%% file: TATW_IT09.tex
\documentclass[journal]{IEEEtran}
\usepackage{color}

\input{preamble_anima}

\begin{document}

\title{A Large-Deviation Analysis of the Maximum- Likelihood Learning of Markov Tree Structures}

\author{Vincent Y.~F.~Tan,~\IEEEmembership{Student Member,~IEEE,}
        Animashree Anandkumar,~\IEEEmembership{Member,~IEEE,}\\
        Lang Tong,~\IEEEmembership{Fellow,~IEEE,} and Alan S. Willsky,~\IEEEmembership{Fellow,~IEEE}
         \thanks{\scriptsize Submitted May 06, 2009. Revised Oct 19, 2010. Accepted Nov 18, 2010.}
          \thanks{\scriptsize  V.\ Y.\ F.\ Tan and A.\ S.\ Willsky are with the
        Stochastic Systems Group, Laboratory for Information and Decision Systems,
        Massachusetts Institute of Technology, Cambridge, MA 02139,
        USA. Email: {\tt\{vtan,willsky\}@mit.edu}. }
                \thanks{\scriptsize  A.\ Anandkumar is with the Center for Pervasive Communications and Computing, Electrical Engineering and
Computer Science Dept., University of California, Irvine, USA 92697. Email: {\tt a.anandkumar@uci.edu}. }
        \thanks{\scriptsize  L.~Tong  is with the School of Electrical and
Computer Engineering, Cornell University, Ithaca, NY 14853, USA.
 Email: {\tt ltong@ece.cornell.edu}.}
 \thanks{\scriptsize This work is supported by A*STAR, Singapore, by a MURI funded through
ARO Grant W911NF-06-1-0076 and by AFOSR Grant FA9550-08-1-0180. This work is also supported    by  UCI setup funds and the Army Research
Office MURI Program under award W911NF-08-1-0238.  The material in this paper was presented  in part at the International Symposium on Information Theory (ISIT), Seoul, Korea, June 2009 \cite{Tan09isit}. }}

% The U. S. Government is authorized to
%reproduce and distribute reprints for Government purposes
%notwithstanding any copyright notation thereon.
\maketitle

\begin{abstract}
The problem of maximum-likelihood (ML) estimation of
discrete tree-structured  distributions is considered. Chow and Liu
 established that  ML-estimation reduces to the construction
of a maximum-weight spanning tree using the empirical mutual
information quantities as the edge weights. Using the theory of large-deviations, we analyze the exponent associated with
 the    error probability of the event that the ML-estimate of the Markov tree structure differs from the true tree structure, given a set of  independently drawn samples.  By exploiting the fact that the output of  ML-estimation is a tree, we establish
that the error exponent is equal to the exponential rate of decay
of a single dominant {\em crossover} event.  We prove that in this
dominant crossover event, a non-neighbor node pair replaces a true edge
of the distribution that is along  the path of edges in the true tree graph
connecting the  nodes in the non-neighbor pair. Using ideas from
Euclidean information theory, we then analyze the scenario of
ML-estimation in the {\em very noisy} learning regime and show that the error
exponent can be approximated as a ratio, which is interpreted as the {\em
signal-to-noise} ratio (SNR) for learning tree distributions. We show via numerical experiments that in this regime, our SNR approximation is accurate.
\end{abstract}

\begin{IEEEkeywords} Error exponent,  Euclidean information theory, Large-deviations principle,   Markov structure,  Maximum-Likelihood distribution estimation, Tree-structured distributions.
\end{IEEEkeywords}

\input{intro_v3_anima}
\input{prelims_v3}
\input{ldp_mi_v3}
\input{errorexp_v4}

\input{euc_v3}

\input{nonunique_v3}

\input{numerical_v2}

\input{conclusion_v2}
\section*{Acknowledgments}

The authors thank the anonymous referees and Associate Editor A.\ Krzyzak  who have helped to  improve the exposition. One reviewer, in particular, helped us highlight the connection of this work with robust hypothesis testing, leading to Section~\ref{sec:ht}. The authors acknowledge  Lizhong Zheng, Marina Meil\u{a}, Sujay Sanghavi, Mukul Agrawal, Alex Olshevsky  and Timo Koski for many stimulating discussions.

%\newpage
\appendices

\input{appendix_v3}

%\newpage
\bibliographystyle{IEEEtran}
\bibliography{isitbib,animabib}

\newpage

\begin{IEEEbiographynophoto}{\bf Vincent Y. F. Tan} (S'07) received the B.A.\ and M.Eng.\ 
degrees in electrical engineering  from Cambridge
University in 2005 and the Ph.D.\ degree  in electrical engineering and computer science  from the Massachusetts
Institute of Technology in 2011.  He was also a research intern with Microsoft Research in   2008 and 2009.  He will join the electrical and computer engineering department at the University of Wisconsin-Madison as a postdoctoral associate in 2011. His  research interests include statistical
signal processing, machine learning and information theory. 

Dr.\ Tan received the Public Service Commission (PSC) Scholarship in 2001
and the National Science Scholarship  (PhD) from the Agency for Science Technology
and Research (A*STAR) in 2006. In 2005, he received the Charles Lamb Prize, a Cambridge University
Engineering Department prize awarded annually to the candidate who
demonstrates the greatest proficiency in electrical engineering. He has
served as a reviewer for the IEEE
Transactions on Signal Processing and  the Journal of Machine Learning Research.
\end{IEEEbiographynophoto}

\begin{IEEEbiographynophoto}
{\bf Anima Anandkumar} (S'02,M'09) received her B.Tech in Electrical
Engineering from the Indian Institute of Technology (IIT) Madras in
2004 and her MS and PhD degrees  in Electrical Engineering from
Cornell University, Ithaca, NY in 2009. She was at the Stochastic
Systems Group at MIT, Cambridge, MA as a post-doctoral researcher. She
has been an assistant professor at EECS Dept. at University of
California Irvine since July 2010.

She is the recipient of the 2009 Best Thesis Award by the ACM
Sigmetrics Society, 2008 IEEE Signal Processing Society Young Author
Best Paper Award, 2008 IBM Fran Allen PhD fellowship, and student
paper award at 2006 IEEE International
Conference on Acoustic, Speech, and Signal Processing (ICASSP). Her
research interests are in the
area of learning and inference of graphical models. She has
served as a reviewer for IEEE Transactions on Information Theory, IEEE
Transactions on Signal Processing, Transactions on Wireless Communications, and IEEE Signal Processing Letters.
\end{IEEEbiographynophoto}

\begin{IEEEbiographynophoto}{\bf Lang Tong} (S'87,M'91,SM'01,F'05) is the Irwin and Joan
Jacobs Professor in Engineering at Cornell University
Ithaca, New York. He received the B.E. degree from Tsinghua University,
Beijing, China, in 1985, and M.S. and Ph.D.
degrees in electrical engineering in 1987 and 1991,
respectively, from the University of Notre Dame, Notre Dame, Indiana.
He was a Postdoctoral Research Affiliate at the Information
Systems Laboratory, Stanford University in 1991.
He was  the 2001 Cor Wit Visiting Professor at
the Delft University of Technology and had held
visiting positions at Stanford University, and U.C. Berkeley.

Lang Tong is a Fellow of IEEE. He received the 1993 Outstanding Young
Author Award from the IEEE Circuits and Systems Society,
the 2004 best paper award (with Min Dong) from IEEE Signal Processing Society,
and the 2004 Leonard G. Abraham Prize Paper Award from the
IEEE Communications Society (with Parvathinathan Venkitasubramaniam
and  Srihari Adireddy). He is also a coauthor of seven student paper awards.
He received Young Investigator Award from the Office of Naval Research.

Lang Tong's research is in the general area of statistical
signal processing, communications and networking,
and information theory.  He has served as an Associate Editor
for the IEEE Transactions on Signal Processing, the IEEE Transactions
on Information Theory, and IEEE Signal Processing Letters.
He was named as a 2009-2010 Distinguished Lecturer by the IEEE Signal
Processing Society.
\end{IEEEbiographynophoto}

\begin{IEEEbiographynophoto}{\bf Alan S. Willsky} (S'70--M'73--SM'82--F'86)
received his S.B. in 1969 and his Ph.D. in 1973 from the Department of Aeronautics and Astronautics at the Massachusetts Institute of Technology, Cambridge.

He joined the  MIT faculty in 
 1973 and is the Edwin Sibley
Webster Professor of Electrical Engineering and
Director of the Laboratory for Information and
Decision Systems. He was a founder of Alphatech,
Inc. and Chief Scientific Consultant, a role in which
he continues at BAE Systems Advanced Information Technologies. From 1998 to 2002, he served on the U.S. Air Force Scientific
Advisory Board. He has delivered numerous keynote addresses and is coauthor
of the text {\em Signals and Systems} (Englewood Cliffs, NJ: Prentice Hall, 1996). His research interests are in the development
and application of advanced methods of estimation, machine learning, and
statistical signal and image processing.

Dr.\ Willsky received several awards including the 1975 American Automatic
Control Council Donald P. Eckman Award, the 1979 ASCE Alfred Noble Prize,
the 1980 IEEE Browder J. Thompson Memorial Award, the IEEE Control Systems
Society Distinguished Member Award in 1988, the 2004 IEEE Donald G.\
Fink Prize Paper Award, the Doctorat Honoris Causa from Universit de Rennes
in 2005 and the IEEE Signal Processing Society Technical Achievement Award in 2010. He and his students, colleagues, and postdoctoral associates have also received a variety of Best Paper Awards at various conferences and for papers in
journals, including the 2001 IEEE Conference on Computer Vision and Pattern
Recognition, the 2003 Spring Meeting of the American Geophysical Union, the
2004 Neural Information Processing Symposium, Fusion 2005, and the 2008
award from the journal Signal Processing for the outstanding paper in the year
2007.
\end{IEEEbiographynophoto}

\end{document}

%% file: preamble_anima.tex
\usepackage[mathscr]{eucal}
\usepackage[cmex10]{amsmath}
\usepackage{epsfig,epsf,psfrag,amssymb,amsfonts,latexsym,url,amsmath,graphicx,bm,cite, xcolor,url}%,cases}
\usepackage{verbatim}
\usepackage{bm}
\usepackage{latexsym, amssymb, amsmath}
\newcommand{\bd}{\begin{definition}}
\newcommand{\ed}{\end{definition}}
\def\bcase{\vskip .1cm \begin{numcases}}
\def\ecase{\end{numcases} \vskip .1cm \noindent}

\def\0{{\mathbf{0}}}
\def\beq{\begin{equation}}
\def\eeq{\end{equation}}
\def\beqn{\vskip .1cm \begin{eqnarray}}
\def\eeqn{\end{eqnarray} \vskip .1cm \noindent}
\def\beqnn{\vskip .01cm \begin{eqnarray*}}
\def\eeqnn{\end{eqnarray*} \vskip .01cm \noindent}

\def\nn{\nonumber}
\newcommand{\defeq}{:=}
\newcommand{\calO}{\mathcal{O}}
\newcommand{\calZ}{\mathcal{Z}}
\newcommand{\calQ}{\mathcal{Q}}
\newcommand{\calY}{\mathcal{Y}}
\newcommand{\calF}{\mathcal{F}}
\newcommand{\calK}{\mathcal{K}}

\newcommand{\calX}{\mathcal{X}}

\def\calC{{\mathcal{C}}}

\newcommand{\calG}{\mathcal{G}}
\newcommand{\tilJ}{\widetilde{J}}
\newcommand{\tilK}{\widetilde{K}}

\newcommand{\diam}{\mathrm{diam}}
\newcommand{\calE}{\mathcal{E}}

\newcommand{\calV}{\mathcal{V}}
\newcommand{\calB}{\mathcal{B}}
\newcommand{\calT}{\mathcal{T}}
\newcommand{\calA}{\mathcal{A}}
\newcommand{\calD}{\mathcal{D}}

\newcommand{\bx}{\mathbf{x}}
\newcommand{\bR}{\mathbb{R}}

\newcommand{\nbd}{\mathrm{nbd}}
\newcommand{\calP}{\mathcal{P}}
\newcommand{\calU}{\mathcal{U}}

\newcommand{\bP}{\mathbb{P}}
\newcommand{\bE}{\mathbb{E}}

\newcommand{\bL}{\mathbf{L}}

\newcommand{\bs}{\mathbf{s}}

\newcommand{\bK}{\mathbf{K}}

\newcommand{\one}{\mathbf{1}}

\newcommand{\argmin}{\operatornamewithlimits{argmin}}
\newcommand{\argmax}{\operatornamewithlimits{argmax}}
\newcommand{\hP}{\widehat{P}}

\newcommand{\estP}{P_{{\text{est}}}}

\newcommand{\hPML}{P_{{\text{\tiny ML}}}}

\newcommand{\hJ}{\widehat{J}}

\newcommand{\hTML}{T_{\text{\tiny ML}}}
\newcommand{\Path}{\mathrm{Path}}

\newcommand{\beps}{\bm{\epsilon}}
\newcommand{\var}{\mathrm{Var}}

\newcommand{\hcalE}{\calE_{{\text{\tiny ML}}}}

\newtheorem{theorem}{Theorem}
\newtheorem{lemma}[theorem]{Lemma}
\newtheorem{proposition}[theorem]{Proposition}
\newtheorem{corollary}[theorem]{Corollary}
\newtheorem{remark}{Remark}
\newtheorem{example}{Example}
\newtheorem{definition}{Definition}
\newtheorem{fact}{Fact}

\newenvironment{myproof}{\noindent{\em Proof:} \hspace*{1em}}{
    \hspace*{\fill} $\Box$ }

\newcommand{\bprf}{\begin{myproof}}
\newcommand{\eprf}{\end{myproof}}

%% file: intro_v3_anima.tex
\section{Introduction}
The estimation of a multivariate distribution from samples is a classical and an important generic problem
in machine learning and statistics and is challenging for high-dimensional
multivariate distributions. In this respect, graphical models~\cite{Lau96}
provide a significant simplification of joint distribution as the distribution can be factorized according to a graph defined on the set of nodes. Many specialized algorithms~\cite{CL68,Che07,Wai06, Lee06,Joh07,Mei06,Dud04} exist for exact and approximate learning of
graphical models Markov on sparse graphs. 

There are many applications of learning graphical models, including  clustering and dimensionality reduction. Suppose we have $d$ genetic variables and
we would like to group the ones that are similar together. Then the construction
of a graphical model provides a visualization of the relationship between genes. Those genes that have
high degree are highly correlated to  many other genes ({\em e.g.}, those in its neighborhood). The learning of a graphical
model may also provide the means to judiciously remove redundant genes from
the model, thus reducing the dimensionality of the data, leading to more efficient
inference of the effects of the genes subsequently.

When the underlying graph is a tree, the Chow-Liu
algorithm~\cite{CL68} provides an efficient method for   the
maximum-likelihood (ML) estimation of the probability distribution
from a set of i.i.d.\ samples drawn  from the distribution. By
exploiting the Markov tree structure, this algorithm  reduces the ML-estimation
problem to solving a maximum-weight spanning tree (MWST) problem. In
this case, it is known that  the ML-estimator  learns
the distribution correctly asymptotically, and hence, is  consistent~\cite{Cho73}.

While consistency is an important qualitative property for any estimator, the study of the rate of convergence, a precise quantitative property, is also of great practical interest. We are interested in the rate of convergence of the ML-estimator (Chow-Liu algorithm) for
tree distributions as we increase the number of samples. 
Specifically,  we study the rate of decay of the error probability or
the error exponent  of the ML-estimator in learning the {\em tree
structure} of the unknown distribution.  A larger exponent means that the error probability in structure learning decays more rapidly. In other words, we need relatively few samples to ensure that the error probability is below some fixed level $\delta>0$.  Such models are thus ``easier'' to learn. We address the following
questions: Is there exponential decay of the probability of error in structure learning
as the number of samples tends to infinity? If so, what is the exact
error exponent, and how does it depend on the parameters of the
distribution? Which edges of the true tree are most-likely to be in
error; in other words, what is the nature of the most-likely
error in the ML-estimator? We provide concrete and intuitive answers to the above
questions, thereby providing insights into how the parameters of the distribution influence the error exponent associated with learning the structure of discrete tree distributions.

\subsection{Main Contributions}
There are three main contributions in this paper. First, using the large-deviation principle (LDP)~\cite{Den00} we
prove that the most-likely error in  ML-estimation is a tree which
differs from the true tree by a single edge. Second, again using the LDP, we derive the
exact  error exponent for ML-estimation of tree structures.  Third,
we provide a succinct and intuitive closed-form approximation for the error
exponent which is tight in the {\em very noisy} learning regime, where the individual
samples are not too informative about the tree structure. The
approximate error exponent has a very intuitive explanation as  the
{\em signal-to-noise ratio} (SNR) for learning.

% and relate them to the tree structure of the true distribution

We analyze the {\em error exponent} (also called the inaccuracy rate) for the
estimation of the structure of the unknown tree distribution. For the error
event that the structure of the ML-estimator $\hcalE$ given $n$
samples differs from the true tree structure $\calE_P$ of the
unknown distribution $P$, the  error exponent is given by
\begin{equation}
K_P ~ := ~ \lim_{n\rightarrow \infty} -\frac{1}{n} \log
\bP(\{\hcalE\ne\calE_P\}). \label{eqn:J_intro}
\end{equation}
To the best of our knowledge, error-exponent analysis for
tree-structure learning has not been considered before (See
Section~\ref{sec:related} for a brief survey of the existing
literature on learning graphical models from data).

Finding the error exponent $K_P$ in~\eqref{eqn:J_intro} is 
not straightforward since  in general, one has to  find the {\em
dominant} error  event with the {\em slowest} rate of decay among all
possible error events~\cite[Ch.\ 1]{Den00}. For
learning the structure of trees, there are a total of $d^{d-2}-1$ possible error
events,\footnote{Since the ML output  $\hcalE$ and the true structure
$\calE_P$ are both spanning trees over $d$ nodes and since there are
$d^{d-2}$ possible spanning trees~\cite{West:book}, we have
$d^{d-2}-1$  number of possible error events.} where $d$ is the
dimension (number of variables or nodes) of the unknown tree distribution $P$. Thus, in principle, one has to consider the information projection \cite{Csis03} of $P$ on all these error trees.  This rules out
brute-force information projection  approaches for finding the error exponent in
\eqref{eqn:J_intro}, especially for high-dimensional data.

%We term these  $d^{d-2}-1$ number of spanning trees, other than the
%true tree structure corresponding to the unknown distribution, as
%the {\em error trees}. The tree having the slowest decay of the
%probability of error of learning it is known as the {\em dominant
%error tree}.

In contrast, we establish that   the search  for the dominant error
event for learning the structure of the tree can be limited to   a polynomial-time  search
space (in $d$).  Furthermore, we establish that  this dominant error
event of the ML-estimator is given by a tree which  differs from the
true tree by only a single edge. We provide a polynomial algorithm
with $\calO(\diam(T_P)\, d^2)$ complexity  to find the error
exponent in~\eqref{eqn:J_intro}, where $\diam(T_P)$ is the diameter of the tree $T_P$. We heavily exploit the mechanism of the ML
Chow-Liu algorithm~\cite{CL68} for tree learning to establish these
results, and specifically, the fact that the ML-estimator tree distribution
depends {\em only} on the  relative order of the empirical mutual
information quantities between all the  node pairs (and not their absolute values).

%We exploit the fact that the ML Chow-Liu learning algorithm outputs $\hcalE$ as a
%MWST using the empirical mutual
%information (MI) quantities as the edge weights. Hence, the relative
%order of the empirical MI quantities between different node pairs
%has an impact on the accuracy of ML-estimation. To this end, for any
%two pairs of nodes in the tree (or equivalently variables of the
%distribution), we   analyze the probability of the crossover event
%that their empirical MI quantities are in the opposite order of
%their actual MI quantities.  Applying Sanov's theorem~\cite[Ch.\ 11]{Cov06}, we
%establish the rate of decay for each crossover event.
%
%It is, however, crucial to note that not all the crossover events
%necessarily lead to an error: even if the order of some of the
%empirical MI quantities are wrong, the final tree output $\hcalE$
%could still be equal to the true tree $\calE_P$. We identify  a set
%of crossover events, each of which always lead to an error in the
%output of the ML-estimator. Here, we exploit the tree structure to note
%that an error definitely occurs if the empirical MI between a non-neighbor node pair  $(u,v)$
%with respect to $\calE_P$ is greater than the empirical MI of some true edge along the unique path connecting the nodes $u$ and $v$.

Although we provide a computationally-efficient way   to compute the
error exponent in \eqref{eqn:J_intro}, it   is not available in
closed-form.  In Section~\ref{sec:euc}, we use  Euclidean
information theory~\cite{Bor06,Bor08} to obtain an approximate error
exponent in closed-form, which can be interpreted as the
signal-to-noise ratio (SNR) for tree structure learning. Numerical
simulations on various discrete graphical models verify that  the
approximation is tight in the very noisy regime.

%The set of such error trees which differ from the true tree in only
%a single edge is obtained as follows: consider any non-neighbor node
%pair with respect to the true tree and replace any edge of the true
%tree which is part of the path connecting the two nodes in the pair.

In Section~\ref{sec:nonunique}, we extend our results  to the
case when the true distribution $P$ is not a tree. In this case, given samples drawn independently from $P$, we intend to learn the {\em optimal projection} $P^*$ onto the set of trees.
Importantly, if $P$ is not a tree, there may be several trees that
are optimal projections~\cite{Cho73} and this requires careful
consideration of the error events.   We derive the error exponent
even in this scenario.

%\Comment Is the error exponent we establish the very best possible?
%
%\Comment I realize that our techniques carry forward (it is a
%subproblem) for some fixed edge to be learned wrong. (or
%neighborhood of some node being wrong): we should emphasize this as
%well: because in structure learning people talk about false
%discovery rates.
%
%\Comment At so many places, I have used ML-estimator. change to MLE.
%
%\Comment I know this amounts to changing a lot of things. But in the
%literature, it is called inaccuracy rate for MLE. Error exponents is
%more for hypothesis testing and we want to avoid this confusion. I
%haven't changed anything for now.

\subsection{Related Work}\label{sec:related}
The seminal work by Chow and Liu in~\cite{CL68} focused on learning tree models from data samples. The authors showed that the learning of the optimal tree distribution essentially decouples into two distinct steps: (i) a structure learning step and (ii) a parameter learning step. The structure learning step, which is the focus on this paper, can be  performed efficiently using a max-weight spanning tree algorithm with the empirical mutual information quantities as the edge weights. The parameter learning step is a maximum-likelihood estimation procedure where the parameters of the learned model are equal to those of the empirical distribution. Chow and Wagner~\cite{Cho73}, in a follow-up paper, studied the consistency properties of the Chow-Liu algorithm for learning trees. They concluded that if the true distribution is Markov on a unique tree structure, then the Chow-Liu learning algorithm is asymptotically consistent. This implies that as the number of samples tends to infinity, the probability that the learned structure differs from the (unique) true structure tends to zero.

Unfortunately, it is known that the exact learning of general graphical models is NP-hard~\cite{Kar01}, but there have been several works to learn approximate models. For example, Chechetka and Guestrin~\cite{Che07} developed good approximations for learning thin junction trees~\cite{Bac02} (junction trees where the sizes of the maximal cliques are small). Heckerman~\cite{Hec95} proposed  learning the structure of Bayesian networks by using the Bayesian Information Criterion~\cite{Schwarz} (BIC) to penalize more complex models and by putting priors on various structures. Other authors used the maximum entropy principle or (sparsity-enforcing) $\ell_1$ regularization as  approximate graphical model learning techniques. In particular, Dudik {\it et al.}~\cite{Dud04} and Lee {\it et al.}~\cite{Lee06} provide strong consistency guarantees on the learned distribution in terms of the log-likelihood of the samples. Johnson {\it et al.}~\cite{Joh07} also used a similar technique known as maximum entropy relaxation (MER) to learn discrete and Gaussian graphical models.  Wainwright {\it et al.}\ \cite{Wai06} proposed a regularization method for learning the graph structure based on $\ell_1$ logistic regression and provided strong theoretical guarantees for learning the correct structure as the number of samples, the number of variables, and the neighborhood size grow. In a similar work, Meinshausen and Buehlmann~\cite{Mei06} considered learning the structure of arbitrary Gaussian models using the Lasso~\cite{Tib96}. They show that the error probability of learning the wrong structure, under some mild technical conditions on the neighborhood size, decays exponentially even when the size of the graph $d$ grows with the number of samples $n$.  However, the rate of decay is not  provided explicitly. Zuk {\it et al.}~\cite{Zuk06} provided bounds on the limit inferior and limit superior of the error rate  for learning the structure of  Bayesian networks but, in contrast to our work, these bounds are not asymptotically tight. In addition, the work in Zuk {\it et al.}~\cite{Zuk06} is intimately tied to the BIC~\cite{Schwarz}, whereas our analysis is for the Chow-Liu ML tree learning algorithm~\cite{CL68}. A modification of the Chow-Liu learning algorithm has also been applied to learning the structure of latent trees where only a subset of variables are observed \cite{Choi10}. 

There have also been a series of papers~\cite{Pan03, Ant01, Cha05, Hut01} that
quantify the deviation of the empirical information-theoretic quantities from
their true values by employing techniques from large-deviations
theory. Some ideas from these papers will turn out to be important
in the subsequent development because we exploit conditions under which the empirical mutual information
quantities do not differ ``too much'' from their nominal values. This
will ensure that structure learning succeeds with high probability.
\subsection{Paper Outline}
This paper is organized as follows: In Sections~\ref{sec:prelim} and~\ref{sec:CL},
we state the system model and the problem statement and provide the necessary preliminaries on undirected
graphical models and the Chow-Liu algorithm~\cite{CL68} for learning tree distributions. In Section~\ref{sec:ldp_mi}, we derive an analytical expression for the crossover rate of two node pairs.
We then relate the crossover rates to the overall error exponent in Section~\ref{sec:ee}.  We also discuss some connections of the problem we solve here with robust hypothesis testing. 
In Section~\ref{sec:euc}, we leverage on ideas in Euclidean information theory to state sufficient conditions that allow approximations of the crossover rate and the error exponent. We obtain an intuitively appealing closed-form expression.  By redefining the error event, we extend our results to the case when the true distribution is not a tree in Section~\ref{sec:nonunique}. We compare the true and approximate crossover rates by performing numerical experiments for a given graphical model in Section~\ref{sec:num}. Perspectives and extensions are discussed in Section~\ref{sec:concl}.

%% file: prelims_v3.tex
\section{System Model and Problem Statement}\label{sec:prelim}
\subsection{Graphical Models}
An {\em undirected graphical model}~\cite{Lau96} is a probability 
distribution that factorizes according to the structure of an 
underlying undirected graph. More explicitly, a vector of random 
variables $\bx\defeq[x_1, \ldots, x_d]^T$ is said to be {\em Markov} 
on a graph $\calG=(\calV,\calE)$ with vertex set $\calV=\{1,\ldots , 
d\}$ and edge set  $\calE\subset  \binom{\calV}{2} $ if 
\begin{equation}
 P(x_i|x_{\calV \setminus \{i\}}) ~=~ P(x_i|x_{\nbd(i)}),\quad \forall\, i\in \calV, \label{eqn:localMarkov}
\end{equation}
where $\nbd(i)$ is the set of neighbors of $i$ in $\calG$, {\it i.e.}, $\nbd(i):=\{j\in\calV: (i,j)\in \calE\}$. Eq.\ 
\eqref{eqn:localMarkov} is called the (local) Markov property and 
states that  if random variable $x_i$ is conditioned on its neighboring random variables, then $x_i$ is independent of the rest of the variables in the graph.

In this paper, we assume that each random variable $x_i\in \calX$, and we also
assume that  $\calX=\{1,\ldots,|\calX|\}$ is a {\em known finite} set.\footnote{The analysis of learning the
structure of jointly Gaussian variables where $\calX=\bR$ is deferred to a companion paper \cite{Tan10sp}. The subsequent analysis carries over straightforwardly to the case where $\calX$ is a countably infinite set. } Hence, 
the joint distribution $P\in  \calP(\calX^d)$, where $\calP(\calX^d)$ is the probability simplex of all distributions supported on 
$\calX^d$. 

%Formally, the underlying probability (measure) space is $(\calX^d, 
%\wp( \calX^d),P)$, where $\wp( \calX^d)$ is the power set (set of all subsets) of 
%$\calX^d$ and $P$ is a probability measure (or distribution) on the 
%$\sigma$-field $\wp( \calX^d)$. 

Except for Section~\ref{sec:nonunique}, we limit our analysis in this paper to the set of strictly 
positive\footnote{A distribution $P$ is said to be strictly positive if  
$P(\bx)>0$ for all $\bx\in \calX^d$.} graphical models $P$, in which  
the graph of $P$ is a   tree on the  $d$ nodes, denoted 
$T_P=(\calV, \calE_P)$. Thus, $T_P$ is an undirected, acyclic and 
connected graph with vertex set $\calV=\{1,\ldots , d\}$ and edge 
set  $\calE_P$, with $d-1$ edges.  Let $\calT^d$ be the set of 
{\em spanning trees} on $d$ nodes, and hence, $T_P \in \calT^d$.  Tree 
distributions possess the following factorization property~\cite{Lau96} 
\begin{equation}
P(\bx) ~=~ \prod_{i\in \calV} P_i(x_i)  \prod_{(i,j)\in \calE_P} \frac{P_{i,j}(x_i, x_j) }
{P_i(x_i) P_j(x_j)},\label{eqn:tree}
\end{equation}
where $P_i$ and $P_{i,j}$ are the marginals on node $i\in \calV$ and 
edge $(i,j)\in \calE_P$ respectively. Since $T_P$ is spanning, $P_{i,j}\ne P_iP_j$ for all $(i,j)\in \calE_P$. Hence, there is a substantial 
simplification of the joint distribution which arises from the Markov tree dependence. In 
particular, the distribution is completely specified by the set of 
edges $\calE_P$ and  pairwise marginals $P_{i,j}$ on the edges of 
the tree $(i,j)\in \calE_P$. In Section~\ref{sec:nonunique}, we 
extend our analysis to general distributions which are not 
necessarily Markov on a tree.

\subsection{Problem Statement}

In this paper, we consider a learning problem, where we are given a 
set of $n$ i.i.d.  
$d$-dimensional samples   
 $\bx^n ~\defeq~ \{ 
\bx_1, \ldots, \bx_n\}$ from an unknown  distribution   $P\in 
\calP(\calX^d)$, which is Markov with respect to a tree $T_P\in \calT^d$. Each sample or observation  $\bx_k := [x_{k,1}, \ldots, 
x_{k,d}]^T$ is a vector of $d$ dimensions where each entry can only take on one of a finite number of values in the alphabet $\calX$. 

Given $\bx^n$, the ML-estimator of the unknown 
distribution $P$ is defined as
\begin{equation} 
\hPML ~\defeq~ \argmax_{Q\in\calD(\calX^d,\calT^d)} \,\,   \sum_{k=1}^n 
\log Q(\bx_k), \label{eqn:clopt1}
\end{equation} 
where $\calD(\calX^d,\calT^d)\subset \calP(\calX^d)$ is defined as the set of all tree distributions 
on the alphabet $\calX^d$ over $d$ nodes. 

In 1968, Chow and Liu showed 
that the above ML-estimate $\hPML$ can be found efficiently via a MWST algorithm~\cite{CL68}, and is described in 
Section \ref{sec:CL}. We denote the tree graph of the ML-estimate $\hPML$ by 
$\hTML=(\calV, \hcalE)$ with   vertex set $\calV$ and edge set 
$\hcalE$.

Given a tree distribution $P$, define the probability of the error 
event that the set of edges is {\em not} estimated correctly by the 
ML-estimator as
\begin{equation}
\calA_n~\defeq~\left\{\hcalE \ne \calE_P \right\} \label{eqn:err}
\end{equation}
We denote $\bP:=P^n$ as the $n$-fold {\em  product probability measure} of the $n$ samples $\bx^n$ which are  drawn i.i.d.\ from $P$. In this paper, we are interested in studying the {\em rate} or {\em error exponent}\footnote{In the maximum-likelihood estimation literature (e.g.\ \cite{Kester&Kallenberg:86Stat, Bahadur&Zabell&Gupta:80}) if the limit in~\eqref{eqn:J} exists,  $K_P$ is also typically known as the inaccuracy rate. We will be using the terms rate, error exponent and inaccuracy rate interchangeably in the sequel. All these terms refer to $K_P$. }  $K_P$ at which the above
error probability exponentially decays with the number of samples 
$n$, given by,  
\begin{equation}
K_P ~\defeq~  \lim_{n\rightarrow \infty} -\frac{1}{n} \log 
\bP(\calA_n), \label{eqn:J}
\end{equation}
whenever the limit exists. Indeed, we will prove that the limit in~\eqref{eqn:J} exists in the sequel. With the $\doteq$ notation\footnote{The 
  $\doteq$ notation (used in \cite{Cov06}) denotes equality to the first order in the 
exponent. For two real sequences $\{a_n\}$ and $\{b_n\}$, $a_n\doteq 
b_n$ if and only if $\lim_{n\rightarrow \infty} 
\frac{1}{n}\log(a_n/b_n) = 0$.}, \eqref{eqn:J} can be written as 
\begin{equation}
\bP(\calA_n) ~\doteq~ \exp(-nK_P).\label{eqn:Kdoteq}
\end{equation}
A positive error exponent  $(K_P>0)$ implies an  
exponential decay of error probability in ML structure learning, and 
we will establish necessary and sufficient conditions to ensure  
this.

Note that we are only interested in quantifying the probability of 
the error in learning the {\em structure} of $P$ in~\eqref{eqn:err}. We 
are not concerned about the parameters that define the ML tree 
distribution $\hPML$.  Since there are only finitely many (but a
super-exponential number of) structures, this is in fact akin to an 
ML problem where the parameter space is discrete and finite~\cite{Ryu99}. Thus, under some mild technical conditions,  we can expect exponential decay in the probability of error as mentioned in~\cite{Ryu99}. Otherwise, we can only expect convergence with rate $\mathcal{O}_p(1/\sqrt{n})$ for  estimation of parameters that belong to a continuous parameter space~\cite{Vantrees}. In this work, we  quantify  the error exponent for learning tree structures using the ML learning procedure precisely.

%We will address the optimality the ML learning algorithm with respect to the error exponent~\eqref{eqn:J} in Section~\ref{sec:bahadur_opt}.
%\Comment {\bf At this moment this is not clear at all, not 
%appropriate here} It is known that as $n\rightarrow \infty$, the 
%empirical distribution $\hP$ tends to the true distribution $P$  
%\cite{Cov06}. Thus, $\hcalE$ approaches $\calE_P$, the true tree 
%structure, since $\hPML$ also approaches $P$ \Comment {\bf how is 
%this known?}  But at what rate does this happen for a given tree 
%distribution $P$? Is the error decay exponential? In this paper, we 
%use the LDP~\cite{Dembo} to quantify the exponential rate at which 

\section{Maximum-Likelihood Learning of Tree Distributions from Samples}\label{sec:CL}
In this section, we review the classical Chow-Liu 
algorithm~\cite{CL68} for learning the ML tree 
distribution $\hPML$ given a set of $n$ samples $\bx^n$ drawn i.i.d. 
from a tree distribution $P$. Recall the ML-estimation problem in~\eqref{eqn:clopt1}, where $\hcalE$ denotes the set of edges of the 
tree $\hTML$ on which $\hPML$ is tree-dependent.  Note that since 
$\hPML$ is tree-dependent, from~\eqref{eqn:tree}, we have the result 
that it is completely specified by the structure $\hcalE$ and 
consistent pairwise marginals $\hPML(x_i,x_j)$ on its edges 
$(i,j)\in \hcalE$.

In order to obtain the ML-estimator, we need the notion of a {\em 
type} or {\em empirical distribution} of $P$, given $\bx^n$, defined as
\begin{equation}
\hP(\bx;\bx^n)\defeq \frac{1}{n}\sum_{k=1}^n 
\mathbb{I}\{ \bx_k = \bx\},\label{eqn:emp_dis}
\end{equation}
where $\mathbb{I}\{\bx_k=\bx\}=1$ if $\bx_k=\bx$ and equals 0 otherwise.  For convenience, in the rest of the paper, we 
will denote the empirical distribution by  $\hP(\bx)$ instead of 
$\hP(\bx;\bx^n)$. 

%It is known that the type is a {\em sufficient 
%statistic} for estimating $P$ from i.i.d.\ samples~\cite[Ch.\ 11]{Cov06}. The Chow-Liu algorithm~\cite{CL68} is  based, in part, on the following fact. 

\begin{fact}
The ML-estimator in~\eqref{eqn:clopt1} is equivalent to the following optimization problem:
\begin{equation} 
\hPML ~=~\argmin_{Q\in\calD(\calX^d,\calT^d)} \,\, D(\hP  \, ||\, Q), 
\label{eqn:clopt2}
\end{equation}
where $\hP$ is the empirical distribution of $\bx^n$, given by~\eqref{eqn:emp_dis}. In~\eqref{eqn:clopt2}, $D(\hP\,||\, Q)=\sum_{\bx\in\calX^d}\hP(\bx)\log\frac{\hP(\bx)}{Q(\bx) }$ denotes the Kullback-Leibler divergence (or relative entropy) \cite[Ch.\ 1]{Cov06} between the probability distributions $\hP,Q \in\calP(\calX^d)$.
\end{fact}
\begin{IEEEproof} 
By the definition of the KL-divergence, we have 
\begin{align}
nD(\hP  \, ||\, Q) &= -nH(\hP) -n\sum_{\bx\in\calX^d} \hP(\bx) \log Q(\bx) ,   \\
&= -nH(\hP)   - \sum_{k=1}^n  \log Q(\bx_k) , \label{eqn:prob_mass}
\end{align} where we use the fact that  the empirical distribution $\hP$ in \eqref{eqn:emp_dis} 
assigns a probability 
mass of $1/n$ to each sample $\bx_k$.
% Since the first term in 
%\eqref{eqn:prob_mass} is a constant for the optimization over $Q$, 
%the statement holds.
\end{IEEEproof}
The minimization over the second variable in \eqref{eqn:clopt2} is also known as the {\em reverse I-projection}~\cite{Csi84,Csis03} of $\hP$ onto the set of tree distributions $\calD(\calX^d,\calT^d)$. We now state the main result of the Chow-Liu tree learning algorithm~\cite{CL68}. In this paper, 
with a slight abuse of notation, we denote the mutual information  $I(x_i;x_j)$
between two random variables $x_i$ and $x_j$ corresponding to nodes $i$ and $j$ as:  
\begin{equation}
I(P_{i,j}) := \sum_{(x_i,x_j)\in 
\calX^2} P_{i,j}(x_i,x_j)\log\frac{P_{i,j}(x_i,x_j)}{P_i(x_i)P_j(x_j)}  .
\end{equation}
Note that the definition above uses only the marginal of $P$ restricted to $(x_i,x_j)$.  If  $e=(i,j)$, then we will also denote the mutual information as $I(P_e)=I(P_{i,j})$. 

%\subsection{The Chow-Liu Algorithm for Learning Trees}
\begin{theorem}[Chow-Liu Tree Learning~\cite{CL68}] The structure and parameters of the ML-estimate $\hPML$ in \eqref{eqn:clopt1} are given by 
\begin{align} \hcalE &~=~ 
\argmax_{\calE_Q : Q \in \calD(\calX^d,\calT^d)} \,\, \sum_{e\in 
\calE_Q} I(\hP_e), \label{eqn:mwst}\\
\hPML(x_i,x_j)&~=~ \hP_{i,j}(x_i,x_j), \qquad \forall\,  (i,j)\in 
\hcalE, \label{eqn:params2} 
\end{align} 
where $\hP$ is the empirical 
distribution in~\eqref{eqn:emp_dis} given the data $\bx^n$, and 
$I(\hP_e)=I(\hP_{i,j})$ is the {\em empirical mutual information} of random variables $x_i$ and $x_j$, which is a function of the empirical distribution $\hP_{e}$.
 \end{theorem}
\begin{IEEEproof} For a fixed tree distribution $Q \in \calD(\calX^d,\calT^d)$,   
$Q$ admits the factorization in \eqref{eqn:tree}, and we have
\begin{align}
& D(\hP\,||\, Q) + H(\hP)\nn \\
 = &-\!\sum_{\bx\in \calX^d} \!\hP(\bx) \log \left[  \prod_{i\in \calV} Q_i(x_i) \!\! \prod_{(i,j)\in \calE_Q}\!\! \frac{Q_{i,j}(x_i, x_j) }{Q_i(x_i) Q_j(x_j)} \!\right], \\
=& - \sum_{i\in \calV} \sum_{x_i\in \calX} \hP_i(x_i) \log Q_i(x_i)  \nn \\
 & - \sum_{(i,j)\in \calE_Q} \sum_{(x_i,x_j) \in \calX^2} \hP_{i,j}(x_i,x_j) \log \frac{Q_{i,j}(x_i, x_j) }{Q_i(x_i) Q_j(x_j)}. \label{eqn:cl_marg}
\end{align}
For a fixed structure $\calE_Q$, it can be shown~\cite{CL68} that 
the above quantity is minimized when the pairwise marginals over the 
edges of $\calE_Q$ are set to that of $\hP$, {\it i.e.}, for all $Q\in \calD(\calX^d,\calT^d)$,   
\begin{align}
& D(\hP\,||\, Q) + H(\hP)\nn \\
\geq &-\sum_{i\in \calV} \sum_{x_i\in \calX} \hP_i(x_i) \log \hP_i(x_i)  \nn\\
 &-\sum_{(i,j)\in \calE_Q} \sum_{(x_i,x_j) \in \calX^2} \hP_{i,j}(x_i,x_j) \log \frac{\hP_{i,j}(x_i, x_j) }{\hP_i(x_i) \hP_j(x_j)}. \label{eqn:cl_proj}
\\ =& \sum_{i\in \calV} H(\hP_i) -   \sum_{(i,j)\in \calE_Q} I(\hP_e). \label{eqn:cl_final}
\end{align}  The first term in~\eqref{eqn:cl_final} is a 
constant with respect to $Q$. Furthermore, since $\calE_Q$ is the 
edge set of the tree distribution $Q\in\calD(\calX^d,\calT^d)$, the 
optimization for the ML tree distribution $\hPML$ reduces to the MWST
search for the optimal edge set as in~\eqref{eqn:mwst}.
\end{IEEEproof}

Hence, the optimal tree probability distribution $\hPML$ is the 
reverse I-projection of $\hP$ onto the optimal tree structure given by 
\eqref{eqn:mwst}.  Thus, the optimization problem in 
\eqref{eqn:clopt2} essentially reduces to a search for the {\em structure} of 
$\hPML$. The structure of $\hPML$ completely determines its 
distribution, since the parameters are given by the empirical 
distribution in \eqref{eqn:params2}.  To solve~\eqref{eqn:mwst}, we 
use the samples $\bx^n$ to compute the empirical distribution $\hP$ 
using \eqref{eqn:emp_dis}, then use $\hP$ to compute $I(\hP_e)$, for 
each node pair $e\in \binom{\calV}{2}$. Subsequently, we use the set of 
empirical mutual information quantities $\{I(\hP_e): e\in \binom{\calV}{2}\}$ 
as the edge weights for the MWST problem.\footnote{If we use the true 
mutual information quantities as inputs to the MWST, then   the true edge set
$\calE_P$ is  the output.} 

%Note that the search for the MWST is not the same as that for 
%largest set of mutual information quantities as one has to take into 
%consideration the spanning tree constraint. 

We  see that the Chow-Liu MWST spanning tree algorithm is an 
efficient way of solving the ML-estimation problem, especially when 
the dimension $d$ is large. This is because there are $d^{d-2}$ 
possible spanning trees over $d$ nodes~\cite{West:book} ruling out 
the possibility for performing an exhaustive search for the optimal 
tree structure. In contrast, the MWST can be found, say using  
Kruskal's algorithm~\cite{Cor03,Kruskal}  or Prim's algorithm~\cite{Prim}, 
in $\mathcal{O}(d^2\log d)$ time. 

%% file: ldp_mi_v3.tex
\section{LDP for Empirical Mutual Information}\label{sec:ldp_mi}
The goal of this paper is to characterize the error exponent for ML tree learning $K_P$ in~\eqref{eqn:J}. As a first step, we
consider a simpler event, which may potentially lead to an error in
ML-estimation. In this section, we derive the LDP rate  for this event, and in the next
section, we use the result to derive $K_P$, the exponent associated to the error event $\calA_n$ defined in \eqref{eqn:err}.

Since the ML-estimate uses the empirical mutual information  quantities as
the edge weights for the MWST algorithm, the  relative values of the
empirical mutual information  quantities have an impact on the accuracy of ML-estimation. In other words, if  the order of these empirical
quantities is different from the true order then it can potentially
lead to an error in the estimated edge set. Hence, it is crucial to
study the probability of the event that the empirical mutual information  quantities
of any two node pairs is different from the true order.
%
%The occurrence of the event {\em may} potentially lead to an error
%in structure learning. However, it is important that the occurrence
%of such an event does not necessarily mean that the learned
%structure $\hcalE$ will be incorrect. This is because the learned
%structure has to be a tree hence the MWST algorithm precludes the
%inclusion of some edges. In the next section, we will describe such
%crossover events relate to error event of interest in
%\eqref{eqn:err}.

Formally, let us consider two distinct node pairs  with no common nodes $e
,e' \in \binom{\calV}{2}$ with unknown distribution $P_{e,e'}\in
\calP(\calX^4)$, where the notation $P_{e,e'}$ denotes the marginal of the tree-structured graphical model $P$ on the nodes in the set $\{e,e'\}$. Similarly, $P_e$ is the marginal of $P$ on edge $e$. Assume that the order
of the true mutual information quantities follow $I(P_e)>I(P_{e'})$.
A {\em crossover event}\footnote{The event $\calC_{e,e'}$ in \eqref{eqn:Aee} depends on the number of samples $n$ but we suppress this dependence for convenience. } occurs if the corresponding empirical mutual
information quantities are of the reverse order, given by
\begin{equation}
\calC_{e,e'}~\defeq~ \left\{I(\hP_e) \le I(\hP_{e'})   \right\}.
\label{eqn:Aee}
\end{equation}
As the number of samples $n \to \infty$, the
empirical quantities approach the true ones, and hence, the
probability of the above event decays to zero. When the decay is
exponential, we have a LDP for the above event, and we term its rate
 as the {\em crossover rate for empirical mutual
information} quantities, defined as
\begin{equation}
J_{e,e'} ~\defeq~ \lim_{n\rightarrow \infty} -\frac{1}{n} \log \bP\left(\calC_{e,e'}  \right), \label{eqn:Jee} %=\lim_{n\rightarrow \infty} -\frac{1}{n} \log  \bP( \hI_{e'} \ge \hI_e )  \label{eqn:Jee}
\end{equation}
assuming the limit in~\eqref{eqn:Jee} exists. Indeed, we show in the proof of Theorem~\ref{thm:cross_emp} that the limit exists. Intuitively (and as seen in our numerical simulations in Section~\ref{sec:num}), if the difference between the true mutual information
quantities $I(P_e)- I(P_{e'})$ is large ({\it i.e.}, $I(P_e)\gg I(P_{e'})$), we
expect  the probability of the crossover event $\calC_{e,e'}$  to be small. Thus, the rate of decay would be faster
 and hence, we expect the crossover rate $J_{e,e'}$ to be large.
In the following, we  see that  $J_{e,e'}$ depends not only on the difference
of mutual information quantities $I(P_e)-I(P_{e'})$, but also on
the {\em distribution} $P_{e,e'}$ of the variables on node pairs $e$ and
$e'$, since the distribution $P_{e,e'}$ influences the accuracy of
estimating them.

\begin{theorem}[Crossover Rate for Empirical MIs]\label{thm:cross_emp}
The crossover rate for
a pair of empirical mutual information quantities in \eqref{eqn:Jee}
is given by
\begin{eqnarray}
J_{e,e'} ~=~ \inf_{Q\in \calP(\calX^4) } \left\{D(Q\, ||\, P_{e,e'}) :
I(Q_{e'}) =  I(Q_e)\right\}, \label{eqn:Jee2}
\end{eqnarray}
where $Q_e,Q_{e'}\in \calP(\calX^2)$ are marginals of $Q$ over node
pairs $e$ and $e'$, which do not share common nodes, {\it i.e.},
\begin{subequations}\label{eqn:Qe}
\begin{align}  
Q_e(x_e)&~:=~\sum_{x_{e'}\in\calX^2} Q(x_e,x_{e'}),\\
Q_{e'}(x_{e'}) &~:=~\sum_{x_e\in\calX^2} Q(x_e,x_{e'}). \end{align}
\end{subequations}
The infimum in~\eqref{eqn:Jee2} is attained by some distribution $Q^*_{e,e'}\in \calP(\calX^4)$ satisfying
$I(Q_{e'}^*) =  I(Q_{e}^*)$ and $J_{e,e'}>0$.
\end{theorem}
\begin{IEEEproof} ({\it Sketch})
The proof hinges on Sanov's theorem~\cite[Ch.\ 11]{Cov06} and the contraction principle in large-deviations~\cite[Sec.\
III.5]{Den00}. The existence of the minimizer follows from the compactness of the constraint set and
Weierstrass' extreme value theorem~\cite[Theorem 4.16]{Rudin}. The rate
$J_{e,e'}$  is strictly positive since we assumed, {\em a-priori}, that
the two node pairs $e$ and $e'$ satisfy $I(P_e)
> I(P_{e'})$. As a result, $Q_{e,e'}^*\ne P_{e,e'}$ and  $D(Q_{e,e'}^*\,||\,
P_{e,e'})>0$. See Appendix~\ref{prf:cross_emp} for the details.
\end{IEEEproof}

In the above theorem, which is analogous to Theorem 3.3 in~\cite{Cha05}, we derived the crossover rate
$J_{e,e'}$ as a constrained minimization  over a submanifold of
distributions in $\calP(\calX^4)$ (See Fig.\ \ref{fig:proj}), and also proved the existence of an
optimizing distribution $Q^*$. However, it is not easy to further
simplify the rate expression in \eqref{eqn:Jee2} since the
optimization is non-convex.

Importantly, this means that it is not clear how the parameters of
the distribution $P_{e,e'}$ affect the rate $J_{e,e'}$, hence ~\eqref{eqn:Jee2} is not  intuitive to aid in understanding the relative ease or difficulty in estimating particular tree-structured distributions. In Section~\ref{sec:euc}, we assume that $P$ satisfies some (so-called very noisy learning) conditions and use Euclidean information
theory~\cite{Bor06,Bor08} to approximate the rate in
\eqref{eqn:Jee2} in order to gain insights as to how the distribution
parameters affect the crossover rate $J_{e,e'}$ and ultimately, the
error exponent $K_P$ for learning the tree structure.

\begin{remark}
Theorem \ref{thm:cross_emp} specifies the crossover rate $J_{e,e'}$
when the two node pairs $e$ and $e'$ do not have any common nodes.
If $e$ and $e'$ share  one node, then the distribution  $P_{e,e'}\in
\calP(\calX^3)$ and here, the crossover rate for empirical mutual
information is
\begin{eqnarray}
J_{e,e'}~=~ \inf_{Q\in \calP(\calX^3) } \left\{D(Q\, ||\, P_{e,e'}) :
I(Q_{e'}) = I(Q_e)\right\}. \label{eqn:Jee_3nodes}
\end{eqnarray}
%The results in the sequel do not depend on whether $e$ and $e'$ share a common node.
\end{remark}
In Section~\ref{sec:euc}, we obtain an approximate closed-form expression for $J_{e,e'}$. The expression, provided in Theorem~\ref{thm:euc}, does not depend on whether $e$ and $e'$ share a node.

%The majority of the results in the sequel do not depend on whether
%$e$ and $e'$ share a node. Hence, we will assume from now, for
%simplicity, that $e$ and $e'$ do not share a node.

\subsection*{Example: Symmetric Star  Graph}\label{sec:star}
\begin{figure}
\centering
\input{star}
\caption{The star graph with $d=9$. $Q_a$ is the joint distribution on any pair of variables that form an edge e.g., $x_1$ and $x_2$. $Q_b$ is the joint distribution on any pair of variables that do not form an edge e.g., $x_2$ and $x_3$. By symmetry, all crossover rates are equal. }
\label{fig:star}
\end{figure}
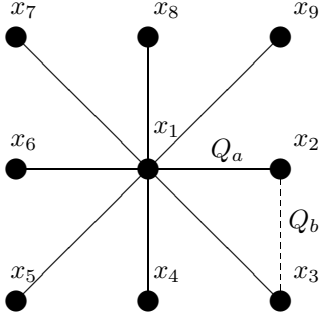

It is now instructive to study a simple example to see how the
overall error exponent $K_P$ for structure learning in~\eqref{eqn:J}
depends on the set of crossover rates $\{J_{e,e'}: e,e'\in
\binom{\calV}{2}\}$. We consider a graphical model $P$ with an associated
tree $T_P=(\calV, \calE_P)$ which is a $d$-order  star with a
central node $1$ and outer nodes $2,\ldots, d$, as shown in Fig.\
\ref{fig:star}. The edge set is given by $\calE_P=\{(1,i):
i=2,\ldots, d\}$.

We assign the joint distributions $Q_a, Q_b \in \calP(\calX^2)$ and
$Q_{a,b}\in \calP(\calX^4)$ to the variables in this graph in the
following specific way:
\begin{enumerate}
\item  $P_{1,i} \equiv Q_a$ for all $2\le i\le d$.
\item $P_{i,j} \equiv Q_b$ for all $2\le i,j \le d$, $i\ne j$.
\item $P_{1,i,j,k} \equiv Q_{a,b}$ for all $2\le i,j,k\le d$, $i\ne j \ne k$.
\end{enumerate}
Thus, we have identical pairwise  distributions $P_{1,i}\equiv Q_a$
of the central node $1$ and any other node $i$, and also identical
pairwise distributions $P_{i,j}\equiv Q_b$ of any  two distinct
outer nodes $i$ and $j$.  Furthermore, assume that
$I(Q_a)>I(Q_b)>0$. Note that the distribution
$Q_{a,b}\in\calP(\calX^4)$ completely specifies the above
graphical model with a  star graph. Also, from the above specifications, we see that $Q_a$ and $Q_b$ are the marginal distributions of $Q_{a,b}$ with respect to to node pairs $(1,i)$ and $(j,k)$ respectively {\it i.e.},
\begin{subequations} \label{eqn:Qa}
\begin{align}
Q_{a}(x_1,x_i) &=\sum_{(x_j,x_k) \in\calX^2}P_{1,i,j,k}(x_1,x_i,x_j,x_k),\\
Q_{b}(x_j,x_k) &=\sum_{(x_1,x_i)\in\calX^2}P_{1,i,j,k}(x_1,x_i,x_j,x_k).
\end{align}
\end{subequations}

Note that each  crossover event between any non-edge $e'$ (necessarily of length 2) and an edge $e$ along its path results
in an error in the learned structure since it leads to $e'$ being
declared an edge instead of $e$. Due to the symmetry, all
such crossover rates between pairs $e$ and $e'$  are equal.
 By the ``worst-exponent-wins" rule \cite[Ch.\ 1]{Den00}, it is more likely to have a single
crossover event than multiple ones. Hence, the error exponent is
equal to the crossover rate between an edge and a non-neighbor pair
in the symmetric star graph. We state this formally in the following proposition.

\begin{proposition}[Error Exponent for symmetric star graph]\label{prop:star}
For the symmetric graphical model with star graph and   $ Q_{a,b}$
as described above,  the   error exponent for structure learning $K_P$
in \eqref{eqn:J}, is  equal to the crossover rate between an edge
and a non-neighbor node pair \begin{equation} K_P ~=~J_{e,e'}, \quad \text{  for any  }\quad  e\in \calE_P, \,
e'\notin \calE_P,\label{eqn:star_rate}\end{equation} where from
\eqref{eqn:Jee2},  the crossover rate is given by \begin{equation} J_{e,e'}\!=\! \inf_{R_{1,2,3,4}\in
\calP(\calX^4) }\!\! \left\{D(R_{1,2,3,4}  || Q_{a,b}) :
I(R_{1,2}) \!=\! I(R_{3,4})\right\},  \end{equation} with $R_{1,2}$ and $R_{3,4}$
as the marginals of $R_{1,2,3,4}$, {\it e.g.},% {\it i.e.},
\begin{equation} \label{eqn:Rmarg}
R_{1,2}(x_1,x_2) =\sum_{(x_3,x_4)\in\calX^2} R_{1,2,3,4}(x_1,x_2,x_3,x_4).%,\quad R_{3,4} =\sum_{x_1,x_2} R_{1,2,3,4}.
\end{equation}
\end{proposition}

%\begin{equation} R_{1,2}\defeq\sum\limits_{x_{3,4}} R_{1,2,3,4},\quad
%R_{3,4}\defeq \sum\limits_{x_{1,2}} R_{1,2,3,4}.\end{equation}

\begin{IEEEproof}
Since there are only two distinct distributions $Q_a$ (which corresponds to a true edge) and $Q_b$ (which corresponds to a non-edge), there is only {\em one} unique rate $J_{e,e'}$, namely the expression in~\eqref{eqn:Jee2} with $P_{e,e'}$ replaced by $Q_{a,b}$. If the event $\calC_{e,e'}$, in \eqref{eqn:Aee}, occurs, an error definitely occurs. This corresponds to the case where {\em any one} edge $e\in \calE_P$ is replaced by {\em any other} node pair $e'$ not in $\calE_P$.\footnote{Also see  theorem~\ref{thm:dom} and its proof for the argument that the dominant error tree differs from the true tree by a single edge.}
\end{IEEEproof}

%\begin{figure}
%\centering{
%\begin{psfrag}
%\psfrag{1}[l]{\scriptsize$1$}\psfrag{2}[l]{\scriptsize$2$}
%\includegraphics[width=.65\columnwidth]{figs/star}
%\end{psfrag}}
%\caption{The ``star'' graph with $d=9$. See the description of this
%example and Prop.\ \ref{prop:star} in the text.} \label{fig:star}
%\end{figure}
%For general graph structures, however, we must first come up with upper and lower bounds for the error event $\calA_n$
%\begin{equation}
%J_U\le\lim_{n\rightarrow \infty}-\frac{1}{n} \log \bP(\calA_n) \le J_L\label{eqn:JU}
%\end{equation}
%where $J_L$ and $J_U$ are functions of $\{J_{e,e'}\}$. It turns out that in the above ``star'' graph example, $J=J_U=J_L$. However, without any special structure on $P$, the bounds in~\eqref{eqn:JU} are, in general, loose. For example, for the three-node case ($d=3$),
%\begin{equation}
%J_L=\max\left\{J_{e_1,e_3},J_{e_2,e_3} \right\}, \,\, J_U=\min\left\{J_{e_1,e_3},J_{e_2,e_3} \right\}
%\end{equation}
%where the edges $e_1,e_2,e_3$ satisfy $I_{e_1}>I_{e_2}>I_{e_3}$. For graphs with more than three nodes, the upper and lower bounds will be even looser.

Hence, we have derived the error exponent for learning a symmetric star
graph through the crossover rate $J_{e,e'}$ between any node pair
$e$ which is an edge in the star graph and another node pair $e'$
which is not an edge.

The symmetric star graph possesses symmetry in the distributions and hence it is easy to relate $K_P$ to a sole crossover rate. In general, it is not straightforward to derive the error exponent
$K_P$ from the set of crossover rates $\{J_{e,e'}\}$ since they may not all be
equal and more importantly, crossover events for different node
pairs affect the learned structure $\hcalE$ in a complex manner. In the next
section, we provide an exact expression for $K_P$ by identifying the (sole)
crossover event related to a dominant error tree.  Finally, we remark that the  crossover event $\calC_{e,e'}$ is  related to the notion of neighborhood selection in the graphical model learning literature~\cite{Mei06,Wai06}.

% Note that the rate
%in \eqref{eqn:star_rate} is exactly equal to \eqref{eqn:Jee2} with
%$P_{e,e'}$ replaced by $Q_{a,b}$. In this case, it is easy to relate
%$K$ to the crossover event since, by construction, each possible
%crossover event results in exactly the same rate, given in
%\eqref{eqn:star_rate}.

%% file: star.tex
\begin{picture}(115,120)
%\linethickness{0.3mm}
\put(0,50){\line(1,0){100}}
\put(50,0){\line(0,1){100}}
\put(50,50){\line(1,1){50}}
\put(50,50){\line(1,-1){50}}
\put(50,50){\line(-1,1){50}}
\put(50,50){\line(-1,-1){50}}
\put(0,50){\circle*{8}}
\put(100,100){\circle*{8}}
\put(100,0){\circle*{8}}
\put(0,100){\circle*{8}}
\put(0,0){\circle*{8}}
\put(50,50){\circle*{8}}
\put(100,50){\circle*{8}}
\put(50,0){\circle*{8}}
\put(50,100){\circle*{8}}
\put(57,65){\makebox (0,0){$x_1$}}
\put(110,60){\makebox (0,0){$x_2$}}
\put(110,10){\makebox (0,0){$x_3$}}
\put(57,10){\makebox (0,0){$x_4$}}
\put(3,10){\makebox (0,0){$x_5$}}
\put(3,60){\makebox (0,0){$x_6$}}
\put(3,110){\makebox (0,0){$x_7$}}
\put(80,57){\makebox (0,0){$Q_a$}}
\put(109,30){\makebox (0,0){$Q_b$}}
\put(57,110){\makebox (0,0){$x_8$}}
\put(110,110){\makebox (0,0){$x_9$}}
\linethickness{0.005mm}
\multiput(100,0)(0,4){13}{\line(0,1){2}}
\end{picture} 
 

%% file: errorexp_v4.tex
\section{Error Exponent for Structure Learning}\label{sec:ee}
The analysis in the previous section characterized the rate
$J_{e,e'}$ for the crossover event $\calC_{e,e'}$ between two
empirical mutual information pairs. In this section, we connect
these set of rate functions $\{ J_{e,e'}\}$ to the quantity of
interest, viz., the error exponent for ML-estimation of edge set $K_P$
in~\eqref{eqn:J}.

Recall that the event $\calC_{e,e'}$ denotes an error in estimating
the order of mutual information quantities. However, such events
$\calC_{e,e'}$ need not necessarily lead to the error event
$\calA_n$ in~\eqref{eqn:err} that the ML-estimate of the edge set
$\hcalE$ is different from the true set $\calE_P$. This is because
the ML-estimate $\hcalE$ is a tree and this  global  constraint
implies that certain crossover events can be ignored. In the
sequel, we will identify useful crossover events through the notion
of a {\em dominant error tree}.

\subsection{Dominant Error Tree}

% be error exponent of the event: ``The tree $\widetilde{\calT}_P\in \calT_{\backslash P}$ is learned from the Chow-Liu algorithm given $n$ samples generated from $P$, a distribution Markov on $T_P $. ''

We can decompose the error event for structure estimation $\calA_n$
in \eqref{eqn:err} into a set of mutually-exclusive events
\begin{equation}
\bP(\calA_n ) = \bP \Bigg(\bigcup_{ T \in \calT^d\setminus \{T_P\}}
\calU_n(T) \Bigg)=\sum_{ T \in \calT^d\setminus \{T_P\}} \bP\left(
\calU_n(T)\right), \label{eqn:union3}
\end{equation}
where each $\calU_n (T)$ denotes the event that the graph of the ML-estimate  $\hTML$ is a tree
$T$ different from the true tree $T_P$. In other words,
\begin{equation}
\calU_n(T)~:=~\left\{\begin{array}{ll}
\left\{\hTML=T\right\},  & \text{ if   } \,\, T \in \calT^d\setminus \{T_P\},\\
\emptyset,  & \text{ if   } \,\, T = T_P.\\
\end{array}\right.
\end{equation}
Note that $\calU_n(T)\cap \calU_n(T')= \emptyset$ whenever   $T\ne T'$. The large-deviation rate or the
exponent for each  error event $\calU_n (T)$ is
\begin{equation}
\Upsilon(T)~\defeq~ \lim_{n\rightarrow\infty} -\frac{1}{n} \log\bP
\left(\calU_n (T) \right),\label{eqn:kt}
\end{equation}
whenever the limit exists.  Among all the error events $\calU_n(T)$,
we identify the dominant one with the slowest rate of decay.
\begin{definition}[Dominant Error Tree]
A {\em dominant error tree} $T_P^*=(\calV, \calE_P^*)$ is a spanning
tree given by\footnote{We will use the notation $\argmin$ extensively in the sequel. It is to be understood that if there is no unique minimum ({\it e.g.\ }in~\eqref{eqn:domtree}), then we arbitrarily choose one of the minimizing solutions.}
\begin{equation}
T_P^*~\defeq~ \argmin_{T \in \calT^d\setminus \{T_P\}} \,\,
\Upsilon(T).\label{eqn:domtree}
\end{equation}
\end{definition}
Roughly speaking, a dominant error tree is the tree that is the
most-likely asymptotic output of the ML-estimator in the event of
an error. Hence, it belongs to the set $\calT^d\setminus \{T_P\}$.
In the following, we note that the error exponent in~\eqref{eqn:J} is equal to
the exponent of the dominant error tree.

\begin{proposition}[Dominant Error Tree \& Error Exponent]\label{prop:domtree}
The error exponent $K_P$ for structure learning is equal to the  exponent
$\Upsilon(T_P^*)$ of the dominant error tree $T_P^*$.\beq
K_P = \Upsilon(T_P^*). \label{eqn:UpsilonTP}\eeq
\end{proposition}

\begin{IEEEproof}
From \eqref{eqn:kt}, we can write
\begin{equation}
\bP  \left(\calU_n (T) \right)\doteq \exp(-n\Upsilon(T)), \quad \forall \, T\in\calT^d\setminus \{T_P\}.
\end{equation}
Now from~\eqref{eqn:union3}, we have
\begin{eqnarray}
\bP(\calA_n ) \doteq
\sum_{ T \in \calT^d\setminus \{T_P\} }  \exp\left(-n
\Upsilon(T)\right)  \doteq     \exp\left(-n   \Upsilon(T_P^*)
\right) ,
\end{eqnarray}
from the ``worst-exponent-wins'' principle~\cite[Ch.\
1]{Den00} and the definition of the dominant error tree $T_P^*$ in~\eqref{eqn:domtree}.
\end{IEEEproof}

\begin{figure}
\centering
\input{replace}
\caption{The path associated to the non-edge $e'=(u,v)\notin\calE_P$, denoted $\Path(e';\calE_P)\subset \calE_P$, is the set of edges along the unique path linking the end points of $e'=(u,v)$. The edge $r(e')=\argmin_{e\in\Path(e';\calE_P) }J_{e,e'}$ is the dominant replacement edge associated to $e'\notin \calE_P$. }
\label{fig:replace}
\end{figure}
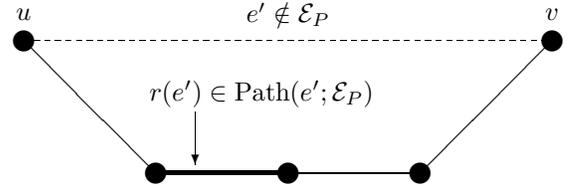

Thus, by identifying a dominant error tree $T_P^*$, we can find the
error exponent $K_P=\Upsilon(T_P^*) $. To this end, we
revisit the crossover events $\calC_{e,e'}$ in \eqref{eqn:Aee},
studied in the previous section. Consider a non-neighbor node pair
$e'$ with respect to $\calE_P$ and the unique path of edges in
$\calE_P$ connecting the two nodes, which we denote as $\Path(e';\calE_P)$. See Fig.\ \ref{fig:replace}, where we define the notion of the path given a non-edge $e'$. Note that $e'$ and $\Path(e';\calE_P)$ necessarily form a cycle; if we replace any  edge $e\in \calE_P$
along the path of the non-neighbor node pair $e'$, the resulting edge set
$\calE_P \setminus \{e\} \cup \{e'\}$ is  still  a spanning tree.
Hence, all such replacements are feasible outputs of the ML-estimation in the event of an error. As a result, all such crossover
events $\calC_{e,e'}$ need to be considered for the error event for
structure learning $\calA_n$ in \eqref{eqn:err}. However, for the
error exponent $K_P$, again by the ``worst-exponent-wins" principle,
we only need to consider the crossover event between each
non-neighbor node pair $e'$ and its  dominant replacement edge $r(e') \in
\calE_P$ defined below.

\begin{definition}[Dominant Replacement Edge]
For each non-neighbor node pair $e' \notin \calE_P$, its {\em  dominant
replacement edge} $r(e')\in \calE_P$ is defined as  the edge in the
unique path along $\calE_P$ connecting the nodes in $e'$ having the
minimum crossover rate
%Given an edge $e'\notin \calE_P$, a {\em replacement edge} $e\in \calE_P$, is any edge along the unique path $\Path(e';\calE_P)\subset \calE_P$ connecting the vertices in $e'$. See Fig.\ \ref{fig:star}(b). The {\em dominant replacement edge} $r(e') \in\Path(e';\calE_P)$ is the replacement edge that satisfies
\begin{equation}
r(e')~\defeq~ \argmin_{e\in \Path(e';\calE_P)} \,\, J_{e,e'},
\label{eqn:dre}
\end{equation}
 %The dominant replacement edge is the edge $r(e')\in \Path(e';\calE_P)$ that results in the smallest rate $ J_{e,e'}$ keeping $e'$ fixed.
where the crossover rate $J_{e,e'}$ is given by \eqref{eqn:Jee2}.
\end{definition}

We are now ready to characterize the error exponent $K_P$ in terms of
the crossover rate between non-neighbor node pairs and their dominant
replacement edges.

\begin{theorem}[Error exponent as a single crossover event]  \label{thm:dom}
The error exponent for ML-tree estimation in~\eqref{eqn:J} is given
by
\begin{equation}
K_P ~=~ J_{r(e^*),e^*}~=~ \min_{e'\notin \calE_P}  \min_{e\in \Path(e';\calE_P)} \,\,
J_{e,e'} ,\label{eqn:Jfinal2}
\end{equation}
where $r(e^*)$ is the dominant replacement edge,  defined in
\eqref{eqn:dre}, associated to  $e^*\notin\calE_P$ and $e^*$ is the optimizing non-neighbor node pair
\begin{equation}
e^* ~\defeq~ \argmin_{e'\notin \calE_P} \,\, J_{r(e')
, e'}. \label{eqn:eprime}
\end{equation}
The dominant error tree $T_P^*=(\calV,\calE_P^*)$ in
\eqref{eqn:domtree} has edge set 
\begin{equation}
\calE_P^* =\calE_P\cup  \{e^*\}
\setminus \{r(e^*) \}. 
\end{equation}
In fact, we also have the following (finite-sample) upper bound on the error probability: 
\begin{equation}
\bP(\calA_n)\le \frac{(d-1)^2(d-2)}{2}\binom{n+1+|\calX|^4}{n+1} \exp(-nK_P), \label{eqn:finite_sam}
\end{equation}
for all $n\in\mathbb{N}$. 
\end{theorem}
\begin{IEEEproof} {\it (Sketch)} The edge set of the dominant error tree $\calE_P^*$ differs from $\calE_P$ in exactly one edge (See Appendix~\ref{prf:dom}). This is because if $\calE_P^*$ were to differ from $\calE_P$ in strictly more than one edge, the resulting error exponent would not be the minimum, hence contradicting Proposition~\ref{prop:domtree}. To identify the dominant error tree, we use the union bound as in~\eqref{eqn:union3} and the ``worst-exponent-wins'' principle~\cite[Ch.\ 1]{Den00}, to conclude that the rate that dominates is the minimum $J_{r(e') , e'}$ over all possible non-neighbor node pairs $e'\notin \calE_P$.  See Appendix~\ref{prf:dom} for the details.
\end{IEEEproof}

The above theorem relates the set of crossover rates $\{J_{e,e'}\}$, which
we characterized in the previous section, to the overall error
exponent $K_P$, defined in~\eqref{eqn:J}. Note that the result in~\eqref{eqn:Jfinal2} and also the existence of the limit in \eqref{eqn:J}  means that the error probability is {\em tight to first order in the exponent} in the sense that  $\bP(\calA_n)\doteq\exp(-nK_P)$. This is in contrast to the work in~\cite{Zuk06}, where bounds on the  upper and lower limit on the sequence $-\frac{1}{n}\log\bP(\calA_n)$   were established.\footnote{However, in \cite{Zuk06}, the authors analyzed the learning of general (non-tree) Bayesian networks.} We numerically compute
the error exponent $K_P$ for different discrete distributions in
Section~\ref{sec:num}.

From~\eqref{eqn:Jfinal2}, we see that if at least one of the
crossover rates $J_{e,e'}$ in the minimization is zero, the overall
error exponent $K_P$ is zero. This observation is important for the
derivation of necessary and sufficient conditions for $K_P$ to be
positive, and hence, for the error probability to decay
exponentially in the number of samples $n$.
%\begin{corollary}
%The error exponent $J$ is given by
%\begin{equation}
%J= J_{r(e^*), e^*}\label{eqn:Jfinal}
%\end{equation}
%where $e^*\notin \calE_P$ is given in~\eqref{eqn:eprime}.
%\end{corollary}
%We remark that this is an exponentially tight bound in the sense
%of~\eqref{eqn:JU} {\it i.e.}\ $J=J_U=J_L$ because the tree that is
%``most-likely'' to be learned and {\em yet} have a different
%structure from $\calE_P$ is identified. This is precisely the
%dominant error tree $T_P^*$.

%\vspace{1em}

\subsection{Conditions for Exponential Decay}
We now provide  necessary and sufficient conditions that ensure
that $K_P$ is strictly positive. This is obviously of crucial
importance since if $K_P >0$, this implies exponential decay of the desired
probability of error $\bP(\calA_n)$, where the error event $\calA_n$ is defined in~\eqref{eqn:err}. 
%\begin{definition}[Proper Forest]
%A  proper fforest on $d$ nodes is an undirected acyclic graph
%that has (strictly) fewer than $d-1$ edges.
%\end{definition}
\begin{theorem}[Equivalent Conditions for Exponential Decay]\label{thm:JzeroI}
Assume that $T_P$, the original structure is acyclic ({\em i.e.}, it may not be connected). Then, the following three statements are equivalent.
\begin{enumerate}
\item[(a)] The probability of error $\bP(\calA_n)$ decays exponentially {\it i.e.},
\begin{equation}
K_P>0.
\end{equation}
\item[(b)] The mutual information quantities satisfy:
\begin{equation}
I(P_{e'}) <  I(P_{e}),\quad \forall \, e\in \Path(e';\calE_P),  \,   e'\notin \calE_P.
\end{equation}
\item[(c)] $T_P$ is not a proper forest.\footnote{A  proper forest on $d$ nodes is an undirected, acyclic graph  that has (strictly) fewer than $d-1$ edges.}
\end{enumerate}
\end{theorem}
%\bprf \Comment short proof here. Can't we make the condition
%simpler? $I(P_{e'})\ne I(P_e)$ where $e$ is part of the path
%connecting nodes in $e'$. This is easier to understand. \eprf
\begin{IEEEproof} ({\it Sketch})
We first show that (a) $\Leftrightarrow$ (b). \\
($\Rightarrow$) We assume statement (a) is true {\it i.e.}, $K_P>0$ and prove that statement (b) is true. Suppose, to the contrary, that $I(P_{e'})= I(P_{e})$ for some $e\in \Path(e';\calE_P)$ and some $e'\notin \calE_P$. Then $J_{r(e'),e'}=0$, where $r(e')$ is the replacement edge associated to $e'$.  By~\eqref{eqn:Jfinal2}, $K_P=0$, which is a contradiction. \\
($\Leftarrow$) We now prove that statement (a) is true assuming statement (b) is true {\it i.e.}, $I(P_{e'})< I(P_{e})$ for all $e\in \Path(e';\calE_P)$ and $e'\notin \calE_P$. By Theorem~\ref{thm:cross_emp},   the crossover rate  $J_{r(e') , e'}$ in~\eqref{eqn:Jee2}  is positive for all $e'\notin\calE_P$. From~\eqref{eqn:Jfinal2}, $K_P>0$ since there are only finitely many $e'$, hence the minimum in \eqref{eqn:eprime} is attained at some non-zero value, {\it i.e.}, $K_P=\min_{e'\notin\calE_P} J_{r(e'),e'}>0.$ 

Statement (c) is equivalent to statement  (b). The proof of this claim makes use of the positivity condition that $P(\bx)>0$ for all $\bx\in \calX^d$ and the fact that if variables $x_1$, $x_2$ and $x_3$ form Markov chains $ x_1 - x_2 - x_3$ and  $x_1 - x_3 - x_2$, then $x_1$ is necessarily  {\em jointly independent} of  $(x_2,x_3)$. Since this proof is rather lengthy, we refer the reader to  Appendix \ref{prf:JzeroI} for the details.
\end{IEEEproof}

Condition (b) states that, for each non-edge $e'$, we need
$I(P_{e'})$ to be strictly smaller  than the mutual information of its
dominant replacement edge $ I(P_{r(e')})$. Condition (c) is a
 more intuitive condition for exponential decay of the
probability of error $\bP(\calA_n)$. This is an important result since it says that for {\em any} non-degenerate tree distribution in which all the pairwise joint distributions are not product distributions ({\it i.e.}, not a proper forest), then we have exponential decay in the error probability. The learning of proper forests is discussed in a companion paper \cite{Tan10jmlr}.

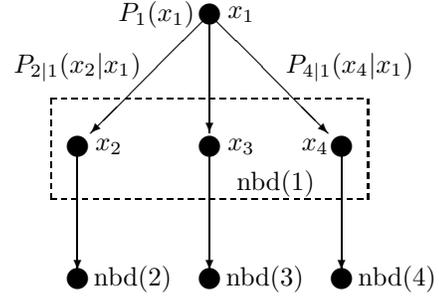
\begin{figure}
\centering
\begin{picture}(120,110)
\put(0,0){\circle*{8}} \put(0,50){\circle*{8}}
\put(50,0){\circle*{8}} \put(50,50){\circle*{8}}
\put(100,0){\circle*{8}} \put(100,50){\circle*{8}}
\put(50,100){\circle*{8}} \put(62,100){\makebox (0,0){$x_1$}}
\put(12,50){\makebox (0,0){$x_2$}}
\put(21,00){\makebox (0,0){$\nbd(2)$}}
\put(71,00){\makebox (0,0){$\nbd(3)$}}
\put(121,00){\makebox (0,0){$\nbd(4)$}} \put(62,50){\makebox
(0,0){$x_3$}} \put(90,50){\makebox (0,0){$x_4$}}
\put(75,36){\makebox (0,0){$\nbd(1)$}} \put(30,100){\makebox
(0,0){$P_{1}(x_1)$}} \put(103,80){\makebox
(0,0){$P_{4|1}(x_4|x_1)$}} \put(0,50){\vector(0,-1){47}}
\put(50,50){\vector(0,-1){47}} \put(100,50){\vector(0,-1){47}}
\put(50,100){\vector(-1,-1){45}} \put(50,100){\vector(0,-1){45}}
\put(50,100){\vector(1,-1){45}}
\multiput(-10,30)(4,0){30}{\line(1,0){2}}
\multiput(-10,68)(4,0){30}{\line(1,0){2}}
\multiput(-10,30)(0,4){10}{\line(0,1){2}}
\multiput(110,30)(0,4){10}{\line(0,1){2}}
 \put(0,80){\makebox (0,0){$P_{2|1}(x_2|x_1)$}}
\end{picture}
\caption{Illustration for Example~\ref{eg:bayesnet}.}
\label{fig:bayesnet}
\end{figure}

In the following example, we describe a simple random process for constructing a distribution $P$ such that all three conditions in Theorem \ref{thm:JzeroI} are satisfied with probability one (w.p.\ 1). See Fig.~\ref{fig:bayesnet}.
\begin{example} \label{eg:bayesnet}
Suppose the structure of $P$, a spanning tree distribution with graph $T_P=(\calV, \calE_P)$, is fixed and $\calX=\{0,1\}$. Now, we assign the parameters of $P$ using the following procedure. Let $x_1$ be the root node. Then randomly draw the parameter of the Bernoulli distribution $P_1(x_1)$  from a uniform distribution on $[0,1]$ {\it i.e.}, $P_1(x_1=0) = \theta_{x_1^0}$ and $\theta_{x_1^0} \sim \mathcal{U}[0,1]$. Next let
 $\nbd(1)$ be the set of neighbors of $x_1$. Regard the set of variables $\{x_j:j\in \nbd(1)\}$ as the children\footnote{Let $x_1$ be the root of the tree. In general, the children of a node $x_k$  ($k\ne 1$) is the set of nodes connected to $x_k$ that are further away from the root than $x_k$.   } of $x_1$. For each $j\in   \nbd(1)$, sample both $P(x_j=0|x_1=0)= \theta_{x_j^0|x_1^0}$ as well as  $P(x_j=0|x_1=1)= \theta_{x_j^0|x_1^1}$   from independent uniform distributions on $[0,1]$ {\it i.e.}, $\theta_{x_j^0|x_1^0}\sim \mathcal{U}[0,1]$ and $\theta_{x_j^0|x_1^1}\sim \mathcal{U}[0,1]$. Repeat this procedure for all children of $x_1$. Then repeat the process for all other children. This construction results in a joint distribution $P(\bx)>0$ for all $\bx\in \calX^d$ w.p.\ 1. In this case, by continuity, all mutual informations are distinct w.p.\ 1, the graph is not a proper forest w.p.\ 1 and the rate $K_P>0$ w.p.\ 1.
\end{example}

This example demonstrates that  $\bP(\calA_n)$ decays exponentially for {\em almost every} tree distribution. More precisely, the tree distributions in which  $\bP(\calA_n)$ does not decay exponentially has  measure zero in $\calP(\calX^d)$.

\subsection{Computational Complexity } \label{sec:computational_complexity}
Finally, we provide an upper bound on the computational complexity to compute $K_P$ in \eqref{eqn:Jfinal2}. Our upper bound on the computational   complexity depends on  the {\em diameter} of the tree $T_P=(\calV,\calE_P)$  which is defined as
\begin{equation}
\diam(T_P) :=\max_{u,v\in\calV} \,\, L(u,v),
\end{equation}
where $L(u,v)$ is the length (number of hops) of the unique path between nodes $u$ and $v$. For example, $L(u,v)=4$ for the non-edge $e'=(u,v)$ in the subtree in Fig.~\ref{fig:replace}.
\begin{theorem}[Computational Complexity for $K_P$]\label{thm:comp}
The number of computations of $J_{e,e'}$ to compute $K_P$, denoted
$N(T_P)$, satisfies
\begin{equation}
N(T_P)~\le~ \frac{1}{2}\diam(T_P) (d-1)(d-2).\label{eqn:comp_old}
\end{equation}
\end{theorem}
\begin{IEEEproof}
Given a non-neighbor node pair $e'\notin\calE_P$, we perform a maximum of $\diam(T_P)$ calculations to determine the dominant replacement edge $r(e')$ from~\eqref{eqn:dre}.  Combining this with the fact that there are a total of $|\binom{\calV}{2} \setminus \calE_P|=\binom{d}{2}-(d-1)=\frac{1}{2} (d-1)(d-2)$ node pairs not in $\calE_P$, we obtain the upper bound.
\end{IEEEproof}

Thus, if the diameter of the tree $\diam(T_P)$ is relatively low and
independent of number of nodes $d$, the complexity is quadratic in
$d$. For instance, for a  star graph, the diameter $\diam(T_P) =2$.
For a balanced tree,\footnote{A balanced tree is one where no leaf is much farther away from the root than any other leaf. The length of the longest direct path between any pair of nodes is $\mathcal{O}(\log d)$.} $\diam(T_P)
=\mathcal{O}(\log d)$, hence the number of computations is
$\mathcal{O}(d^2 \log d)$.

\subsection{Relation of The Maximum-Likelihood Structure Learning Problem to Robust Hypothesis Testing} \label{sec:ht}
\input{uncertainty}

We now take a short detour and discuss the relation between the analysis of the learning problem and {\em robust hypothesis testing}, which was first considered by Huber and Strassen in~\cite{Huber}. Subsequent work was done in~\cite{Pandit, Zeitouni&Gutman:91IT, Unnikrishnan} albeit for differently defined uncertainty classes known as moment classes. %The robust hypothesis testing setup  is different from the learning problem in this paper. 

We hereby consider an alternative but related problem. Let $T_1,\ldots, T_M$ be the $M=d^{d-2}$ trees with $d$ nodes. Also let $\calQ_1,\ldots,\calQ_M\subset\calD(\calX^d,\calT^d)$ be the subsets of tree-structured graphical models Markov on $T_1,\ldots, T_M$ respectively.  The structure learning problem is similar to   the $M$-ary  hypothesis testing problem between the uncertainty classes of distributions $\calQ_1,\ldots, \calQ_M$. The uncertainty class $\calQ_i$ denotes the set of tree-structured graphical models with different {\em parameters} (marginal $\{P_i:i\in\calV\}$ and pairwise distributions $\{P_{i,j}:(i,j)\in\calE_P\}$) but Markov on the same tree $T_i$. 

In addition, we note that  the  probability simplex $\calP(\calX^d)$ can be partitioned into $M$   subsets\footnote{From the definition in \eqref{eqn:Bi}, we see that the relative interior of the subsets are pairwise disjoint. We discuss the scenario when $P$ lies on the boundaries of these subsets  in Section~\ref{sec:nonunique}. } $\calB_1,\ldots,\calB_M\subset\calP(\calX^d)$ where each $\calB_i, i=1,\ldots, M$ is defined as 
\begin{equation} \label{eqn:Bi}
\calB_i:= \bigcup_{P'\in\calQ_i} \left\{ Q :D(P'\,||\,Q )\le \min_{R\in\cup_{j\ne i}\calQ_i} D( P' \, ||\,R)\right\}.
\end{equation}
See Fig.~\ref{fig:uncertainty}.  According to the ML criterion in~\eqref{eqn:clopt2}, if  the type $\hP$ belongs to $\calB_i$, then the $i$-th tree is favored.

In~\cite{Tan10:ISIT}, a subset of the authors of this paper considered the Neyman-Pearson setup of a  robust   binary hypothesis testing problem where the  null hypothesis corresponds to the true tree model $P$  and the (composite) alternative hypothesis corresponds to the set of distributions Markov on some  erroneous tree $T_Q\ne T_P$. The false-alarm probability  was constrained to be smaller than  $\alpha>0$ and optimized for worst-case type-II (missed detection) error exponent using the Chernoff-Stein Lemma \cite[Ch.\ 12]{Cov06}. It was established that  the worst-case error exponent can be expressed in closed-form in terms of the mutual information of so-called {\em bottleneck edges}, {\it i.e.}, the edge and non-edge pair that have the smallest mutual information difference. However, in general, for the binary hypothesis testing problem, the error event {\em   does  not} decompose into a union of  local events. This is in contrast to error exponent for learning the ML tree $K_P$, which can be computed by considering {\em local crossover events} $\calC_{e,e'}$, defined in \eqref{eqn:Aee}.

%For more discussion on how the error exponent for the learning problem relates to the worst-case type-II error exponent for the  hypothesis testing problem, the reader is referred to \cite[Section IV.D]{Tan10:ISIT}. 

Note that $\{\hP\in\calB_i\}$ corresponds to a {\em global event} since each $\calB_i\subset\calP(\calX^d)$. The large-deviation analysis techniques we utilized to obtain the error exponent $K_P$ in Theorem~\ref{thm:dom} show that such global  error events can be also decomposed into a collection of local crossover events $\calC_{e,e'}$. These local events depend only on the type {\em restricted} to pairs of nodes $e$ and $e'$ and are more intuitive for assessing (and analyzing) when and how an error can occur during the Chow-Liu learning process. % allows us to analyze the error probability using large deviation techniques applied to the smaller state space $\calX^4$.  

%However, the rest of the paper is devoted  to the study of the ML structure learning error exponent $K_P$, defined in \eqref{eqn:J}, instead of the error exponent associated to robust  hypothesis testing. 

%\cite{Unnikrishnan}

%In general, when we consider an arbitrary distribution 
%$\widetilde{P}\in \calP(\calX^d)$ which need not be a tree 
%distribution , the Chow-Liu algorithm can be adapted to  solve the 
%following divergence-minimization problem:
%\begin{equation}
%\min_{Q\in\calD(\calX^d,\calT^d)} \,\, D(\widetilde{P}  \, ||\, Q) 
%=\sum_{\bx\in \calX^d}\widetilde{P}(\bx) \log 
%\frac{\widetilde{P}(\bx)}{Q(\bx)} . \label{eqn:clopt3}
%\end{equation}
%For this reason, $\hPML$ is also known as the {\em reverse 
%information projection}~\cite{Csi84,Csis03} of $\hP$ onto the set of 
%trees $\calD(\calX^d,\calT^d)$. 

%% file: replace.tex
\begin{picture}(200,70)
\put(0,50){\line(1,-1){50}}
\put(150,00){\line(1,1){50}}
\put(50,00){\line(1,0){100}}
\linethickness{0.5mm}
\put(50,00){\line(1,0){50}}
\put(0,50){\circle*{8}}
\put(50,0){\circle*{8}}
\put(100,0){\circle*{8}}
\put(150,0){\circle*{8}}
\put(200,50){\circle*{8}}
\put(100,60){\makebox (0,0){$e'\notin\calE_P$}}
\put(0,60){\makebox (0,0){$u$}}
\put(200,60){\makebox (0,0){$v$}}
\put(90,30){\makebox (0,0){$r(e')\in\Path(e';\calE_P)$}}
\linethickness{0.005mm}
\multiput(0,50)(4,0){50}{\line(1,0){2}}
\put(65,23){\vector(0,-1){20}}
\end{picture} 

%% file: uncertainty.tex
\begin{figure}
\centering
\begin{picture}(100,100)
\thicklines
\put(0,0){\line(1,2){50}}
\put(50,100){\line(1,-2){50}}
\put(0,0){\line(1,0){100}}
\put(50,0){\line(0,1){33}}
\put(50,33){\line(2,1){27}}
 \put(50,33){\line(-2,1){27}}
 \put(50,68){\makebox (0,0){$\calB_1$}}
  \put(25,30){\makebox (0,0){$\calB_2$}}
    \put(75,30){\makebox (0,0){$\calB_3$}}
        \put(85,75){\makebox (0,0){$\calP(\calX^d)$}}
\put(70,10){\circle*{2}}        
        \put(77,10){\makebox (0,0){$\hP$}}
\end{picture}
\caption{The partitions of the simplex associated to our learning problem are given by $\calB_i$, defined in \eqref{eqn:Bi}. In this  example, the type $\hP$ belongs to $\calB_3$ so the tree associated to partition $\calB_3$ is favored. }
\label{fig:uncertainty}
\end{figure}
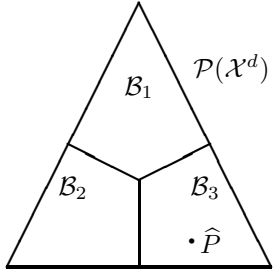

%% file: euc_v3.tex
\section{Euclidean Approximations}\label{sec:euc}

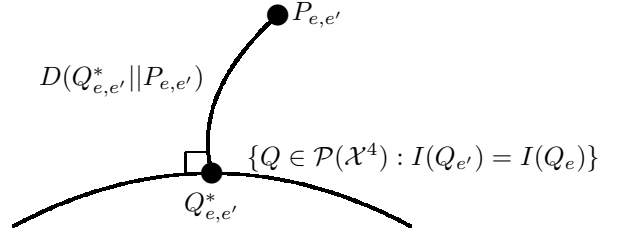
\begin{figure}
\centering
\begin{picture}(200,80)
%\color{blue}
\put(100,80){\circle*{8}}
%\color{red}
\put(75,20){\circle*{8}}
%\color{black}
\put(65,20){\line(0,1){8}}
\put(65,28){\line(1,0){8}}
\linethickness{.3mm}
\qbezier(0,0)(75,40)(150,0)
\qbezier(100,80)(67,50)(75,20)
\put(115,80){\makebox (0,0){$P_{e,e'}$}}
\put(75,7){\makebox (0,0){$Q_{e,e'}^*$}}
\put(155,26){\makebox (0,0){$\{Q\in\calP(\calX^4):I(Q_{e'}) =  I(Q_{e})\}$}}
%\put(125,20){\vector(-1,-1){15}}
\put(41,55){\makebox (0,0){$D(Q_{e,e'}^*||P_{e,e'})$}}
\end{picture}
\caption{A geometric interpretation of~\eqref{eqn:Jee2} where $P_{e,e'}$ is projected onto the submanifold of probability distributions $\{Q\in\calP(\calX^4):I(Q_{e'}) =  I(Q_{e})\}$. }
\label{fig:proj}
\end{figure}
In order to gain more insight into the error exponent, we make use
of {\em Euclidean approximations}~\cite{Bor08} of
information-theoretic quantities to obtain an approximate but
closed-form solution to~\eqref{eqn:Jee2}, which is non-convex and
hard to solve exactly.  In addition, we note that  the dominant error event results from an edge and a non-edge that satisfy the conditions for which the Euclidean approximation is valid, {\it i.e.}, the very-noisy condition given later in Definition~\ref{def:vn}. This justifies our approach we adopt in this section.  Our use of Euclidean approximations for various information-theoretic quantities  is akin to various problems considered in other contexts in information theory~\cite{Bor06,Bor08,Abb08}.
%\begin{itemize}
%\item The very noisy hypothesis testing problem~\cite{Bor06}.
%\item The very noisy source coding with helper problem~\cite{Bor08}.
%\item The very noisy degraded broadcast channel problem~\cite{Bor08}.
%\item The very noisy linear universal decoding for compound channels problem~\cite{Abb08}.
%\end{itemize}

We first approximate the crossover rate $J_{e,e'}$ for
any two node pairs $e$ and $e'$, which do not share a common node. The joint distribution on $e$ and $e'$, namely $P_{e,e'}$ belongs to the set $\calP(\calX^4)$. Intuitively, the crossover rate $J_{e,e'}$ should depend on the ``separation'' of the mutual information values
$I(P_e)$ and $I(P_{e'})$, and also on the uncertainty of the difference between mutual
information estimates $I(\hP_e)$ and $I(\hP_{e'})$. We will see that
the approximate rate also depends on these mutual information quantities given by a simple  expression which can be
 regarded as the signal-to-noise ratio
(SNR) for learning.

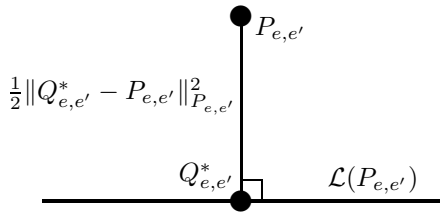
\begin{figure}
\centering
\begin{picture}(150,80)
%\color{blue}
\put(75,70){\circle*{8}}
%\color{red}
\put(75,0){\circle*{8}}
%\color{black}
\put(83,0){\line(0,1){8}}
\put(83,8){\line(-1,0){8}}
\linethickness{.3mm}
\put(0,0){\line(1,0){150}}
\put(75,0){\line(0,1){70}}
\put(90,65){\makebox (0,0){$P_{e,e'}$}}
\put(62,9){\makebox (0,0){$Q_{e,e'}^*$}}
%\put(145,25){\makebox (0,0){$\epsilon=Q_{e,e'}^*-P_{e,e'} $}}
%\put(155,10){\makebox (0,0){$\epsilon^T (s_{e} - s_{e'})  = I(P_{e'})-I(P_{e})  $}}
\put(125,8){\makebox (0,0){$\mathcal{L}(P_{e,e'})$}}
%\put(113,13){\vector(0,-1){20}}
\put(30,40){\makebox (0,0){$ \frac{1}{2}\|Q_{e,e'}^*-P_{e,e'}\|_{P_{e,e'}}^2$}}
\end{picture}
\caption{Convexifying the objective results in a least-squares problem. The objective is converted into a quadratic as in~\eqref{eqn:defJeetilde} and the linearized constraint set $\mathcal{L}(P_{e,e'})$ is given~\eqref{eqn:linconstr}. }
\label{fig:proj2}
\end{figure}

Roughly speaking, our strategy is to ``convexify'' the objective and
the constraints in~\eqref{eqn:Jee2}. See Figs.~\ref{fig:proj} and \ref{fig:proj2}. To do so, we recall that if $P$ and $Q$ are two discrete distributions with the same support $\calY$, and
they are close entry-wise, the KL divergence can be
approximated~\cite{Bor08} as
\begin{align}
D(Q\, &||\, P)= - \sum_{a\in\calY} Q(a)\log \frac{P(a)}{Q(a)},\\
&= - \sum_{a\in\calY} Q(a)\log \left[1+ \left(\frac{P(a)-Q(a)}{Q(a)} \right)\right], \\
%&\stackrel{\approx}{\le}   \sum_{\bx\in \calX^d} Q(\bx) \left[ \frac{P(\bx)-Q(\bx)}{Q(\bx)} - \left(\frac{Q(\bx)-P(\bx)}{Q(\bx)}\right)^2 \right],  \label{eqn:log_ineq}\\
&=\frac{1}{2} \sum_{a\in\calY} \frac{(Q(a)-P(a))^2}{Q(a)} +o( \|Q-P\|_{\infty}^2 )  ,  \label{eqn:ineq}\\
&= \frac{1}{2}\| Q - P \|_Q^2  +o( \|Q-P\|_{\infty}^2 ) , \label{eqn:dqp_approx1}
\end{align}
where $\|y\|_w^2$ denotes the weighted squared norm of $y$, {\it i.e.}, $ \|y\|_w^2 ~\defeq~ \sum_i y_i^2/w_i$. The equality  in~\eqref{eqn:ineq} holds because $\log(1+t)= \sum_{i=1}^{\infty}(-1)^{i+1} t^i/i$ for $t\in (-1,1]$. The difference between the divergence and the Euclidean approximation becomes tight as $\epsilon=\|P-Q\|_{\infty}\to 0$. Moreover, it remains tight even if the subscript $Q$ in \eqref{eqn:dqp_approx1} is changed to a distribution $Q'$ in the vicinity of $Q$~\cite{Bor08}. That is, the difference between $\|Q-P\|_Q$ and $\|Q-P\|_{Q'}$ is  negligible compared to either term when $Q'\approx  Q$. Using this fact and the assumption that  $P$ and $Q$ are two discrete distributions that   are close entry-wise,
\begin{equation}
D(Q\,||\, P)~\approx~   \frac{1}{2}\| Q - P \|_P^2. \label{eqn:dqp_approx}
\end{equation}
In fact, it is also known~\cite{Bor08} that if $\|P-Q\|_{\infty}< \epsilon$ for some $\epsilon>0$, we also have $D(P\, ||\, Q) \approx D(Q\, ||\, P)$.

In the following, to make our statements  precise, we will use the notation $\alpha_1 \approx_{\delta} \alpha_2$ to denote that two real numbers $\alpha_1$ and $\alpha_2$ are in the  $\delta$ neighborhood of each other, {\it i.e.}, $|\alpha_1-\alpha_2|<\delta$.\footnote{In the following, we will also have continuity statements  where given $\epsilon>0$ and $\alpha_1 \approx_{\epsilon} \alpha_2$, implies that there exists some  $\delta=\delta(\epsilon)>0$ such that $\beta_1 \approx_{\delta } \beta_2$. We will be casual about specifying what the $\delta$'s are. } We will also need the following notion of information density to
state our approximation for $J_{e,e'}$.
\begin{definition}[Information Density]
Given a pairwise joint distribution $P_{i,j}$ on $\calX^2$  with marginals
$P_i$ and $P_j$, the {\em information density}~\cite{Pin64, Lan06}
function, denoted by $s_{i,j}:\calX^2\rightarrow \bR$, is defined as
\begin{equation}
s_{i,j}(x_i, x_j) ~\defeq~ \log
\frac{P_{i,j}(x_i,x_j)}{P_i(x_i)P_j(x_j)},\quad \forall \, (x_i,
x_j)\in \calX^2 .  \label{eqn:info_dens}
\end{equation}
%We will also use the notation $s_e=s_{i,j}$ if $e=(i,j)$.
\end{definition}
Hence, for each node pair $e=(i,j)$, the information density $s_e$ is also a random variable whose
expectation is simply the mutual information between $x_i$ and $x_j$, {\it i.e.}, $\bE[s_{e}]=I(P_e)$.

Recall that we also assumed in Section~\ref{sec:prelim} that $T_P$ is a spanning tree,  which implies that for all node pairs $ (i,j) $, $P_{i,j}$ is {\em not} a product distribution, {\it i.e.},
$P_{i,j} \ne P_i  P_j$, because if it were, then $T_P$ would be disconnected. We now define a condition for which our approximation  holds.
\begin{definition}[$\epsilon$-Very Noisy Condition] \label{def:vn}
We say that  $P_{e,e'}\in\calP(\calX^4)$, the joint distribution on node pairs $e$ and $e'$, satisfies the {\em $\epsilon$-very noisy condition} if
\begin{equation}
\|P_e-P_{e'}\|_{\infty}\! := \!\max_{(x_i,x_j)\in \calX^2 } | P_e(x_i,x_j) -P_{e'}(x_i,x_j) |\!< \!\epsilon.\label{eqn:eps_vn}
\end{equation}
\end{definition}
This condition is needed because if \eqref{eqn:eps_vn} holds, then by continuity of the mutual information, there exists a $\delta>0$ such that $I(P_e)\approx_{\delta} I(P_{e'})$, which means that the mutual information quantities are difficult to distinguish and the approximation in \eqref{eqn:dqp_approx1} is accurate.\footnote{Here and in the following, we do not specify the exact value of $\delta$ but we simply note that as $\epsilon\to 0$, the approximation in \eqref{eqn:dqp_approx} becomes tighter. } Note that proximity of the mutual information values is not sufficient for the approximation to hold since we have seen from Theorem~\ref{thm:cross_emp}  that $J_{e,e'}$ depends not only on the mutual information quantities but on the entire joint distribution $P_{e,e'}$.
%\Comment No longer needed: [A1:] For all node pairs $ (i,j) \in \calV^2$, every element of $P_{i,j}$ is strictly positive {\it i.e.}\ $\forall \,(x_i,x_j)\in \calX^2, P_{i,j}(x_i,x_j) >0 $. %{\it i.e.}\ $P_e\in \calP_{\delta}(\calX^2) =\{Q\in \calP(\calX^2): Q(x_i,x_j)\ge\delta >0, \,\,\forall (x_i,x_j)\in \calX^2\}$. $\delta$ can be arbitrarily small but non-zero.
%|Q^*(x_i,x_j,x_k,x_l)- P_{e,e'}(x_i,x_j,x_k,x_l)|< \epsilonilon

We now define the {\em approximate crossover rate} on disjoint node pairs $e$ and $e'$ as
\begin{equation}
\widetilde{J}_{e,e'}~ \defeq~ \inf 
\left\{ \frac{1}{2}\| Q - P_{e,e'} \|_{P_{e,e'}}^2  : Q\in \mathcal{L}(P_{e,e'}) \right\},  \label{eqn:defJeetilde}
\end{equation}
where the (linearized) constraint set  is
\begin{align}
%\mathcal{L}(P_{e,e'}) \!:=\!\left\{Q:(Q  -P_{e,e'})^T  (s_e-s_{e'}) = I(P_{e'})-I(P_e)\right\}.
\mathcal{L}(P_{e,e'}) := \Big\{  &Q\in\calP(\calX^4):  I(P_e)+\left\langle\nabla_{P_e}I(P_e), Q-P_{e,e'}\right\rangle  \nn\\
& \, =  \, I(P_{e'})+\left\langle \nabla_{P_{e'}}I(P_{e'}), Q-P_{e,e'}\right\rangle \Big\}, \label{eqn:linconstr}
\end{align}
where $\nabla_{P_e}I(P_e)$ is the gradient vector of the mutual information with respect to the joint distribution $P_e$. We also define the approximate error exponent as
\begin{equation}
\widetilde{K}_P ~\defeq~ \min_{e'\notin \calE_P} \, \min_{e\in \Path(e';
\calE_P)} \, \widetilde{J}_{e,e'}. \label{eqn:Ktilde}
\end{equation}
We now provide the expression for the approximate crossover rate $\widetilde{J}_{e,e'}$ and also state the conditions under which the
approximation is asymptotically accurate in $\epsilon$.\footnote{We say that a collection of approximations $\{\widetilde{\theta}(\epsilon):\epsilon>0\}$ of a true parameter $\theta$ is {\em asymptotically accurate in $\epsilon$} (or simply asymptotically accurate) if the approximations  converge to $\theta$ as $\epsilon\to 0$, {\it i.e.}, $\lim_{\epsilon\to 0} \widetilde{\theta}(\epsilon)=\theta$.}

\begin{theorem}[Euclidean approximation of $J_{e,e'}$]\label{thm:euc}
The approximate crossover rate for the empirical mutual information quantities, defined in \eqref{eqn:defJeetilde}, is given by
\begin{equation}
\widetilde{J}_{e,e'}~=~\frac{(\bE[s_{e'}-s_{e}])^2}{2\,
\var(s_{e'}-s_{e})}~=~\frac{(I(P_{e'})-I(P_{e}))^2}{2\,
\var(s_{e'}-s_{e}) } , \label{eqn:Jee3}
\end{equation}
where $s_e$ is the information density  defined
in~\eqref{eqn:info_dens} and the expectation and variance are both
with respect to $P_{e,e'}$.
%Furthemore, if $P_{e,e'}$ satisfies the $\epsilon$-very noisy condition for some $\epsilon>0$, then there exists a $\delta>0$ such that
%\begin{equation}
%J_{e,e'} ~\approx_{\delta} ~ \widetilde{J}_{e,e'},   \label{eqn:Jeeapprox}
%\end{equation}
Furthermore, the approximation~\eqref{eqn:Jee3} is asymptotically accurate, {\it i.e.}, as $\epsilon\to 0$ (in the definition of $\epsilon$-very noisy condition), we have that $\widetilde{J}_{e,e'}\to J_{e,e'}$.

%Furthermore, the approximation is tight when
%$Q^*_{e,e'}$, the optimizing distribution in~\eqref{eqn:Jee2} is
%close to $P_{e,e'}$ entry-wise   {\it i.e.}, $\|P_{e,e'} -
%Q^*_{e,e'}\|_{\infty}<\epsilon'$, then

%Furthermore, if all the optimizing distributions $Q^*_{e,e'}$, are close to the corresponding $P_{e,e'}$ entry-wise, then the approximate error exponent is also close to the true error exponent
%$$
%K_P~\approx~\tilK_P.
%$$
\end{theorem}
\begin{IEEEproof} ({\it Sketch})
Eqs.\ \eqref{eqn:defJeetilde} and \eqref{eqn:linconstr} together define a least squares problem. Upon simiplification of the solution, we obtain~\eqref{eqn:Jee3}. See Appendix~\ref{prf:euc} for the details.
\end{IEEEproof}

We also have an additional result for the Euclidean approximation for the overall error exponent $K_P$. The proof is clear from the definition of $\tilK_P$ in \eqref{eqn:Ktilde} and the continuity of the min function. 
\begin{corollary}[Euclidean approximation of $K_P$] \label{cor:euc}
The approximate error exponent $\widetilde{K}_P$ is asymptotically accurate if  all joint distributions  in the set $\{P_{e,e'}:e\in \Path(e;\calE_P), e'\notin\calE_P\}$  satisfy the $\epsilon$-very noisy condition. 
%when either one of the following conditions is true.
%%, there exists a $\delta>0$ such that
%%\begin{equation}
%%K_P ~\approx_{\delta} ~\tilK_P.\label{eqn:K_Papprox}
%%\end{equation}
%\begin{enumerate}
%\item[(a)] The joint distribution $P_{r(e'),e'}$ satisfies the $\epsilon$-very noisy condition for every $e'\notin \calE_P$.
%\item[(b)] The joint distribution $P_{r(e^*),e^*}$, where $e^*$ is defined in \eqref{eqn:eprime} and $r(e^*)$ is the dominant replacement edge associated to non-edge $e^*$,  satisfies the $\epsilon$-very noisy condition  but all the other distributions on the non-neighbor node pairs $e'\notin \calE_P\cup \{e^*\}$ along with their dominant replacement edges $r(e')$ {\em do not} satisfy the  $\epsilon$-very noisy condition. %, then~\eqref{eqn:K_Papprox} holds for some $\delta>0$.
%\end{enumerate}
%The approximation \eqref{eqn:K_Papprox}
\end{corollary}
%\begin{IEEEproof}
%Since $\tilK_P =\tilJ_{r(e^*),e^*}$ (by the definition in \eqref{eqn:Ktilde}) and $\tilJ_{r(e^*),e^*}$ is asymptotically accurate (by Theorem \ref{thm:euc}) so is $\tilK_P$. 
%\end{IEEEproof}
%\bprf \Comment check the above statement. the optimizing $e^*$ may
%be different for approximate and exact ones.
% See Appendix..\eprf

Hence, the expressions for the crossover rate $J_{e,e'}$ and the error exponent  $K_P$
are vastly simplified  under the $\epsilon$-very noisy condition on the joint distributions $P_{e,e'}$.  The approximate
crossover rate $\widetilde{J}_{e,e'}$ in~\eqref{eqn:Jee3} has a very
intuitive meaning. It is proportional to the square of the
difference between the mutual information quantities of $P_e$ and
$P_{e'}$. This corresponds exactly to our initial intuition -- that
if $I(P_e)$ and $I(P_{e'})$ are well separated $(I(P_e)\gg
I(P_{e'}))$ then the crossover rate has to  be large.
$\widetilde{J}_{e,e'}$ is also weighted by the precision (inverse
variance) of $(s_{e'}- s_e)$. If this variance   is large then
 we are   uncertain about the estimate
$I(\hP_e)-I(\hP_{e'})$, and crossovers are more likely, thereby
reducing the  crossover rate $\widetilde{J}_{e,e'}$.

We now comment on our assumption of $ P_{e,e'}$ satisfying the $\epsilon$-very noisy condition,
under which the approximation is tight as seen in Theorem~\ref{thm:euc}. When $ P_{e,e'}$ is  $\epsilon$-very noisy, then we have $I(P_e)\approx_{\delta} I(P_{e'})$, which implies that the optimal solution of \eqref{eqn:Jee2} $Q^*_{e,e'}\approx_{\delta'} P_{e,e'}$. When $e$ is an edge and $e'$ is a non-neighbor node
pair, this implies that it is very hard to distinguish the relative magnitudes of the empiricals $I(\hP_e)$ and
$I(\hP_{e'})$. Hence, the particular problem of learning the
distribution $P_{e,e'}$ from samples is {\em very noisy}. Under these conditions, the approximation in~\eqref{eqn:Jee3} is accurate.

In summary, our approximation in \eqref{eqn:Jee3} takes into account
not only the absolute difference between the mutual information
quantities $I(P_e)$ and $I(P_{e'})$, but also the uncertainty in
learning them. The expression in~\eqref{eqn:Jee3} is, in fact, the
SNR for the estimation of the difference between empirical mutual
information quantities.   This answers one of the fundamental questions we posed
in the introduction, viz., that we are now able to distinguish
between distributions that are ``easy'' to learn and those that are
``difficult'' by computing the set of SNR quantities
$\{\widetilde{J}_{e,e'}\}$ in~\eqref{eqn:Jee3}.

%% file: nonunique_v3.tex
\section{Extensions to Non-Tree Distributions}\label{sec:nonunique}
\begin{figure}
\input{reverse_info_proj}
\end{figure}
In all the preceding sections, we dealt exclusively with the case 
where the true distribution $P$ is Markov on a  tree. In this 
section,  we  extend the preceding large-deviation analysis to deal with 
 distributions $P$ that may not be   tree-structured  but in which we estimate a tree distribution from the given set of samples $\bx^n$, using the Chow-Liu ML-estimation procedure. Since the Chow-Liu procedure outputs a tree, it is not possible to learn the structure of $P$ correctly. Hence, it will be necessary to redefine the error event. 

%\Comment My idea is to write a journal for discrete part and include 
%this section as well. Hence, make the following subsection 
%completely for discrete distributions. 
%\footnote{We use $\min$ in the optimization problem inn \eqref{eqn:projection1} becaues the minimum is indeed attained. Clearly the KL-divergence is continuous. The set of tree distributions $\calD(\calX^d,\calT^d)$ can be shown to be compact and hence by Weierstrauss' extreme value theorem, the minimum is attained.  }

When $P$ is not a tree distribution, we analyze 
the properties of the optimal {\em reverse 
I-projection}~\cite{Csis03} of  $P$  onto the set of tree 
distributions, given by the optimization  problem\footnote{The minimum in the optimization problem in \eqref{eqn:projection1} is attained because the KL-divergence is continuous and the set of tree distributions $\calD(\calX^d,\calT^d)$ is compact.}
\begin{equation}  \Pi^*(P) ~\defeq~ \min_{Q\in \calD(\calX^d,\calT^d)} 
\,\, D(P\,||\,Q). \label{eqn:projection1}
\end{equation}
$ \Pi^*(P)$ is the KL-divergence of $P$ to the closest element in $\calD(\calX^d,\calT^d)$. See Fig.~\ref{fig:reverse}. As Chow and Wagner~\cite{Cho73} noted, if $P$ is not a tree, there 
may be several trees  optimizing \eqref{eqn:projection1}.\footnote{This is a technical condition of theoretical interest in  this section. In fact, it can be shown that the set of distributions such that there is more than one tree optimizing \eqref{eqn:projection1} has (Lebesgue) measure zero in $\calP(\calX^d)$.} We denote 
the set of optimal projections as $\calP^*(P)$, given by
\begin{equation}
\calP^*(P) := \{ Q\in \calD(\calX^d,\calT^d): D(P\,||\,Q)= 
\Pi^*(P)\}.\label{eqn:opt_projection}
\end{equation}

%\Comment We cannot write $\argmin$ as a single element. That is the 
%whole point that there may be many optimal projections.
%
%\Comment Suggest moving the example to the Appendix. 

We now illustrate that $\calP^*(P)$ may have more than one element with 
the following example.
\begin{table}[t]
\centering
%\begin{tabular}{cc}
\begin{tabular}{|c|c|c||c|}\hline
$x_1$& $x_2$&$x_3$ & Distribution $P(\bx)$ \\\hline 
0 & 0 &0 & $(1/2-\xi)(1/2-\kappa)$\\\hline
0 & 0 &1 & $(1/2+\xi)(1/2-\kappa)$\\\hline
0 & 1 &0 & $(1/3+\xi)\kappa$\\\hline
0 & 1 &1 & $(2/3-\xi)\kappa$\\\hline
1 & 0 &0 & $(2/3-\xi)\kappa$\\\hline
1 & 0 &1 & $(1/3+\xi)\kappa$\\\hline
1 & 1 &0 & $(1/2-\xi)(1/2-\kappa)$\\\hline
1 & 1 &1 & $(1/2+\xi)(1/2-\kappa)$\\\hline
\end{tabular}
%&
%\begin{picture}(60,60)
%\put(0,0){\line(1,0){60}}
%\put(0,0){\line(1,2){30}}
%\put(30,60){\line(1,-2){30}}
%\put(0,0){\circle*{8}}
%\put(30,60){\circle*{8}}
%\put(60,0){\circle*{8}}
%\put(0,10){\makebox (0,0){$x_2$}}
%\put(60,10){\makebox (0,0){$x_3$}}
%\put(40,60){\makebox (0,0){$x_1$}}
%\end{picture}
%\end{tabular}
\caption{Table of probability values for Example~\ref{eg:degen}. }
\label{tab:degen}
\end{table}

\begin{example}\label{eg:degen}
Consider the parameterized discrete probability distribution $P\in\calP(\{0,1\}^3)$  shown in Table~\ref{tab:degen} where $\xi\in(0,1/3)$ and $\kappa\in(0,1/2)$ are constants.

\begin{proposition}[Non-uniqueness of projection] \label{prop:nonunique}
For sufficiently small $\kappa$, the Chow-Liu MWST algorithm 
(using either Kruskal's~\cite{Kruskal} or Prim's~\cite{Prim} 
procedure) will first include the edge $(1,2)$. Then, it will 
arbitrarily choose between the two remaining edges  $(2,3)$ or 
$(1,3)$.
\end{proposition}
The proof of this proposition is provided in Appendix~\ref{prf:nonunique} where we show that $I(P_{1,2})> I(P_{2,3}) = I(P_{1,3})$ for sufficiently small $\kappa$. Thus, the optimal tree structure $P^*$ is not unique. This in fact corresponds to the case where $P$ belongs to the boundary of some set $\calB_i \subset\calP(\calX^d)$ defined in \eqref{eqn:Bi}.  See Fig.~\ref{fig:eflat} for an information geometric 
interpretation. 
\end{example}

\begin{figure}
\input{equiv_trees2}
\end{figure}

Every tree distribution in $\calP^*(P)$ has the maximum sum 
mutual information weight. More precisely,  we have 
\begin{equation} 
\sum_{e \in \calE_Q} I(Q_e)\! =\! \max_{Q' \in \calD(\calX^d,\calT^d)} \sum_{e \in 
\calE_{Q'}} I(Q'_e),\,\, \forall\,Q \in \calP^*(P).\label{eqn:II}
\end{equation} 
Given \eqref{eqn:II}, we note that  when we use a MWST algorithm to find the optimal 
solution to the problem in~\eqref{eqn:projection1},    ties will  be encountered during the greedy 
addition of edges, as demonstrated in Example~\ref{eg:degen}. Upon 
breaking the ties arbitrarily, we obtain some distribution $Q\in 
\calP^*(P)$. We now provide a sequence of useful definitions that lead to definition of a new error event for which we can perform large-deviation analysis. 

%$\calP^*(P)$ not good.. should be a function of P.

We denote the set of tree structures\footnote{In fact, each tree 
defines a so-called  {\em e-flat submanifold}~\cite{Ama00,Ama01} in 
the set of probability distributions on $\calX^d$ and $\estP$ lies 
in both submanifolds. The so-called {\em m-geodesic} connects $P$ to 
any of its optimal projection $\estP\in\calP^*(P)$. } corresponding to the distributions in 
$\calP^*(P)$ as 
\begin{equation} {\calT}_{\calP^*(P)} ~:=~  
\{T_Q\in \calT^d:  Q\in \calP^*(P) \},\label{eqn:ST_P}
\end{equation} and term it as the set of {\em optimal tree 
projections}. A similar definition applies to the edge sets of 
optimal tree projections
\begin{equation} \calE_{\calP^*(P)}~ 
:=~  \{\calE_Q :T_Q=(\calV,\calE_Q)\in\calT^d, Q \in \calP^*(P) \}.\label{eqn:SE_P}
\end{equation}

%The set ${\calT}_{\calP^*(P)}\subseteq\calT^d$ is finite 
%because $\calT^d$ is a finite set. 

Since the distribution $P$ is unknown, our goal is to estimate the 
optimal tree-projection $\estP$ using the empirical distribution $\hP$, where $\estP$ is given by
\begin{equation}
\estP~\defeq~ \argmin_{Q\in \calD(\calX^d,\calT^d)} \,\, 
D(\hP\,||\,Q).\label{eqn:Pest}
\end{equation} If there are many distributions $Q$, we 
arbitrarily pick one of them. We will see that by redefining the error event, we will have still a LDP.   Finding the reverse I-projection 
$\estP$   can be solved efficiently (in time $\mathcal{O}(d^2\log 
d)$) using the  Chow-Liu algorithm~\cite{CL68} as described in 
Section~\ref{sec:CL}.   

We define $T_{\estP} =(\calV, \calE_{\estP})$ as the graph of $\estP$, which is the learned tree and redefine the new {\em error event} as
\begin{equation}
\calA_n(\calP^*(P)) ~:=~ \left\{\calE_{\estP} ~\notin~ 
\calE_{\calP^*(P)}    \right\} . \label{eqn:err1}
\end{equation}
Note that this new error event essentially reduces to the original 
error event  $\calA_n=\calA_n(\{P\})$ in~\eqref{eqn:err} if 
${\calT}_{\calP^*(P)}$ contains only one member. So if the learned structure belongs to $\calE_{\calP^*(P)} $, there is no error, otherwise an error is declared. We would like 
to analyze the decay of the error probability of  
$\calA_n(\calP^*(P)) $ as defined in~\eqref{eqn:err1}, {\it i.e.}, find the new {\em error exponent}
\begin{equation}
K_{\calP^*(P)} ~:=~ \lim_{n\rightarrow \infty} -\frac{1}{n} \log \bP(\calA_n(\calP^*(P))). \label{eqn:J1}
\end{equation}
It turns out that the analysis of the new event $\calA_n(\calP^*(P))$ is very similar to the analysis performed in Section~\ref{sec:ee}. We  redefine the notion of a dominant replacement edge and the computation of the new rate $K_{\calP^*(P)}$ then follows automatically. 
\begin{definition}[Dominant Replacement Edge]
Fix an edge set $\calE_Q\in \calE_{\calP^*(P)}$. For the error event $\calA_n(\calP^*(P))$ defined in~\eqref{eqn:err1}, given a non-neighbor node pair $e' \notin \calE_Q $, its dominant replacement edge $r(e';\calE_Q)$ with respect to $\calE_Q$, is given by
\begin{equation}
r(e';\calE_Q) ~:=~ \argmin_{\substack{e\in \Path(e';\calE_Q)\\  
\calE_Q\cup\{e'\}\setminus\{e\}\notin \calE_{\calP^*(P)}}} 
\,\, J_{e,e'}, \label{eqn:dre1}
\end{equation}
if there exists an edge $e\in\Path(e';\calE_Q)$ such that 
$\calE_Q\cup\{e'\}\setminus\{e\}\notin \calE_{\calP^*(P)}$. 
Otherwise $r(e';\calE_Q)=\emptyset$. $J_{e,e'}$ is the  crossover rate of 
mutual information quantities defined in \eqref{eqn:Jee}. If $r(e';\calE_Q)$
exists, the corresponding crossover rate is 
\begin{equation}
J_{r(e';\calE_Q),e'}~=~\min_{\substack{e\in \Path(e';\calE_Q)\\  
\calE_Q\cup\{e'\}\setminus\{e\}\notin \calE_{\calP^*(P)}}} 
\,\, J_{e,e'}, \label{eqn:corr_cr}
\end{equation}
otherwise $J_{\emptyset,e'}=+\infty$.
\end{definition}
In \eqref{eqn:dre1}, we are basically fixing an edge set $\calE_Q\in \calE_{\calP^*(P)}$ and excluding  the trees with  $e\in\Path(e';\calE_Q)$ replaced by $e'$ if it belongs to the set of 
optimal tree projections ${\calT}_{\calP^*(P)}$. We further remark 
that in~\eqref{eqn:dre1}, $r(e')$ may not necessarily exist. Indeed, 
this occurs if every tree with $e\in \Path(e';\calE_Q)$ replaced by 
$e'$ belongs to the set of optimal tree projections. This is, however,  {\em not}
an error by the definition of the error event in \eqref{eqn:err1} 
hence, we set $J_{\emptyset,e'}=+\infty$. In addition, we define the {\em dominant non-edge} associated to edge set  $ \calE_Q\in \calE_{\calP^*(P)} $ as:
\begin{equation}
e^*(\calE_Q) ~:=~ \argmin_{e'\notin\calE_Q} \min_{\substack{e\in \Path(e';\calE_Q)\\  
\calE_Q\cup\{e'\}\setminus\{e\}\notin \calE_{\calP^*(P)}}} 
  J_{e,e'}. \label{eqn:dom_non_edge} %\forall\, \calE_Q\in \calE_{\calP^*(P)}  
\end{equation}
Also, the {\em dominant structure} in the set of optimal tree projections is defined as
\begin{equation}
\calE_{P^*} ~:=~ \argmin_{\calE_Q\in \calE_{\calP^*(P)}} \quad J_{r(e^*(\calE_Q); \calE_Q), e^*(\calE_Q)}, \label{eqn:EQstar}
\end{equation}
where the crossover rate $J_{r(e';\calE_Q),e'}$ is defined in \eqref{eqn:corr_cr} and the dominant non-edge $e^*(\calE_Q)$ associated to $\calE_Q$ is defined in \eqref{eqn:dom_non_edge}. Equipped with these definitions, we are now ready to state  the generalization of Theorem~\ref{thm:dom}. 
\begin{theorem}[Dominant Error Tree] \label{thm:dom_err_tree}
For the error event $\calA_n(\calP^*(P))$ defined in~\eqref{eqn:err1}, a dominant error tree (which may not be unique) has edge set  given by 
\begin{equation}
\calE_{P^*}\cup\{e^*(\calE_{P^*})\}\setminus \{ r(e^*(\calE_{P^*}); \calE_{P^*})\}, \label{eqn:E_Pstar}
\end{equation}
where $e^*(\calE_{P^*})$ is the dominant non-edge associated to the dominant structure $\calE_{P^*}\in\calE_{\calP^*(P)}$ and  is defined by \eqref{eqn:dom_non_edge} and \eqref{eqn:EQstar}. Furthermore, the error exponent $K_{\calP^*(P)}$, defined in \eqref{eqn:J1} is given as
\begin{equation}
K_{\calP^*(P)} ~=~ \min_{\calE_Q\in \calE_{\calP^*(P)}}\min_{e'\notin\calE_Q}  \min_{\substack{ e\in 
\Path(e';\calE_Q)\\  \calE_Q\cup\{e'\}\setminus\{e\}\notin 
\calE_{\calP^*(P)}}} J_{e,e'} .\label{eqn:Jfinal1}
\end{equation}
\end{theorem}
\begin{IEEEproof}
The proof of this theorem follows directly  by identifying the dominant error tree belonging to the set $\calT^d\setminus\calT_{\calP^*(P)} $. By further applying the result in Proposition~\ref{prop:domtree} and Theorem~\ref{thm:dom}, we obtain the result via the ``worst-exponent-wins''~\cite[Ch.\ 1]{Den00} principle by minimizing over all trees in the set of optimal projections $\calE_{\calP^*(P)}$ in \eqref{eqn:Jfinal1}.
\end{IEEEproof}
This theorem now allows us to analyze the more general error event $\calA_n(\calP^*(P))$, which includes $\calA_n$ in \eqref{eqn:err} as a special case if the set of optimal tree projections ${\calT}_{\calP^*(P)}$ in~\eqref{eqn:ST_P} is a singleton.

%% file: reverse_info_proj.tex
\centering
\begin{picture}(130, 100)
\qbezier(-20,0)(40,25)(100,0)
%\put(-20,0){\line(1,0){120}}
\put(-20,0){\line(3,5){30}}
\put(100,0){\line(3,5){30}}
%\put(10,50){\line(1,0){120}}
\qbezier(10,50)(40,75)(130,50)
\put(80,90){\circle*{8}}
\put(50,40){\circle*{8}}
\put(78,21){\makebox (0,0){$\calD(\calX^d,\calT^d)$}}
\linethickness{.2mm}
\qbezier(80,90)(39,70)(50,40)
\put(42,80){\makebox (0,0){$\Pi^*(P)$}}
\put(90,90){\makebox (0,0){$P$}}
\put(38,38){\makebox (0,0){$P^*$}}
\end{picture}
\caption{Reverse I-projection~\cite{Csis03} of $P$ onto the set of tree distributions $\calD(\calX^d,\calT^d)$ given by \eqref{eqn:projection1}.}
\label{fig:reverse}

%% file: equiv_trees2.tex
\centering
\begin{picture}(140, 80)
\put(0,80){\line(1,-1){80}}
\put(50,0){\line(1,1){80}}
\put(65,60){\circle*{8}}
\put(40,40){\circle*{8}}
\put(90,40){\circle*{8}}
\put(157,70){\makebox (0,0){$\calD(\calX^d,T_{\estP^{(2)}})$}}
\put(-25,70){\makebox (0,0){$\calD(\calX^d,T_{\estP^{(1)}})$}}
\put(40,26){\makebox (0,0){$\estP^{(1)}$}}
\put(90,26){\makebox (0,0){$\estP^{(2)}$}}
\put(130,48){\makebox (0,0){$D(P||\estP^{(2)})$}}
\put(0,48){\makebox (0,0){$D(P||\estP^{(1)})$}}
\put(25,48){\vector(1,0){15}}
\put(105,48){\vector(-1,0){15}}
\put(65,70){\makebox (0,0){$P$}}
\put(116,5){\makebox (0,0){$e$-flat manifolds}}
\linethickness{.2mm}
\qbezier(65,60)(48,58)(40,40)
\qbezier(65,60)(82,58)(90,40)
\end{picture}
\caption{Each tree  defines an $e$-flat submanifold~\cite{Ama00,Ama01} of probability distributions. These are the two lines as shown in the figure. If the KL-divergences $D(P || \estP^{(1)})$ and $D(P || \estP^{(2)})$ are equal, then $\estP^{(1)}$ and $\estP^{(2)}$ do not have the same structure but both are optimal with respect to the optimization problem in \eqref{eqn:projection1}. An example of such a distribution $P$ is provided in Example~\ref{eg:degen}. }
\label{fig:eflat}

%% file: numerical_v2.tex
\section{Numerical Experiments}\label{sec:num}

\begin{figure}
\centering
\begin{picture}(100,100)
\put(0,0){\line(1,1){50}}
\put(100,0){\line(-1,1){50}}
\put(50,50){\line(0,1){50}}
\put(0,0){\circle*{8}}
\put(50,50){\circle*{8}}	
\put(100,0){\circle*{8}}
\put(50,100){\circle*{8}}
\put(0,11){\makebox (0,0){$x_4$}}
\put(62,50){\makebox (0,0){$x_1$}}
\put(100,11){\makebox (0,0){$x_3$}}
\put(62,100){\makebox (0,0){$x_2$}}
\put(76,75){\makebox (0,0){$e=(1,2)$}}
\put(51,7){\makebox (0,0){$e'=(3,4)$}}
\multiput(0,0)(4,0){25}{\line(1,0){2}}
\end{picture}
\caption{Graphical model used for our numerical experiments. The true model is a symmetric star (cf. Section~\ref{sec:star}) in which the mutual information quantities satisfy $I(P_{1,2})= I(P_{1,3})=I(P_{1,4})$ and by construction, $I(P_{e'})< I(P_{1,2})$ for any non-edge $e'$. Besides, the mutual information quantities on the non-edges are  equal, for example, $I(P_{2,3})=I(P_{3,4}).$ }
\label{fig:star_expt}
\end{figure}
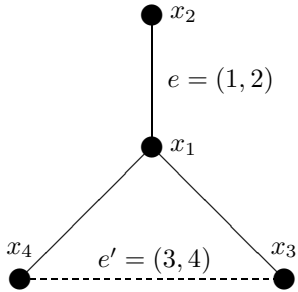

In this section, we perform a series of numerical experiments with the following three objectives:
\begin{enumerate}
\item In Section~\ref{sec:goodness}, we study the accuracy of the Euclidean approximations  (Theorem~\ref{thm:euc}). We do this  by analyzing under which regimes   the approximate  crossover rate $\tilJ_{e,e'}$ in \eqref{eqn:Jee3} is  close to the true crossover rate $J_{e,e'}$ in \eqref{eqn:Jee2}.
\item Since the LDP and error exponent analysis are asymptotic theories, in Section~\ref{sec:sims} we use simulations to study the behavior of  the actual crossover rate, given a finite number of samples $n$. In particular, we study how fast the  crossover rate, obtained from simulations,  converges to the true crossover rate.  To do so, we generate a number of samples from the true distribution and use the Chow-Liu algorithm to learn trees structures. Then we compare the result to the true structure and finally compute the error probability.
\item In Section~\ref{sec:emprate}, we address the issue of  the learner  not having access to the true distribution, but nonetheless wanting to compute an estimate of the crossover rate. The learner only has the samples $\bx^n$ or equivalently, the empirical distribution $\hP$. However, in all the preceding analysis, to compute the true crossover rate $J_{e,e'}$ and the overall error exponent $K_P$, we used the true distribution $P$ and solved the constrained  optimization problem in~\eqref{eqn:Jee2}. Alternatively we computed the approximation in~\eqref{eqn:Jee3}, which is also a function of the true distribution. However, in practice, it is also useful to compute an online estimate of the crossover rate by using the empirical distribution in place of the true distribution in the constrained optimization problem in~\eqref{eqn:Jee2}. This is an estimate of the rate that the learner can compute given the samples. We call this the {\em empirical rate} and formally define it in Section~\ref{sec:emprate}. We  perform  convergence analysis  of the empirical rate  and also numerically verify  the rate of convergence to the true crossover rate.
% We would also like to compare the true crossover rate to the rate obtained by using the empirical distribution in the non-convex optimization problem \eqref{eqn:Jee2} to compute the crossover rate $J_{e,e'}$.
\end{enumerate}

In the following, we will be performing numerical experiments for the undirected graphical  model with four nodes as shown in Fig.\ \ref{fig:star_expt}. We parameterize the distribution with $d=4$ variables with a single parameter $\gamma> 0$ and let $\calX=\{0,1\}$, {\it i.e.}, all the variables are binary. For the parameters, we set $P_1(x_1=0) = 1/3$ and
\begin{subequations}\label{eqn:delta_par}
\begin{align}
P_{i|1}(x_i=0|x_1=0) &= \frac{1}{2}+\gamma , \quad i=2,3,4, \\
P_{i|1}(x_i=0|x_1=1) &= \frac{1}{2}-\gamma,\quad i=2,3,4. 
\end{align}
\end{subequations}
With this parameterization, we see that if $\gamma$ is small, the mutual information $I(P_{1,i})$ for $i=2,3,4$ is also small. In fact if $\gamma=0$, $x_1$ is independent of $x_i$ for $i=2,3,4$ and as a result, $I(P_{1,i})=0$. Conversely, if $\gamma$ is large, the  mutual information $I(P_{1,i})$  increases as the dependence of the outer nodes with the central node increases. Thus, we can vary the size of the mutual information along the edges by varying $\gamma$. By symmetry, there is only one crossover rate and hence this crossover rate is also the error exponent for the error event $\calA_n$ in~\eqref{eqn:err}. This is exactly the same as the symmetric star graph as described in Section~\ref{sec:star}.

\subsection{Accuracy of Euclidean Approximations}\label{sec:goodness}
We first study the accuracy of the Euclidean approximations used to derive the result in Theorem~\ref{thm:euc}. We denote the {\em true rate} as the crossover rate resulting from the non-convex optimization problem~\eqref{eqn:Jee2} and the {\em approximate rate} as the crossover rate computed using the approximation in~\eqref{eqn:Jee3}.

\begin{figure}
\centering
\includegraphics[width=.8\columnwidth]{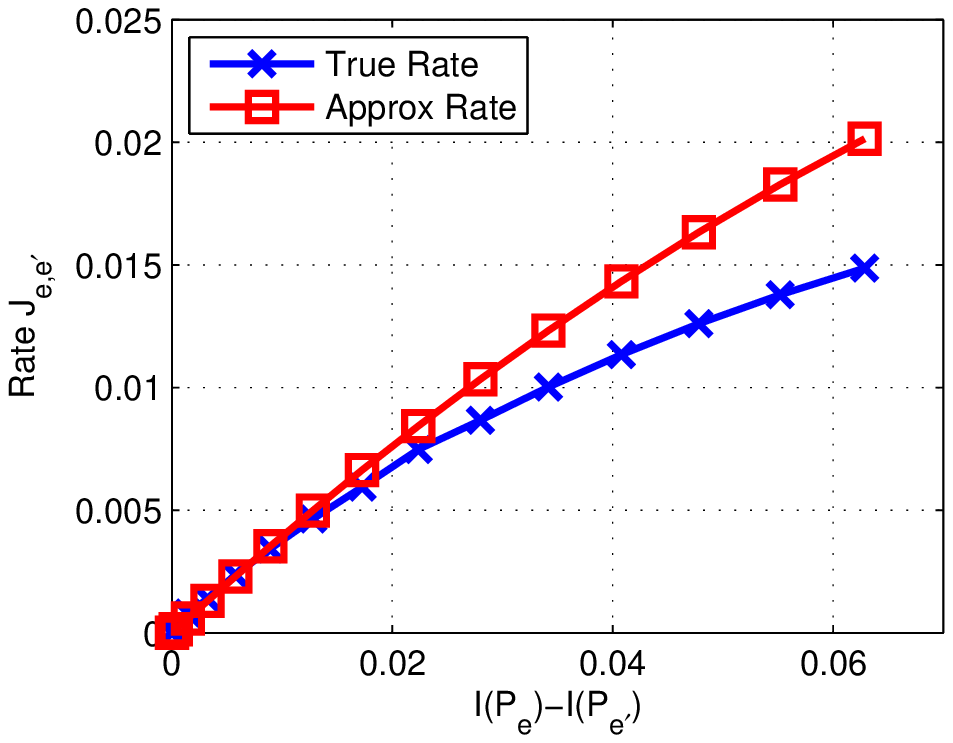}
\caption{Comparison of True and Approximate Rates.}
\label{fig:discrete_euc}
\end{figure}

We vary $\gamma$ from 0 to 0.2 and plot both the true and approximate rates against the difference between the mutual informations $I(P_e)-I(P_{e'})$ in Fig.\ \ref{fig:discrete_euc}, where $e$ denotes any edge and $e'$ denotes any non-edge in the model. The non-convex optimization problem was performed using the Matlab function \texttt{fmincon} in the optimization toolbox. We used several different feasible starting points and chose the best optimal objective value to avoid problems with  local minima. We first note from Fig.\ \ref{fig:discrete_euc} that both rates increase as $I(P_e)-I(P_{e'})$ increases. This is in line with our intuition because if $P_{e,e'}$ is such that $I(P_e)-I(P_{e'})$ is large, the crossover rate is also large. We also observe that if $I(P_e)-I(P_{e'})$ is small, the true and approximate rates are very close. This is in line  with the assumptions for Theorem~\ref{thm:euc}. Recall that if $P_{e,e'}$ satisfies the $\epsilon$-very noisy condition (for some small $\epsilon$), then the mutual information quantities $I(P_e)$ and $I(P_{e'})$ are close  and consequently the true and approximate crossover rates are also close. When the difference between the mutual informations increases, the true and approximate rate separate from each other.

\subsection{Comparison of True Crossover Rate to the Rate obtained from Simulations}\label{sec:sims}
In this section, we compare the true crossover rate in \eqref{eqn:Jee2} to the rate we obtain when we learn tree structures using Chow-Liu with i.i.d.\ samples drawn from $P$, which we define as the {\em simulated rate}.  We fixed $\gamma>0$ in~\eqref{eqn:delta_par} then for each $n$, we estimated the probability of error using the Chow-Liu algorithm as described in Section~\ref{sec:CL}. We state the procedure precisely in the following steps.

\begin{enumerate}
\item Fix $n \in \mathbb{N}$ and sample $n$ i.i.d.\ observations $\bx^n $ from $P$.
\item Compute the empirical distribution $\hP$ and the set of empirical mutual information quantities $\{I(\hP_e): e\in \binom{\calV}{2}\}$.
\item Learn the Chow-Liu tree $\calE_{{\text{\tiny ML}}}$ using a MWST algorithm with $\{I(\hP_e): e\in \binom{\calV}{2}\}$ as the edge weights.
\item If $\calE_{{\text{\tiny ML}}}$ is not equal to $\calE_P$, then we declare an error.
\item Repeat steps 1 -- 4 a total of $M\in \mathbb{N}$ times and estimate the probability of error $\bP(\calA_n) = \# \mathrm{errors}/M$ and the error exponent $-(1/n) \log \bP(\calA_n) $, which is the simulated rate.
\end{enumerate}
If the probability of error  $\bP(\calA_n)$ is very small, then the number of runs $M$ to estimate  $\bP(\calA_n)$ has to be fairly large. This is often the case in error exponent analysis as the sample size needs to be substantial to estimate very small error probabilities.

\begin{figure}
\centering
\begin{tabular}{c}
\includegraphics[width=.75\columnwidth]{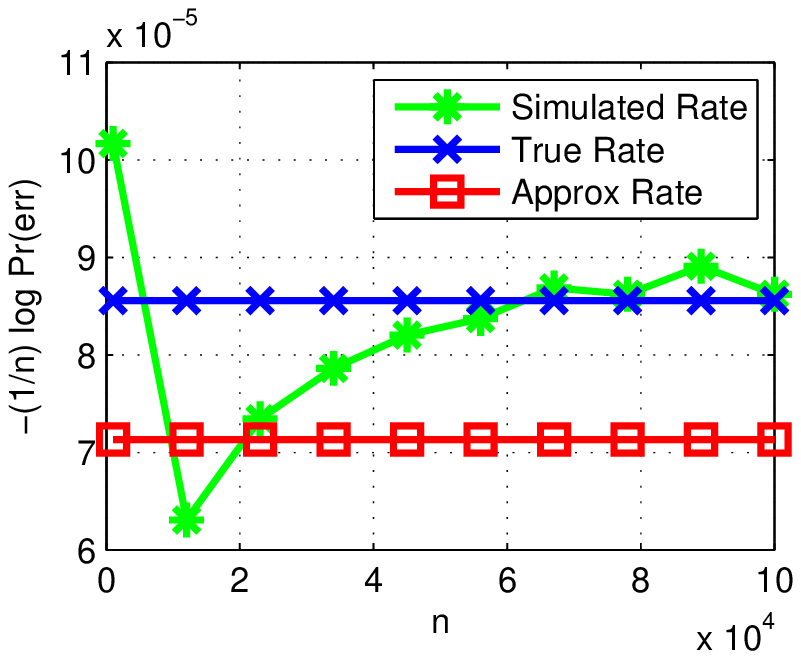}\\
\includegraphics[width=.75\columnwidth]{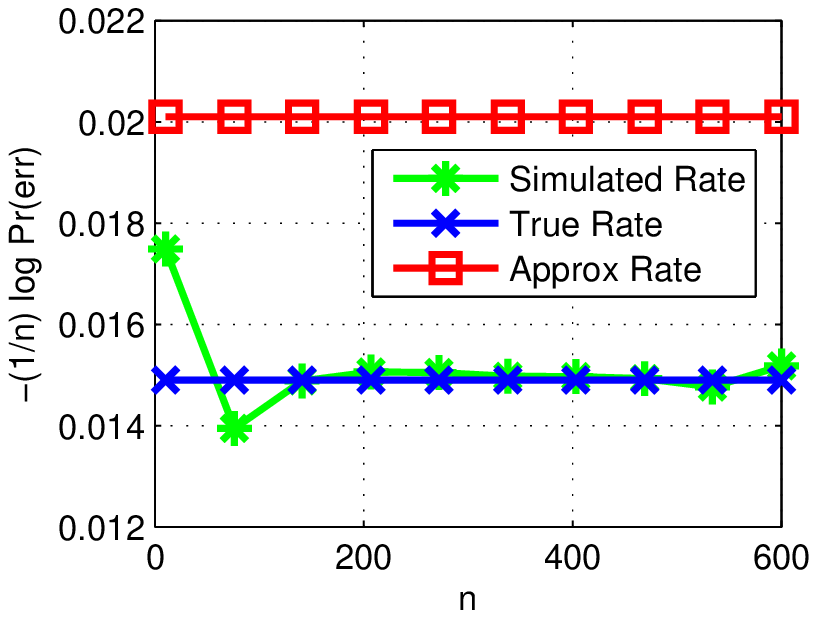}
\end{tabular}
\caption{Comparison of True, Approximate and Simulated Rates  with $\gamma=0.01$ (top) and $\gamma=0.2$ (bottom). Here the number of runs $M= 10^{7}$ for $\gamma=0.01$ and $M= 5\times  10^{8}$ for $\gamma=0.2$. The probability of error is computed dividing the total number of errors by the total number of runs.}
\label{fig:sim_disc}
\end{figure}

In Fig.\ \ref{fig:sim_disc}, we plot the true rate, the approximate rate and the simulated rate when $\gamma=0.01$ (and  $M= 10^{7}$) and $\gamma=0.2$ (and  $M=5\times 10^{8}$). Note that, in the former case, the true rate is higher than the approximate rate and in the latter case, the reverse is true. When $\gamma$ is large ($\gamma=0.2$), there are large differences in the true tree models. Thus, we expect that the error probabilities to be very small and hence $M$ has to be large in order to estimate the error probability correctly but $n$  does not have to be too large for the simulated rate to converge to the true rate. On the other hand, when $\gamma$ is small ($\gamma=0.01$), there are only subtle differences in the graphical models, hence we need a larger number of samples $n$ for the simulated rate to converge to its true value, but $M$ does not have to be large since the error probabilities are not small. The above observations are in line with our intuition.

%Interestingly, we need a large number of samples $n$ in order for the simulated rate to converge to the true value. For instance, for the last data point of the top plot of Fig.\ \ref{fig:sim_disc}, $n= 10^5$. Effectively, this means that we require a total of $M\times n= 10^{12}$ samples to estimate the probability of error and hence the rate accurately. Now, for $\gamma=0.2$, since the difference between the mutual informations is large, we need fewer samples $n$ to estimate the rate.
%
%The  experiment shows that, in general, we require a very large number of samples for the simulated rate to be close to the true rate. This is due to the fact that the theory of large-deviations is an asymptotic theory, which means that we may require $n$ to be very large for the simulated rate to be close to the true rate. In the case of learning trees, this section has shown empirically that this is indeed the case.

\subsection{Comparison of True Crossover Rate to  Rate obtained from the Empirical Distribution} \label{sec:emprate}
In this subsection, we compare the true rate to the {\em empirical rate}, which is defined as
\begin{equation}
\hJ_{e,e'} :=  \inf_{Q\in \calP(\calX^4) }\left\{D(Q\, ||\, \hP_{e,e'}) : I(Q_{e'})=  I(Q_e)\right\}.
\label{eqn:Jeehat}
\end{equation}
The empirical rate $\hJ_{e,e'}=\hJ_{e,e'}(\hP_{e,e'})$ is a function of the empirical distribution $\hP_{e,e'}$. This rate is computable by a learner, who does not have access to the true distribution $P$. The learner only has access to a finite number of samples $\bx^n=\{\bx_1, \ldots, \bx_n\}$. Given $\bx^n$, the learner can compute the empirical probability $\hP_{e,e'}$ and perform the  optimization in~\eqref{eqn:Jeehat}. This is an estimate of the true crossover rate. A natural question to ask is the following: Does the empirical rate $\hJ_{e,e'}$ converge to the true crossover rate $J_{e,e'}$ as $n\to\infty$? The next theorem answers this question in the affirmative.
\begin{theorem}[Crossover Rate Consistency]\label{thm:consistency}
The empirical crossover  rate $\hJ_{e,e'}$ in \eqref{eqn:Jeehat} converges almost surely to the true crossover rate $J_{e,e'}$ in \eqref{eqn:Jee2}, {\it i.e.},
\begin{equation}
\bP\left(\lim_{n\to \infty}   \hJ_{e,e'}=J_{e,e'}  \right)=1.
\end{equation}
\end{theorem}
\begin{IEEEproof} ({\it Sketch})
The proof of this theorem follows from the continuity of $\hJ_{e,e'}$ in the empirical distribution $\hP_{e,e'}$ and the continuous mapping theorem by Mann and Wald~\cite{Man43}. See Appendix~\ref{prf:consistency} for the details.
\end{IEEEproof}
We conclude that the learning of the rate from samples is consistent. Now we perform simulations to determine how many samples are required for the empirical rate to converge to the true rate.

\begin{figure}
\centering
\begin{tabular}{c}
\includegraphics[width=.75\columnwidth]{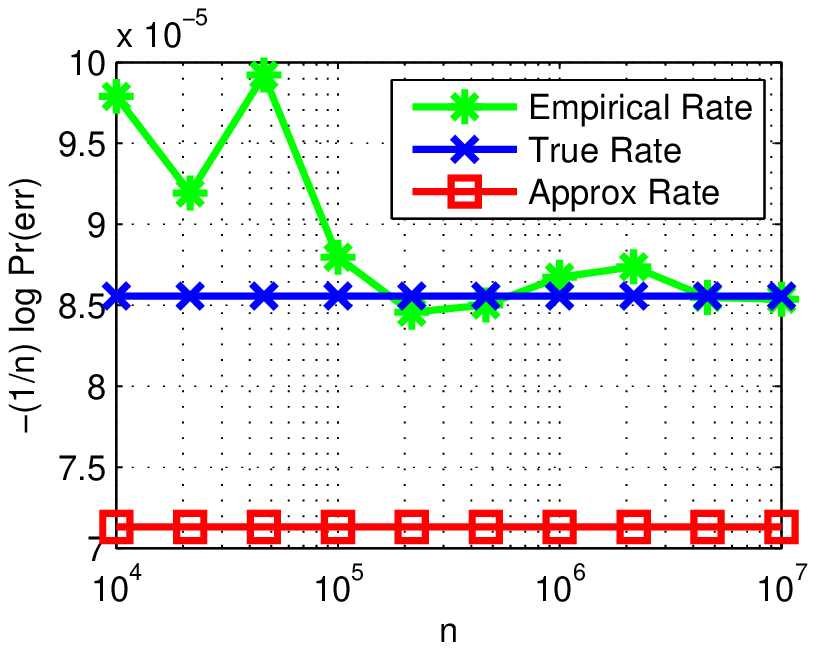}\\
\includegraphics[width=.75\columnwidth]{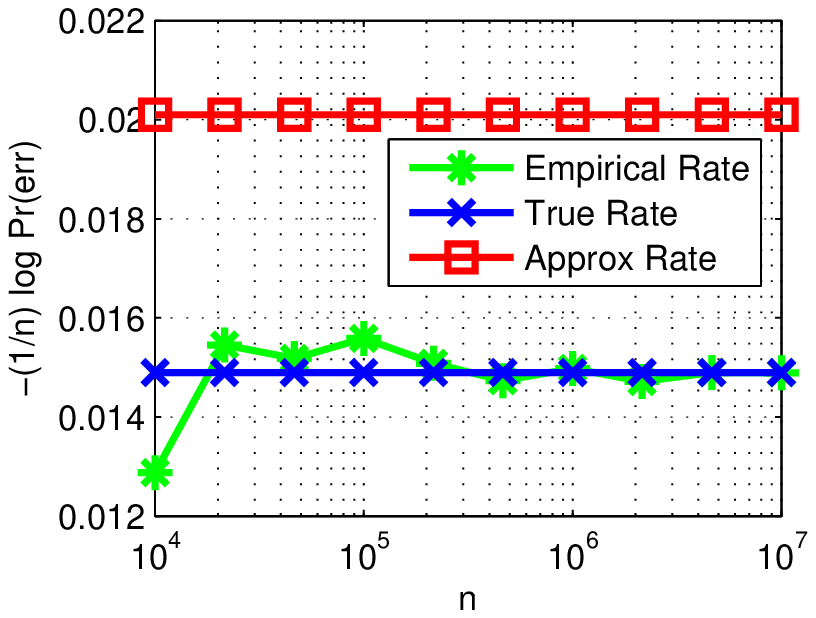}
\end{tabular}
\caption{Comparison of True, Approximate and Empirical Rates   with $\gamma=0.01$ (top) and $\gamma=0.2$ (bottom). Here $n$ is the number of observations used to estimate the empirical distribution.}
\label{fig:emp_disc}
\end{figure}

We set $\gamma=0.01$ and $\gamma=0.2$ in~\eqref{eqn:delta_par}. We then drew $n$ i.i.d.\ samples from $P$ and computed the empirical distribution $\hP_{e,e'}$. Next, we solved the optimization problem in~\eqref{eqn:Jeehat} using the \texttt{fmincon} function in Matlab, using different initializations and compared the empirical rate to the true rate. We repeated this for several values of $n$ and the results are displayed in Fig.\ \ref{fig:emp_disc}. We see that  for $\gamma=0.01$, approximately $n= 8\times 10^6$ samples are required for the empirical distribution to be close enough to the true distribution so that the empirical rate converges to the true rate.

%Next, we consider the Gaussian case with $\gamma=0.01$ and $\gamma=0.2$ in~\eqref{eqn:gamma}. Once again, we compared the true rate to the empirical rate in Fig.\ \ref{fig:emp_gauss}. In this case, we observe that for $\gamma=0.01$, approximately $n= 5\times 10^6$ samples are required for the empirical distribution to be close enough to the true distribution so that the empirical rate converges to the true rate.

%We conclude from Fig.\ \ref{fig:emp_disc} that typically, we would require fewer samples to estimate the true rate using the empirical rate in \eqref{eqn:Jeehat} as compared to the simulated rate. This is because to compute the simulated rate, multiple runs (in our notation $M$ runs) are required in order to estimate the probability of error since each set of $n$ samples allows us to learn only one tree, and hence to make one comparison to the true tree model.

%% file: conclusion_v2.tex
\section{Conclusion, Extensions and Open Problems}\label{sec:concl}

In this paper, we presented a solution to the problem of 
finding the error exponent for tree structure learning by  
extensively using tools from large-deviations theory combined with 
facts about tree graphs.   We quantified the error exponent for learning 
the structure and exploited the structure of the true tree to identify the 
dominant tree in the set of erroneous trees. We also drew insights 
from the approximate crossover rate, which can be interpreted as the 
SNR for learning.  These two main results in Theorems~\ref{thm:dom} 
and~\ref{thm:euc} provide the intuition as to how errors occur for 
learning discrete tree distributions via the Chow-Liu algorithm. 

In a companion paper \cite{Tan10sp}, we develop counterparts to the results here for the Gaussian case. Many of the results carry through 
but thanks to the special structure that Gaussian distributions 
possess, we are also able to identify which structures (among the class of trees) are easier to 
learn and which are harder to learn given a fixed set of  correlation 
coefficients on the edges. Using Euclidean information theory, we show that if the parameters on the edges are fixed, the star is the most difficult to learn (requiring many more samples to ensure $\bP(\calA_n)\le \delta$) while the Markov chain is the easiest. The results in this paper have also been extended to  learning high-dimensional   general acyclic models (forests)~\cite{Tan10jmlr}, where $d$ grows with $n$ and typically the growth of $d$ is much faster than that of $n$.

There are many open problems resulting from this paper. One of these involves studying  the optimality of the error exponent associated to the ML Chow-Liu algorithm $K_P$, {\it i.e.}, whether the rate established in Theorem~\ref{thm:dom} is the best (largest) among all consistent estimators of the edge set. Also, since large-deviation rates may not be indicative of the true error probability $\bP(\calA_n)$, results from weak convergence theory \cite{Billingsley} may potentially be applicable to provide better approximations to $\bP(\calA_n)$.  %A preliminary study was conducted in a recent paper~\cite{Tan10:ISIT}.%We defer the study of this to a another paper. 

%We defer the discussion of this set of results to a 
%companion paper.  It is our belief that this work can be extended to  
%analyzing performance of more general algorithms for learning graph 
%structures.

%% file: appendix_v3.tex
\section{Proof of Theorem \ref{thm:cross_emp}} \label{prf:cross_emp}
\begin{IEEEproof}
We divide the proof of this theorem into three steps. Steps 1 and 2  prove the expression in \eqref{eqn:Jee2}. Step 3 proves the existence of the optimizer.

{\it Step 1}: First, we note from Sanov's Theorem~\cite[Ch.\ 11]{Cov06} that the empirical joint distribution on edges $e$ and $e'$ satisfies
\begin{equation}
\lim_{n\rightarrow \infty} -\frac{1}{n} \log\bP(\hP_{e,e'} \in \calF) = \inf \{ D(Q\, ||\,  P_{e,e'}):Q\in \calF  \}
\end{equation}
for any set $\calF \subset \calP(\calX^4)$ that  equals the closure of its interior, {\it i.e.}, $\calF =\mathrm{cl}(\mathrm{int}(\calF))$. We now have a LDP for the sequence of probability measures $\hP_{e,e'}$, the empirical distribution on $(e,e')$. Assuming that $e$ and $e'$ do not share a common node, $\hP_{e,e'}\in\calP(\calX^4)$ is a probability distribution over four variables (the variables in the node pairs $e$ and $e'$). We now define the function $h:\calP(\calX^4) \rightarrow \bR$ as
\begin{equation}
h(Q)~:=~I(Q_{e'})-I(Q_e).\label{eqn:h}
\end{equation}
Since $Q_e = \sum_{x_{e'}} Q$, defined in \eqref{eqn:Qe} is continuous in $Q$ and the mutual information $I(Q_e)$ is also continuous in $Q_e$, we conclude that $h$ is indeed continuous, since it is the composition of continuous functions. By applying the contraction principle~\cite{Den00} to the sequence of probability measures $\hP_{e,e'}$ and the continuous map $h$, we obtain a corresponding LDP for the new sequence of probability measures $h(\hP_{e,e'})= I(\hP_{e'})- I(\hP_{e})$, where the rate is given by:
\begin{align}
J_{e,e'}&=\inf_{Q\in \calP(\calX^4) } \left\{D(Q\, ||\, P_{e,e'}) : h(Q)\ge 0\right\}, \\
&= \inf_{Q\in \calP(\calX^4) } \left\{D(Q\, ||\, P_{e,e'}) : I(Q_{e'})\ge I(Q_e) \right\}.\label{eqn:Jeeinf}
\end{align}
We now claim that the limit in~\eqref{eqn:Jee} exists. From Sanov's
theorem~\cite[Ch.\ 11]{Cov06}, it suffices to show that the
constraint set $\calF:=\{I(Q_{e'})\ge I(Q_e)\}$
in~\eqref{eqn:Jeeinf} is a regular closed set, {\it i.e.}, it
satisfies $\calF=\mathrm{cl}(\mathrm{int}(\calF))$. This is true
because there are no isolated points in $\calF$ and thus the
interior is nonempty. Hence, there exists a sequence of
distributions $\{Q_n\}_{n=1}^{\infty}\subset\mathrm{int}(\calF)$
such that $\lim_{n\to\infty}D(Q_n||P_{e,e'})= D(Q^*||P_{e,e'})$,
which proves the existence of the limit in \eqref{eqn:Jee}.
%\footnote{In
%\eqref{eqn:Cee}, even though the union is over $r\ge 0$, there are
%only finitely many events  because the collection of all types
%(empirical distributions) is polynomial in
%$n$~\cite[Ch.\ 11]{Cov06}.} .  \\\\
%{\it Step 2}: Next,  the dominant event in the crossover eventof
%mutual information quantities on $e$ and $e'$
%\begin{eqnarray}
%\calC_{e,e'}=\left\{I(\hP_{e'})  \ge I(\hP_{e}) \right\}  \!=\! \bigcup_{r\ge 0} \left\{I(\hP_{e'}) = I(\hP_{e})+r \right\},
%\label{eqn:Cee}
%\end{eqnarray}
%is the equality case, {\it i.e.}, $\{  I(\hP_{e})= I(\hP_{e'})\}$ is
%the dominant event. This is because  $ I(\hP_{e'})- I(\hP_{e})$ is a
%random variable with  mean\footnote{Recall that we assumed {\em
%a-priori} that $I(P_e)>I(P_{e'})$.} $I(P_{e'})- I(P_{e})<0$, and
%hence,  the event   $\{I(\hP_{e'})- I(\hP_{e})= 0\}$ dominates
%$\cup_{r\geq 0}\{I(\hP_{e'})- I(\hP_{e})=r\}$,   {\it i.e.}, $
%\bP(\calC_{e,e'})\doteq \bP(\{  I(\hP_{e})= I(\hP_{e'})\}).$ This
%proves the first assertion in Theorem~\ref{thm:cross_emp}.
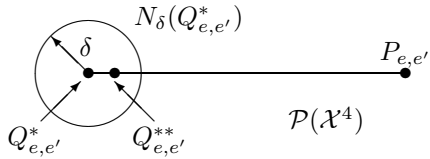
\begin{figure}
\centering
\input{dominantcase}
\caption{Illustration of Step 2 of the proof of Theorem \ref{thm:cross_emp}.}
\label{fig:domcase}
\end{figure}
%\Comment Can u change the part below..

{\it Step 2}: We now show that the optimal solution $Q_{e,e'}^*$, if it exists (as will be shown in Step 3), must satisfy $I(Q_e^*)= I(Q_{e'}^*)$. Suppose, to the contrary, that $Q_{e,e'}^*$ with objective value $D(Q_{e,e'}^*||P_{e,e'})$ is such that $I(Q_{e'}^*)>I(Q_e^*)$. Then $h(Q_{e,e'}^*)>0$, where $h$, as shown above, is continuous. Thus, there exists a $\delta>0$ such that the $\delta$-neighborhood 
\begin{equation}
N_{\delta}(Q_{e,e'}^*):=\{R: \|R-Q_{e,e'}^*\|_{\infty}<\delta\} ,\end{equation} satisfies $h(N_{\delta}(Q_{e,e'}^*)) \subset (0,\infty)$~\cite[Ch.\ 2]{Rudin}. Consider the new distribution (See Fig.~\ref{fig:domcase})
\begin{align}
Q_{e,e'}^{**} &= Q_{e,e'}^* + \frac{\delta}{2}(P_{e,e'} - Q_{e,e'}^*)\\
& = \left(1-\frac{\delta}{2}\right) Q_{e,e'}^* +  \frac{\delta}{2}P_{e,e'}.
\end{align}
Note that $Q_{e,e'}^{**} $ belongs to $ N_{\delta}(Q_{e,e'}^*)$ and hence is a feasible solution of \eqref{eqn:Jeeinf}.  We now prove that $D(Q_{e,e'}^{**}||P_{e,e'})<D(Q_{e,e'}^*||P_{e,e'})$, which  contradicts the optimality of $Q_{e,e'}^*$.
\begin{align}
&D(Q_{e,e'}^{**}\, ||\, P_{e,e'}) \nn\\
&= D\left(\left(1-\frac{\delta}{2}\right) Q_{e,e'}^* +  \frac{\delta}{2}P_{e,e'}\,  \Big\|\,  P_{e,e'} \right), \\
&\le \left(1-\frac{\delta}{2}\right) D(Q_{e,e'}^*\, ||\,  P_{e,e'}) + \frac{\delta}{2} D(P_{e,e'}\, ||\,  P_{e,e'}), \label{eqn:convex}\\
&= \left(1-\frac{\delta}{2}\right) D(Q_{e,e'}^*\,||\, P_{e,e'})\label{eqn:dqp_equal}\\
&< D(Q_{e,e'}^*\, ||\, P_{e,e'}), \label{eqn:deltapos}
\end{align}
where \eqref{eqn:convex} is due to the convexity of the KL-divergence in the first variable~\cite[Ch.\ 2]{Cov06}, \eqref{eqn:dqp_equal} is because $D(P_{e,e'}|| P_{e,e'})=0$ and \eqref{eqn:deltapos} is because $\delta>0$. Thus, we conclude that the optimal solution must satisfy $I(Q_e^*)= I(Q_{e'}^*)$ and the crossover rate can be stated as \eqref{eqn:Jee2}.

{\it Step 3}: Now, we prove the existence of the minimizer $Q^*_{e,e'}$,
which will allow us to replace the $\inf$ in \eqref{eqn:Jee2} with
$\min$. First, we note that $D(Q\,||\, P_{e,e'})$ is continuous in
both variables and hence continuous and the first variable $Q$. It
remains to show that the constraint set
\begin{equation}
\Lambda := \{Q\in \calP(\calX^4): I(Q_{e'}) =  I(Q_e)  \}\label{eqn:Lambda}
\end{equation}
is compact, since it is clearly nonempty (the uniform distribution belongs to $\Lambda$). Then we can conclude, by Weierstrass' extreme value theorem~\cite[Theorem 4.16]{Rudin}, that the minimizer $Q^*\in \Lambda$ exists. By the Heine-Borel theorem~\cite[Theorem 2.41]{Rudin}, it suffices to show that $\Lambda$ is bounded and closed. Clearly $\Lambda$ is bounded since $\calP(\calX^4)$ is a bounded set. Now, $\Lambda=h^{-1}(\{0\})$ where $h$ is defined in \eqref{eqn:h}. Since $h$ is continuous and $\{0\}$ is closed (in the usual topology of the real line), $\Lambda$ is closed~\cite[Theorem 4.8]{Rudin}. Hence that $\Lambda$ is compact. We also need to use the fact that $\Lambda$ is compact in the proof of Theorem~\ref{thm:consistency}.
%Now, consider $\Lambda^c=\calP(\calX^4)\setminus \Lambda  = \{Q\in \calP(\calX^4): I(Q_e)\ne I(Q_{e'})\}$. We need to show that $\Lambda^c$ is open in $\calP(\calX^4)$. The set $\Lambda^c$ can be equivalently written as $\Lambda^c = \{Q\in \calP(\calX^4): h(Q)< 0\}$ where $h$ is defined in \eqref{eqn:h}. Since $h$ is continuous, for each $Q_0\in\Lambda^c$, with $h(Q_0)>0$, there exists a $\epsilon>0$ such that every $R\in N_{\epsilon}(Q_0)=\{Q: \|Q-Q_0\|_{\infty}<\epsilon\}$ satisfies $h(R)> 0$. A similar argument holds if $h(Q_0)<0$. Hence, $N_{\epsilon}(Q_0)\subset \Lambda^c$ and $Q_0$ is an interior point. Since $Q_0$ was arbitrary, every point in $\Lambda^c$ is an interior point, which proves $\Lambda^c$ is open, that $\Lambda$ is closed and  hence that $\Lambda$ is compact. We also need to use the fact that $\Lambda$ is compact in the proof of Theorem~\ref{thm:consistency}.
\end{IEEEproof}

%\section{Proof of Proposition \ref{prop:star}: Error Exponent for ``Star'' graph}
%\begin{IEEEproof}
%Since there are only two distinct distributions $Q_a$ (which corresponds to a true edge) and $Q_b$ (which corresponds to a non-edge), there is only {\em one} unique rate $J_{e,e'}$, namely the expression in \eqref{eqn:Jee2} with $P_{e,e'}$ replaced by $Q_{a,b}$. If the event $\calA_{e,e'}$ occurs, an error definitely occurs. This corresponds to the case where {\em any one} edge $e\in \calE_P$ is replaced by {\em any other} edge not in $\calE_P$.
%\end{IEEEproof}

\section{Proof of Theorem \ref{thm:dom}} \label{prf:dom}
\begin{IEEEproof}
We first claim that $\calE_P^*$, the edge set corresponding to the
dominant error tree, differs from $\calE_P$ by  exactly one
edge.\footnote{This is somewhat analogous to the fact that the
second-best MWST differs from the MWST by  exactly one
edge~\cite{Cor03}.} To prove this claim, assume, to the contrary,
that $\calE_P^*$ differs from $\calE_P$ by two edges. Let
$\hcalE=\calE':= \calE_P\setminus \{e_1,e_2\} \cup \{e_1', e_2'\}$, where
$e_1', e_2'\notin\calE_P$ are the two edges that have replaced
$e_1,e_2\in \calE_P$ respectively. Since $T'=(\calV,\calE')$ is a tree, these edges
cannot be arbitrary and specifically, $\{e_1, e_2\} \in
\{\Path(e_1';\calE_P)\cup \Path(e_2';\calE_P)\}$ for the tree
constraint to be satisfied. Recall that  the rate of the event that
the output of the ML algorithm is $T'$ is given by $\Upsilon(T')$
in \eqref{eqn:kt}.  Then consider the probability of the joint event (with
respect to the probability measure $\bP=P^n$).

%\Comment Can u make diagrams for these cases? I am sending handdrawn
%ones.

%{\em Case 1: }

Suppose that  $e_i\in  \Path(e_i';\calE_P)$ for $i=1,2$ and
$e_i\notin  \Path(e_j';\calE_P)$ for $i,j=1,2$ and $ i\ne j$. See Fig.~\ref{fig:case1}. Note that
the true mutual information quantities satisfy $I(P_{e_i}) >
I(P_{e_i'})$. We  prove this claim by contradiction that suppose
$I(P_{e_i'}) \ge I(P_{e_i})$ then,  $\calE_P$  does not have
maximum weight because if the non-edge $e_i'$ replaces the true edge
$e_i$, the resulting tree\footnote{The resulting graph is indeed a
tree because $\{e_i'\}\cup \Path(e_i';\calE_P)$ form a cycle so if
any edge is removed, the resulting structure does not have any
cycles and is connected, hence it is a tree.  See
Fig.~\ref{fig:replace}. }  would have higher weight, contradicting
the optimality of the true edge set  $\calE_P$, which is the MWST
with the true mutual information quantities as edge weights. More precisely, we can compute the exponent when $T'$ is the output of the MWST algorithm:
%Then the rate of this event is no less than if either $e_1'$ replaced $e_1$ or $e_2'$ replaced $e_2$ only. This is because
\begin{align} 
\Upsilon(T')&= \lim_{n\to \infty} -\frac{1}{n} \log \bP\left(
\bigcap_{i=1,2} \{I(\hP_{e_i'})\geq I(\hP_{e_i})\}\right), \\
%&\le \, \min \{\bP(\{e_1' \text{ replaces } e_1 \}) ,\,  \bP(  \{e_2' \text{ replaces } e_2 \}), \nn\\
%&\doteq \,  \min(\exp(-nJ_{e_1,e_1'}),\exp(-nJ_{e_2,e_2'}),\exp(-nJ_{e_1,e_2'}),\exp(-nJ_{e_2,e_1'})),\nn\\
&\geq \max_{i=1,2}\lim_{n\to \infty} -\frac{1}{n} \log \bP\left( \{
I(\hP_{e_i'})\geq I(\hP_{e_i})\}\right), \\  &=\,  \max  \left\{ J_{e_1,e_1'} \, ,\,  J_{e_2,e_2'} \right\}.\label{eqn:ldpmax}
\end{align}
Now  $J_{e_i, e_i'} = \Upsilon(T_i)$ where $ T_i:= (\calV,  \calE_P\setminus
\{e_i\} \cup\{e_i'\})$. From Prop.\ \ref{prop:domtree},
the error exponent associated to  the dominant error tree, {\it i.e.},  $K_P=
\min_{T\ne T_P} \Upsilon(T)$ and from \eqref{eqn:ldpmax}, the dominant
error tree cannot be $T'$ and should differ from $T_P$ by one and only one
edge.

\begin{figure}
\centering
\input{prfthm5}
\caption{Illustration of the proof of Theorem \ref{thm:dom}. The dominant error event involves only one crossover event. }
\label{fig:case1}
\end{figure}
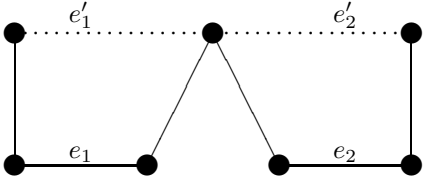

\iffalse
{\em Case 2: } $e_i\in  \Path(e_j';\calE_P)$ for $i,j=1,2$. See Fig.~\ref{fig:case2}. Now, we
have $I(P_{e_i}) \geq I(P_{e_j'})$ for $i,j=1,2$ and
%\begin{equation}
%\Bigg[ \Bigl(\bigcap_{\substack{i=1,2 \\j \ne i}}
%\{I(\hP_{e_i'})\geq I(\hP_{e_j})\}\Bigr) \bigcup \Bigl(\bigcap_{i=1,2}
%\{I(\hP_{e_i'})\geq I(\hP_{e_i})\}\Bigr)\Bigg]
%\end{equation}
\begin{align} \Upsilon(T')= \lim_{n\to \infty}  \, &\, -\frac{1}{n} \log
\bP\Bigg[ \Big(\bigcap_{\substack{i=1,2 \\j \ne i}}
\{I(\hP_{e_i'})\geq I(\hP_{e_j})\}\Big)
\nn \\ &  \bigcup \Big(\bigcap_{i=1,2}
\{I(\hP_{e_i'})\geq I(\hP_{e_i})\}\Big)\Bigg]. \end{align}
%&\le \, \min \{\bP(\{e_1' \text{ replaces } e_1 \}) ,\,  \bP(  \{e_2' \text{ replaces } e_2 \}), \nn\\
%&\doteq \,  \min(\exp(-nJ_{e_1,e_1'}),\exp(-nJ_{e_2,e_2'}),\exp(-nJ_{e_1,e_2'}),\exp(-nJ_{e_2,e_1'})),\nn\\
Hence, we have
\begin{equation}
\Upsilon(T') \geq \min\left\{ \max_{i=1,2}
J_{e_i,e_i'}\, ,\,  \max_{ i,j=1,2 ,\,  j \ne i} J_{e_i,
e_j'}\right\}.
\end{equation}
Again, $T'$ cannot be the dominant error tree.\fi

The similar conclusion holds for   the two other cases (i)  $e_i\in
\Path(e_i';\calE_P)$ for $i=1,2$, $e_2\in  \Path(e_1';\calE_P)$  and
$e_1\notin  \Path(e_2';\calE_P)$ and (ii)    $e_i\in
\Path(e_i';\calE_P)$ for $i=1,2$, $e_1\in  \Path(e_2';\calE_P)$  and
$e_2\notin  \Path(e_1';\calE_P)$. In other words, the dominant error tree differs from the true tree by one   edge. 

We now use the    ``worst-exponent-wins principle''~\cite[Ch.\
1]{Den00}, to conclude that the rate that dominates is the minimum
$J_{r(e') , e'}$ over all possible $e'\notin \calE_P$, namely
$J_{r(e^*) , e^*}$ with $e^*$ defined in \eqref{eqn:eprime}. More precisely,
\begin{align}
\bP(\calA_n)&= \bP\Bigg( \!\!\bigcup_{e'\notin\calE_P}\!\! \{ e' \text{
replaces any } e\in \Path(e'; \calE_P)\text{ in } \hTML\} \!\!
\Bigg), \nn\\
&= \!\bP \Bigg(\! \bigcup_{e'\notin\calE_P}\bigcup_{e\in\Path(e';\calE_P)} \!\!\{ e'
\text{ replaces } e \text{ in } \hTML \} \!
\Bigg), \\
&\le \sum_{e'\notin\calE_P} \sum_{e\in\Path(e';\calE_P)} \bP (   \{ e' \text{ replaces } e \text{ in } \hTML\} ), \label{eqn:unionbd} \\
&= \sum_{e'\notin\calE_P} \sum_{e\in\Path(e';\calE_P)} \bP  (   \{
I(\hP_{e'})\ge I(\hP_e)  \}  ),
\label{eqn:empmi_cross}\\
%&= \sum_{e'\notin\calE_P} \sum_{e\in\Path(e';\calE_P)} \bP (  \calC_{e,e'} ), \label{eqn:cee}\\
&\doteq \sum_{e'\notin\calE_P} \sum_{e\in\Path(e';\calE_P)} \exp(-nJ_{e,e'}), \label{eqn:defJee} \\
&\doteq \exp\left(-n\min_{e'\notin\calE_P} \min_{e\in\Path(e';\calE_P)} J_{e,e'}\right), \label{eqn:expapp}
%&= \exp\left(-nJ_{r(e^*),e^*} \right). \label{eqn:def_Jee}
\end{align}
where~\eqref{eqn:unionbd} is from the union bound, \eqref{eqn:empmi_cross} and \eqref{eqn:defJee}  are from the definitions of the crossover  event and rate respectively (as described in Cases 1 and 2 above) and~\eqref{eqn:expapp} is an
application of the ``worst-exponent-wins'' principle~\cite[Ch.\ 1]{Den00}.
\iffalse\begin{figure}
\centering
\input{prfthm5_2}
\caption{Illustration for Case 2 of the proof of Theorem \ref{thm:dom}.}
\label{fig:case2}
\end{figure}\fi

%Eq.~ deserves further elaboration. We claim that the crossover event of mutual information quantities (defined in \eqref{eqn:Aee}) can be written as
%\begin{equation}
%\calC_{e,e'} = \{ I(\hP_{e'})\ge I(\hP_e) \}=\{ e' \text{ replaces } e \}.
%\end{equation}
%To prove this claim, fix a non-edge $e'$. It suffices to show that $I(P_{e'}) < I(P_e)$ for all $e\in\Path(e';\calE_P)$. Suppose, to the contrary, there exists an edge $\tilde{e}\in \Path(e';\calE_P)$ such that $I(P_{e'}) \ge I(P_{\tilde{e}})$. Then  $\calE_P$  does not have  maximal weight because if the non-edge $e'$ replaces the true edge $\tilde{e}$, the resulting tree\footnote{The resulting graph is a indeed tree because $\{e'\}\cup \Path(e';\calE_P)$ forms a loop so if any edge is removed, the resulting structure does not have any cycles and is connected, hence it is a tree.  See Fig.~\ref{fig:replace}. }  would have higher weight, contradicting the optimality of the true edge set  $\calE_P$, which is the MWST with the true mutual informations as edge weights.

We conclude\footnote{The notation $a_n\stackrel{.}{\le }b_n$   means that $\limsup_{n\to\infty}\frac{1}{n}\log (a_n/b_n)\le 0$. Similarly,  $a_n\stackrel{.}{\ge}b_n$  means that  $\liminf_{n\to\infty}\frac{1}{n}\log (a_n/b_n)\ge 0$. } from~\eqref{eqn:expapp} that
\begin{equation}
\bP(\calA_n)\stackrel{.}{\le }
\exp(-nJ_{r(e^*),e^*}),\label{eqn:upperexpbd}
\end{equation}
from the definition of the dominant replacement edge $r(e')$ and the
dominant non-edge $e^*$, defined in \eqref{eqn:dre} and
\eqref{eqn:eprime} respectively. The lower bound follows trivially
from the fact that if $e^*\notin\calE_P$ replaces $r(e^*)$, then the
error $\calA_n$  occurs. Thus, $\{e^*  \text{ replaces
}r(e^*)\}\subset \calA_n$ and
\begin{align}
\bP(\calA_n)&\stackrel{.}{\ge}\bP(\{e^*  \text{ replaces }r(e^*)
\text{ in }\hTML\})   \\
&\doteq \exp(-nJ_{r(e^*),e^*}).  \label{eqn:lowerexpbd}
 \end{align}
Hence, \eqref{eqn:upperexpbd} and \eqref{eqn:lowerexpbd} imply that $\bP(\calA_n)\doteq \exp(-nJ_{r(e^*),e^*}),$ which proves our
main result in \eqref{eqn:Jfinal2}. 

The finite-sample result in \eqref{eqn:finite_sam} comes from the upper bound in \eqref{eqn:expapp} and the following two elementary facts:
\begin{enumerate}
\item The exact number of $n$-types with alphabet $\mathcal{Y}$ is given by  $\binom{n+1+|\mathcal{Y}|}{n+1}$ \cite{Csi97}. In particular, we have
\begin{equation}
\bP(\calC_{e,e'}) \le \binom{n+1+|\calX|^4}{n+1}\exp(-nJ_{e,e'}), \label{eqn:finite_C} 
\end{equation}
for all $n\in\mathbb{N}$, since $\calC_{e,e'}$ only involves the distribution $P_{e,e'}\in\calP(\calX^4)$. Note that the exponent $4$ of $|\calX|^4$ in~\eqref{eqn:finite_C} is an upper bound since if $e$ and $e'$ share a node $P_{e,e'}\in\calP(\calX^3)$.
\item The number of error events $\calC_{e,e'}$ is at most $(d-1)^2(d-2)/2$ because there are $\binom{d}{2}-(d-1)=(d-1)(d-2)/2$ non-edges and for each non-edge, there are at most $d-1$ edges along its path. 
\end{enumerate}
This completes the proof.
\end{IEEEproof}

\section{Proof of Theorem \ref{thm:JzeroI}} \label{prf:JzeroI}
Statement (a) $\Leftrightarrow$ statement (b) was proven in full after  the theorem was stated. Here we provide the proof that (b) $\Leftrightarrow$ (c). Recall that statement (c) says that $T_P$ is not a proper forest. We first begin with a preliminary lemma.
%Proof of part (a):
%\begin{IEEEproof}
%($\Rightarrow$) Assume $K>0$. Suppose, to the contrary, that $I_{e'}= I_{r(e')}$ for some $e'\notin \calE_P$. Then the infimum in~\eqref{eqn:Jee2} equals to 0 because it is attained for some $e' \notin \calE_P$. By~\eqref{eqn:Jfinal}, $K=0$, a contradiction.
%
%($\Leftarrow$) Now assume that $I_{e'}\ne I_{r(e')}$ for all $e'\notin \calE_P$. Then, clearly $J_{r(e') , e'}>0$ for all $e'$. From~\eqref{eqn:Jfinal2}, $K>0$ since there are only finitely many $e'$, hence the minimum in \eqref{eqn:eprime} is attained.
%\end{IEEEproof}
\begin{lemma} \label{lem:jointindep}
Suppose $x,y,z$ are three random variables taking on values on finite sets $\calX,\calY,\calZ$ respectively. Assume that $P(x,y,z)>0$ everywhere. Then $x - y - z$ and $x - z - y$ are Markov chains if and only if $x$ is jointly independent of $y,z$.
\end{lemma}
\begin{IEEEproof}
($\Rightarrow$) That $x - y - z$ is a Markov chain implies that
\begin{equation}
P(z|y,x)= P(z|y),
\end{equation}
or alternatively
\begin{equation}
P(x,y,z)=P(x,y)\frac{P(y,z)}{P(y)}.\label{eqn:hc1}
\end{equation}
Similarly from the fact that $x - z - y$  is a Markov chain, we have
\begin{equation}
P(x,y,z)=P(x,z)\frac{P(y,z)}{P(z)}. \label{eqn:hc2}
\end{equation}
Equating \eqref{eqn:hc1} and \eqref{eqn:hc2}, and use the positivity to cancel $P(y,z)$, we arrive at
\begin{equation}
P(x|y)=P(x|z).
\end{equation}
It follows that $P(x|y)$ does not depend on $y$, so there is some constant $C(x)$ such that $P(x|y)=C(x)$ for all $y\in \calY$. This immediately implies that $C(x)=P(x)$ so that $P(x|y)=P(x)$. A similar argument gives that $P(x|z)=P(x)$. Furthermore, if $x - y - z$ is a Markov chain, so is $z - y -x$, therefore
\begin{equation}
P(x|y,z)=P(x|y)=P(x).
\end{equation}
The above equation says that $x$ is jointly independent of both $y$ and $z$. \\
($\Leftarrow$) Conversely, if $x$ is jointly independent of both $y$ and $z$, then $x - y - z$ and $x - z - y$ are Markov chains. In fact $x$ is not connected to $y - z$.
\end{IEEEproof}

\begin{IEEEproof}
We now prove (b) $\iff$ (c) using Lemma~\ref{lem:jointindep} and  the assumption that $P(\bx)>0$ for all $\bx\in \calX^d$.\\
%($\Rightarrow$) We prove that statement (b) implies (c). Statement (b) says that $I(P_{e'})\ne I(P_{r(e')})$ for every $e'\notin \calE_P$. Suppose, to the contrary, that $T_P$ is a proper forest. Then there does not exist a path between the two nodes of $e'$ for some $e'\notin \calE_P$. Consider any such $e'$. Let these two disconnected components be $\calG_1=(\calV_1,\calE_1)$ and $\calG_2=(\calV_2,\calE_2)$ such that $\calV_1\cup \calV_2=\calV$. Then the variables in $\calG_1$ are jointly independent of those in $\calG_2$. Let $x_1$ be any one variable in the component $\calG_1$. Then, by  Lemma~\ref{lem:jointindep}, if $x_2,x_3\in \calV_2$, then $x_1 - x_2- x_3$ and $x_1- x_3- x_2$ are Markov chains. This means that $I(P_{e'})=I_{(1,2)}=I_{(2,3)}= I(P_{r(e')})$,  contradicting statement (b). \\
%($\Leftarrow$) We prove that statement (c)  implies (b). Thus, we assume that $P$ is not Markov on a proper forest, {\it i.e.}, it is a spanning tree. Suppose, to the contrary, that   $I(P_{e'})= I(P_{r(e')})$ for some  $e'\notin\calE_P$. Let $e' =(1,2)$ and $r(e')=(2,3)$. Then $x_1 - x_2 - x_3$ and $x_1 - x_3 - x_2$ are Markov chains. From  Lemma~\ref{lem:jointindep}, $x_1$ is jointly independent of $x_2$ and $x_3$, which violates the fact that $T_P$ is a spanning tree. %By the lemma and by reversing the preceding argument, $I_{e'}\ne I_{r(e')}$ for all $e'\notin\calE_P$. Finally, by Theorem \ref{thm:JzeroI}, $K>0$.
($\Rightarrow$) If (b) is true then $I(P_{e'})< I(P_{e})$ for all $e\in\Path(e';\calE_P)$ and for all $e'\notin\calE_P$. Assume, to the contrary, that $T_P$ is a proper forest, {\it i.e.}, it contains at least 2 connected components (each connected component may only have one node), say $\calG_i=(\calV_i, \calE_i)$ for $i=1,2$. Without loss of generality, let $x_1$ be in component $\calG_1$ and $x_2,x_3$ belong to component $\calG_2$. Then since $\calV_1\cap \calV_2=\emptyset$ and $\calV_1\cup \calV_2=\calV$, we have that $x_1$ jointly independent of  $x_2$ and $ x_3$. By  Lemma~\ref{lem:jointindep}, we have the following Markov chains $x_1- x_2 -x_3$ and $x_1- x_3 -x_2$. This implies from the Data Processing Inequality~\cite[Theorem 2.8.1]{Cov06} that $I(P_{1,2})\ge I(P_{1,3})$ and at the same time $I(P_{1,2})\le I(P_{1,3})$ which means that $I(P_{1,2})= I(P_{1,3})$. This contradicts (b) since by taking $e'=(1,2)$, the mutual informations along the path $\Path(e';\calE_P)$ are no longer distinct. \\
($\Leftarrow$) Now assume that (c) is true, {\it i.e.}, $T_P$ is not a proper forest. Suppose, to the contrary, (b) is not true, {\it i.e.}, there exists a $e'\notin \calE_P$ such that $I(P_{e'}) = I(P_{r(e')})$, where $r(e')$ is the replacement edge associated with the non-edge $e'$. Without loss of generality, let $e'=(1,2)$ and $r(e')=(3,4)$, then since $T_P$ is not a proper forest, we have the following Markov chain $x_1-x_3-x_4-x_2$. Now note that $I(P_{1,2}) = I(P_{3,4})$. In fact, because there is no loss of mutual information  $I(P_{1,4}) = I(P_{3,4})$ and hence by the Data Processing Inequality we also have  $x_3-x_1-x_4-x_2$. By using Lemma~\ref{lem:jointindep}, we have $x_4 $ jointly independent of  $x_1$ and $x_3$, hence we have a proper forest, which is a contradiction.
\end{IEEEproof}

%\section{Proof of Theorem \ref{thm:comp}: Computational Complexity for $K_P$}\label{prf:comp}
%\begin{IEEEproof}
%Given a non-neighbor node pair $e'\notin\calE_P$, we perform a maximum of $\zeta(T_P)$ calculations to determine the dominant replacement edge $r(e')$ from~\eqref{eqn:dre}.  Combining this with the fact that there are a total of $|\calV^2\setminus \calE_P|=\binom{d}{2}-(d-1)=\frac{1}{2} (d-1)(d-2)$ not in $\calE_P$, we obtain the upper bound.
%\end{IEEEproof}

%\begin{figure}
%\centering
%\includegraphics[width=.65\columnwidth]{figs/proj}
%\caption{ A geometric interpretation of~\eqref{eqn:Jee2} where $P_{e,e'}$ is projected onto the manifold. }
%\label{fig:proj}
%\end{figure}

%\begin{figure}
%\centering
%\includegraphics[width=.45\columnwidth]{figs/proj2}
%\caption{Convexifying the objective results in a least-squares problem.}
%\label{fig:proj2}
%\end{figure}

\section{Proof of Theorem \ref{thm:euc} }\label{prf:euc}
\begin{IEEEproof}
The proof proceeds in several steps. See Figs.\ \ref{fig:proj} and \ref{fig:proj2} for intuition behind this proof.

{\em Step 1}: Let $Q$ be such that
\begin{equation}
Q(x_i,x_j,x_k,x_l)=P_{e,e'}(x_i,x_j,x_k,x_l)+\epsilon_{i,j,k,l}.
\end{equation}
Thus, the $\epsilon_{i,j,k,l}$'s are the deviations of $Q$ from $P_{e,e'}$. To ensure that $Q$ is a valid distribution we require $\sum \epsilon_{i,j,k,l}=0$. The objective in \eqref{eqn:defJeetilde} can now be alternatively expressed as
\begin{equation}
\frac{1}{2}\beps^T \bK_{e,e'} \beps= \frac{1}{2}\sum_{x_i,x_j,x_k,x_l} \frac{\epsilon_{i,j,k,l}^2}{P_{e,e'}(x_i,x_j,x_k,x_l)},\label{eqn:Kee}
\end{equation}
where $\beps\in \bR^{|\calX|^4}$ is the vectorized version of the deviations $\epsilon_{i,j,k,l}$ and $\bK_{e,e'}$ is a $|\calX|^4\times |\calX|^4$ diagonal matrix containing the entries $1/P_{e,e'}(x_i,x_j,x_k,x_l)$ along its diagonal.

{\em Step 2}: We now perform a first-order Taylor expansion of $I(Q_e)$ in the neighborhood of $I(P_e)$.
\begin{align}
I(Q_e)&= I(P_e)+\beps^T \nabla_{\beps} I(Q_e)\Big|_{\beps=0}+ o(\|\beps\|) ,\\
&= I(P_e)+\beps^T \bs_e  + o(\|\beps\|), \label{eqn:taylor_IQ}
\end{align}
where $\bs_e$ is the length $|\calX|^4$-vector that contains the information density values of edge $e$. Note that because of the assumption that $P$ is not a proper forest,  $P_{i,j}\ne P_i \, P_j$ for all $(i,j)$, hence the linear term does not vanish.\footnote{Indeed if $P_e$ were a product distribution, the linear term in \eqref{eqn:taylor_IQ} vanishes and $I(Q_e)$ is approximately a quadratic in $\beps$ (as shown in \cite{Bor08}).} The constraints can now be rewritten as
\begin{align}
\beps^T \one ~=~ 0,\qquad \beps^T (\bs_{e'} -\bs_{e})  ~ = ~ I(P_{e})-I(P_{e'})
. \label{eqn:linear_constr}
\end{align}
%This is a convex least-squares problem since the objective is quadratic in $\beps$ and the constraints are linear in $\beps$. Thus, the optimal solution is achieved when equality holds in \eqref{eqn:linear_constr}, {\it i.e.}, the constraints can be written as
%\footnote{Suppose the optimal $\beps^*$ is in the interior of the constraint set, then $\beps^*_{\delta}=(1-\delta)\beps^*$ would also be in the constraint set in \eqref{eqn:linear_constr}   for $\delta$ sufficiently small. But $\beps^*_{\delta}$ would have smaller norm, which contradicts the optimality of $\beps^*$. Thus, the optimal solution must be on the boundary, {\it i.e.}, when equality holds. }
or in matrix notation as:
\begin{equation}
\left[ \begin{array}{c}
\bs_{e'}^T-\bs_{e}^T\\ \one^T
\end{array}\right] \beps = \left[ \begin{array}{c}
 I(P_{e})-I(P_{e'})\\ 0
\end{array}\right],  \label{eqn:lin_constr}
\end{equation}
where $\one$ is the length-$|\calX|^4$ vector consisting of all ones. For convenience, we define $\bL_{e,e'}$ to be the matrix in \eqref{eqn:lin_constr}, {\it i.e.},
\begin{equation}
\bL_{e,e'}:=\left[ \begin{array}{c}
\bs_{e'}^T-\bs_{e}^T\\ \one^T
\end{array}\right]  \in\bR^{2\times |\calX|^4}.
\end{equation}
{\em Step 3}: The optimization problem now reduces to minimizing~\eqref{eqn:Kee} subject to the  constraints in \eqref{eqn:lin_constr}. This is a standard least-squares problem. By using the Projection Theorem in Hilbert spaces, we get the solution
\begin{eqnarray}
\beps^* \!=\!  \bK_{e,e'}^{-1}   \bL_{e,e'}^T( \bL_{e,e'} \bK_{e,e'}^{-1} \bL_{e,e'}^T)^{-1} \left[\! \begin{array}{c}
 I(P_{e})\! - \! I(P_{e'})\\ 0
\end{array}\!\right].\label{eqn:eps}
\end{eqnarray}
The inverse of $\bL_{e,e'} \bK_{e,e'}^{-1} \bL_{e,e'}^T$ exists because we assumed $T_P$ is not a proper forest and hence $P_{i,j}\ne P_i P_j$ for all $(i,j)\in\binom{\calV}{2}$.  This is a sufficient condition for the matrix $\bL_{e,e'}$ to have full row rank and thus, $\bL_{e,e'} \bK_{e,e'}^{-1} \bL_{e,e'}^T$ is invertible. Finally, we substitute $\beps^*$ in \eqref{eqn:eps} into~\eqref{eqn:Kee} to obtain
\begin{equation}
\tilJ_{e,e'}=\frac{1}{2}  \left[( \bL_{e,e'} \bK_{e,e'}^{-1} \bL_{e,e'}^T)^{-1}\right]_{11} ( I(P_{e})-I(P_{e'}))^2,
\end{equation}
where $[\mathbf{M}]_{11}$ is the (1,1) element of the matrix $\mathbf{M}$. Define $\psi$ to be the weighting function given by
\begin{equation}
\psi(P_{e,e'}) := \left[( \bL_{e,e'} \bK_{e,e'}^{-1} \bL_{e,e'}^T)^{-1} \right]_{11}.\label{eqn:psi}
\end{equation}
%Note from assumption A1 that $g$ is {\em uniformly continuous} in $P_{e,e'}$ since it is a continuous function on a compact set~\cite{Rudin}.
It now suffices to show that $\psi(P_{e,e'})$ is indeed the inverse variance of $s_e-s_{e'}$. We now simplify the expression for the weighting function $\psi(P_{e,e'})$ recalling how $\bL_{e,e'}$ and $\bK_{e,e'}$ are defined. The product of the matrices in \eqref{eqn:psi} is
\begin{equation}
\bL_{e,e'} \bK_{e,e'}^{-1} \bL_{e,e'}^T=\left[ \begin{array}{cc}
\bE[(s_{e'}- s_e)^2] & \bE[ s_{e'}- s_e]\\ \bE[ s_{e'}- s_e]& 1
\end{array}\right],\label{eqn:prodmatrices}
\end{equation} 
where all expectations are with respect to the distribution $P_{e,e'}$. Note that the determinant of \eqref{eqn:prodmatrices} is $ \bE[(s_{e'}- s_e)^2] - \bE[(s_{e'}- s_e)]^2=\var(s_{e'}- s_e)$. Hence, the (1,1) element of the inverse of  \eqref{eqn:prodmatrices} is simply 
\begin{equation}
 \psi(P_{e,e'})=  \var(s_{e'}- s_e)^{-1}.
\end{equation} 
Now, if $e$ and $e'$ share a node, this proof proceeds in exactly the same way. In particular, the crucial step~\eqref{eqn:taylor_IQ} will also remain the same since the Taylor expansion does not change. This concludes the first part of the proof.

{\it Step 4}: We now prove the continuity statement. The idea is that all the approximations become increasingly exact as $\epsilon$ (in the definition of the $\epsilon$-very noisy condition) tends to zero. More concretely, for every $\delta>0$, there exists a $\epsilon_1>0$ such that if $P_{e,e'}$ satisfies the $\epsilon_1$-very noisy condition,  then
\begin{equation}
|I(P_e) - I(P_{e'})|<\delta\label{eqn:contMI}
\end{equation}
since mutual information is continuous. For every $\delta>0$, there exists a $\epsilon_2>0$ such that  if $P_{e,e'}$ satisfies the $\epsilon_2$-very noisy condition, then
\begin{equation}
\|Q^*_{e,e'}  -P_{e,e'}\|_{\infty}<\delta, \label{eqn:contconstset}
\end{equation}
since if $P_{e,e'}$ is $\epsilon_2$-very noisy it is close to the constraint set $\{Q:I(Q_{e'})\ge I(Q_{e})\}$ and hence close to the optimal solution $Q^*_{e,e'}$. For every $\delta>0$, there exists a $\epsilon_3>0$ such that if $P_{e,e'}$ satisfies the $\epsilon_3$-very noisy condition, then
\begin{equation}
\left| D(  Q^*_{e,e'}|| P_{e,e'})  - \frac{1}{2}\|Q^*_{e,e'} - P_{e,e'}\|_{P_{e,e'}}^2 \right|<\delta, \label{eqn:dqp_tight}
\end{equation}
which follows from the approximation of the divergence and the continuity statement in \eqref{eqn:contconstset}. For every $\delta>0$, there exists a $\epsilon_4>0$ such that if   $P_{e,e'}$ satisfies the $\epsilon_4$-very noisy condition, then
\begin{equation}
\left|I(P_{e}) -\mathbf{s}_e^T (Q^*_{e,e'}-P_{e,e'}) \right|<\delta, \label{eqn:const_tight}
\end{equation}
which follows from retaining only the first term in the  Taylor expansion of the mutual information in \eqref{eqn:taylor_IQ}. Finally, for every $\delta>0$,  there exists a $\epsilon_5>0$ such that if   $P_{e,e'}$ satisfies the $\epsilon_5$-very noisy condition, then
\begin{equation}
|\tilJ_{e,e'} -J_{e,e'} |<\delta ,  \label{eqn:contJee}
\end{equation}
which follows from continuity of the objective  in the constraints \eqref{eqn:const_tight}. Now choose $\epsilon=\min_{i=1,\ldots, 5} \epsilon_i$ to conclude that for every $\delta>0$, there exists a $\epsilon>0$ such that if  $P_{e,e'}$ satisfies the $\epsilon$-very noisy condition, then \eqref{eqn:contJee} holds. This completes the proof.  \end{IEEEproof}

\section{Proof of Proposition \ref{prop:nonunique}} \label{prf:nonunique}
\begin{IEEEproof}
The following facts about $P$ in Table~\ref{tab:degen} can be readily verified:
\begin{enumerate}
\item $P$ is positive everywhere, {\it i.e.}, $P(\bx)>0$ for all $\bx\in\calX^3$.
\item $P$ is Markov on the complete graph with $d=3$ nodes, hence $P$ is not a tree distribution.
\item  The mutual information between $x_1$ and $x_2$  as a function of $\kappa$ is given by
\begin{equation}
I(P_{1,2})=\log 2+ (1-2\kappa)\log(1-2\kappa)+ 2\kappa\log(2\kappa).\nn
\end{equation}
Thus $I(P_{1,2})\to\log 2=0.693$ as $\kappa\to 0$.
\item For any $(\xi,\kappa)\in(0,1/3)\times(0,1/2)$, $I(P_{2,3})=I(P_{1,3})$ and this pair of mutual information  quantities can be made arbitrarily small as $  \kappa\to 0$.
\end{enumerate}
Thus, for sufficiently small $\kappa>0$, $I(P_{1,2})> I(P_{2,3}) = I(P_{1,3})$. We conclude that the Chow-Liu MWST algorithm will first pick the edge $(1,2)$ and then arbitrarily choose between the two remaining edges: $(2,3)$ or $(1,3)$. Thus, optimal tree structure  is not unique. 
\end{IEEEproof}

%\section{Proof of Theorem~\ref{thm:dom_err_tree}: General Dominant Error Tree}\label{prf:dom_err_tree}
%\begin{IEEEproof}
%The proof of this
%\end{IEEEproof}

\section{Proof of Theorem \ref{thm:consistency} }\label{prf:consistency}
We first state  two preliminary lemmas and prove the first one. Theorem~\ref{thm:consistency} will then be an immediate consequence of these lemmas.
\begin{lemma} \label{lemma:continuity}
Let $X$ and $Y$ be two metric spaces and let $\calK \subset X$ be a compact set in $X$. Let $f:X\times Y\rightarrow \bR$ be a continuous real-valued function. Then the function $g:Y\rightarrow \bR$, defined as
\begin{equation}
g(y)~:=~\min_{x\in \calK} \,\, f(x,y),\quad \forall \,y\in Y, \label{eqn:gy}
\end{equation}
is continuous on $Y$. \end{lemma}
%\begin{remark}
%The minimum in \eqref{eqn:gy} is attained by the extreme value theorem.
%\end{remark}
\begin{IEEEproof}
Set the minimizer in \eqref{eqn:gy} to be
\begin{equation}
x(y)~:=~ \argmin_{x\in \calK} f(x,y) .\label{eqn:xy}
\end{equation}
The optimizer $x(y)\in \calK$ exists since $f(x,y)$ is continuous on $\calK$ for each $y\in Y$ and $\calK$ is compact. This follows from Weierstrauss' extreme value theorem~\cite[Theorem 4.16]{Rudin}. We want to show that for $\lim_{y' \rightarrow y}  g(y')= g(y)$. In other words, we need to prove that
\begin{equation}
\lim_{y' \rightarrow y}   f(x(y'), y')\rightarrow f(x(y),y).
\end{equation}
Consider the difference,
\begin{align}
|f(x(y'), x')- & f(x(y),y)| \le |f(x(y),y)-f(x(y), y')| \nn\\
&+ |f(x(y),y')-f(x(y'), y')|. \label{eqn:diff}
\end{align}
The first term in \eqref{eqn:diff} tends to zero as $y'\rightarrow y$ by the continuity of $f$ so it remains to show that the second term, $B_{y'}:=|f(x(y),y')-f(x(y'), y')|\rightarrow 0$, as $y'\rightarrow y$. Now, we can remove the absolute value since by the optimality of $x(y')$, $f(x(y),y')\ge f(x(y'), y')$. Hence,
\begin{equation}
B_{y'}= f(x(y),y')-f(x(y'), y') .
\end{equation}
Suppose, to the contrary, there exists a sequence $\{y_n'\}_{n=1}^{\infty}\subset Y$ with $y_n' \rightarrow y$ such that
\begin{equation}
f(x(y), y_n') - f(x(y_n') , y_n')> \epsilon >0,\quad\forall\, n\in\mathbb{N}. \label{eqn:contr_eps}
\end{equation}
By the compactness of $\calK$, for the sequence $\{x(y_n')\}_{n=1}^{\infty} \subset \calK$, there exists a subsequence $\{x(y_{n_k}')\}_{k=1}^{\infty} \subset \calK$ whose limit is $x^* = \lim_{k \rightarrow \infty} x(y_{n_k}')$ and $x^*\in \calK$~\cite[Theorem 3.6(a)]{Rudin}. By the continuity of $f$
\begin{align}
\lim_{k\rightarrow \infty} f(x(y) , y_{n_k}') &= f(x(y), y), \label{eqn:lim1}\\
\lim_{k\rightarrow \infty} f(x(y_{n_k}') , y_{n_k}') &= f(x^*, y), \label{eqn:lim2}
\end{align}
since every subsequence of a convergent sequence $\{y_n'\}$ converges to the same limit $y$. Now \eqref{eqn:contr_eps} can be written as
\begin{equation}
f(x(y), y_{n_k}') - f(x(y_{n_k}') , y_{n_k}')> \epsilon >0,\quad\forall\, k\in\mathbb{N}.\label{eqn:contr_eps1}
\end{equation}
We now take the limit as $k\to\infty$ of~\eqref{eqn:contr_eps1}. Next, we use \eqref{eqn:lim1} and \eqref{eqn:lim2} to conclude that
\begin{eqnarray}
f(x(y),y) - f(x^*, y)\!>\! \epsilon \Rightarrow  f(x(y),y) >f(x^*, y)+\epsilon ,
\end{eqnarray}
which contradicts the optimality of $x(y)$ in \eqref{eqn:xy}. Thus, $B_{y'}\to 0$  as $y'\to y$ and  $\lim_{y'\to y}g(y')=g(y)$, which demonstrates the continuity of $g$ on $Y$. % {\it i.e.}, $f(x(y),y) = \min_{x\in \calK} f(x,y)$.
\end{IEEEproof}
\begin{lemma}[The continuous mapping theorem~\cite{Man43}] \label{lem:as}
Let $(\Omega,\calB(\Omega), \nu)$ be a probability  space. Let the sequence of random variables $\{X_n\}_{n=1}^{\infty}$ on $\Omega$ converge $\nu$-almost surely to $X$, {\it i.e.},  $X_n\stackrel{a.s.}{\longrightarrow}X$. Let $g:\Omega\to \bR$ be a  continuous function. Then $g(X_n)$ converges $\nu$-almost surely to $g(X)$, {\it i.e.}, $g(X_n) \stackrel{a.s.}{\longrightarrow} g(X)$.
\end{lemma}
%\begin{IEEEproof}
%Define the (measurable) exception set with respect to the sequence of random variables $\{X_n\}$ and the random variable $X$ as $E(\{X_n\},X):=\{\omega\in\Omega:\lim_{n\to \infty}X_n(\omega)\ne X(\omega)\}\subset \calF(\Omega)$. By the continuity of $g$,
%$$
%\lim_{n\to\infty}g(X_n(\omega))=g\left(\lim_{n\to\infty} X_n(\omega)\right) = g(X(\omega)),
%$$
%for all $ \omega\notin E(\{X_n\},X).$ Since this is true for each $\omega\notin E(\{X_n\},X)$ and $\nu(E(\{X_n\},X))=0$ (by a.s.\ convergence of the sequence $\{X_n\}$), we have established the a.s.\ convergence of the sequence $\{g(X_n)\}$.
%\end{IEEEproof}
%\begin{remark}
%From the above proof, we see that for Lemma~\ref{lem:as} to hold, we only require $g$ to be continuous almost everywhere on $\Omega$ with respect to  the measure $\nu$.
%\end{remark}
\begin{IEEEproof}
Now, using Lemmas~\ref{lemma:continuity} and~\ref{lem:as}, we complete the proof of  Theorem \ref{thm:consistency}.  First we note from \eqref{eqn:Jeehat} that $\hJ_{e,e'}=\hJ_{e,e'}(\hP_{e,e'})$, {\it i.e.}, $\hJ_{e,e'}$ is a function of the empirical distribution on node pairs $e$ and $e'$. Next, we note that $D(Q||P_{e,e'})$ is a continuous function in $(Q,P_{e,e'})$. If $\hP_{e,e'}$ is fixed, the expression~\eqref{eqn:Jeehat} is a minimization of $D(Q|| \hP_{e,e'})$, over the compact set\footnote{Compactness of $\Lambda$ was proven in Theorem \ref{thm:cross_emp} cf.\ Eq.\ \eqref{eqn:Lambda}.}  $\Lambda = \{Q\in\calP(\calX^4) : I(Q_{e'})= I(Q_{e})\}$, hence Lemma~\ref{lemma:continuity} applies (with the identifications  $f\equiv D$ and $\Lambda\equiv\mathcal{K}$) which implies that $\hJ_{e,e'}$ is continuous in the empirical distribution $\hP_{e,e'}$. Since the empirical distribution $\hP_{e,e'}$ converges almost surely to $P_{e,e'}$ \cite[Sec.\ 11.2]{Cov06}, $ \hJ_{e,e'}(\hP_{e,e'})$ also converges almost surely to $J_{e,e'}$, by Lemma~\ref{lem:as}.  %by the continuous mapping theorem~\cite{Bil99}, we conclude that $ \hJ_{e,e'}(\hP_{e,e'})$ also converges in probability to $J_{e,e'}$.
\end{IEEEproof}

%% file: dominantcase.tex
\begin{picture}(150,60)
\put(30,30){\circle{60}}
\put(30,30){\line(1,0){120}}
\put(30,30){\circle*{4}}
\put(150,30){\circle*{4}}
\put(40,30){\circle*{4}}
\put(30,30){\vector(-1,1){14}}
\put(29,40){\makebox (0,0){$\delta$}}
\put(10,4){\makebox (0,0){$Q_{e,e'}^*$}}
\put(57,4){\makebox (0,0){$Q_{e,e'}^{**}$}}
\put(12,12){\vector(1,1){15}}
\put(55,12){\vector(-1,1){15}}
\put(150,37){\makebox (0,0){$P_{e,e'}$}}
\put(68,50){\makebox (0,0){$N_{\delta}(Q_{e,e'}^*)$}}
\put(120,12){\makebox (0,0){$\calP(\calX^4)$}}
\end{picture}

%% file: prfthm5.tex
\begin{picture}(150, 60)
\put(0,0){\line(1,0){50}}
\put(0,0){\line(0,1){50}}
\put(50,0){\line(1,2){25}}
\put(75,50){\line(1,-2){25}}
\put(100,0){\line(1,0){50}}
\put(150,0){\line(0,1){50}}
\multiput(0,50)(4,0){38}{\circle*{1}}
\put(0,0){\circle*{8}}
\put(0,50){\circle*{8}}
\put(50,0){\circle*{8}}
\put(75,50){\circle*{8}}
\put(100,0){\circle*{8}}
\put(150,0){\circle*{8}}
\put(150,50){\circle*{8}}
\put(125,4){\makebox (0,0){$e_2$}}
\put(25,4){\makebox (0,0){$e_1$}}
\put(125,57){\makebox (0,0){$e_2'$}}
\put(25,57){\makebox (0,0){$e_1'$}}
\end{picture}

%% file: TATW_IT09.bbl
\begin{thebibliography}{10}
\providecommand{\url}[1]{#1}
\csname url@samestyle\endcsname
\providecommand{\newblock}{\relax}
\providecommand{\bibinfo}[2]{#2}
\providecommand{\BIBentrySTDinterwordspacing}{\spaceskip=0pt\relax}
\providecommand{\BIBentryALTinterwordstretchfactor}{4}
\providecommand{\BIBentryALTinterwordspacing}{\spaceskip=\fontdimen2\font plus
\BIBentryALTinterwordstretchfactor\fontdimen3\font minus
  \fontdimen4\font\relax}
\providecommand{\BIBforeignlanguage}[2]{{%
\expandafter\ifx\csname l@#1\endcsname\relax
\typeout{** WARNING: IEEEtran.bst: No hyphenation pattern has been}%
\typeout{** loaded for the language `#1'. Using the pattern for}%
\typeout{** the default language instead.}%
\else
\language=\csname l@#1\endcsname
\fi
#2}}
\providecommand{\BIBdecl}{\relax}
\BIBdecl

\bibitem{Tan09isit}
V.~Y.~F. Tan, A.~Anandkumar, L.~Tong, and A.~S. Willsky, ``{A Large-Deviation
  Analysis for the Maximum Likelihood Learning of Tree Structures},'' in
  \emph{Proceedings of IEEE International Symposium on Information Theory},
  Seoul, Korea, Jul 2009, pp. 1140 -- 1144.

\bibitem{Lau96}
S.~Lauritzen, \emph{{Graphical Models}}.\hskip 1em plus 0.5em minus 0.4em\relax
  {Oxford University Press, USA}, 1996.

\bibitem{CL68}
C.~K. Chow and C.~N. Liu, ``Approximating discrete probability distributions
  with dependence trees.'' \emph{IEEE Transactions on Information Theory},
  vol.~14, no.~3, pp. 462--467, May 1968.

\bibitem{Che07}
A.~Chechetka and C.~Guestrin, ``{Efficient Principled Learning of Thin Junction
  Trees},'' in \emph{Proc.\ of Neural Information Processing Systems}, 2007.

\bibitem{Wai06}
M.~J. Wainwright, P.~Ravikumar, and J.~D. Lafferty, ``{High-Dimensional
  Graphical Model Selection Using $l_1$-Regularized Logistic Regression},'' in
  \emph{Neural Information Processing Systems}.\hskip 1em plus 0.5em minus
  0.4em\relax MIT Press, 2006, pp. 1465--1472.

\bibitem{Lee06}
{S. Lee and V. Ganapathi and D. Koller}, ``{Efficient structure learning of
  Markov networks using $l_1$-regularization},'' in \emph{{Neural Information
  Processing Systems}}, 2006.

\bibitem{Joh07}
J.~Johnson, V.~Chandrasekaran, and A.~S. Willsky, ``{Learning Markov Structure
  by Maximum Entropy Relaxation},'' in \emph{{Artificial Intelligence and
  Statistics (AISTATS)}}, 2007.

\bibitem{Mei06}
N.~Meinshausen and P.~Buehlmann, ``{High dimensional graphs and variable
  selection with the Lasso},'' \emph{Annals of Statistics}, vol.~34, no.~3, pp.
  1436--1462, 2006.

\bibitem{Dud04}
M.~Dudik, S.~Phillips, and R.~Schapire, ``Performance guarantees for
  regularized maximum entropy density estimation,'' in \emph{{Conference on
  Learning Theory}}, 2004.

\bibitem{Cho73}
C.~K. Chow and T.~Wagner, ``{Consistency of an estimate of tree-dependent
  probability distributions },'' \emph{IEEE Transactions in Information
  Theory}, vol.~19, no.~3, pp. 369 -- 371, May 1973.

\bibitem{Den00}
F.~D. Hollander, \emph{Large Deviations (Fields Institute Monographs,
  14)}.\hskip 1em plus 0.5em minus 0.4em\relax American Mathematical Society,
  Feb 2000.

\bibitem{West:book}
D.~B. West, \emph{Introduction to Graph Theory}, 2nd~ed.\hskip 1em plus 0.5em
  minus 0.4em\relax Prentice Hall, 2000.

\bibitem{Csis03}
I.~Csisz\'ar and F.~Mat\'u\v{s}, ``Information projections revisited,''
  \emph{IEEE Transactions on Information Theory}, vol.~49, no.~6, pp.
  1474--1490, June 2003.

\bibitem{Bor06}
S.~Borade and L.~Zheng, ``{I-Projection and the Geometry of Error Exponents},''
  in \emph{Allerton Conference on Communication, Control, and Computing}, 2006.

\bibitem{Bor08}
------, ``{Euclidean Information Theory},'' in \emph{Allerton Conference on
  Communication, Control, and Computing}, 2007.

\bibitem{Kar01}
D.~Karger and N.~Srebro, ``{Learning Markov networks: maximum bounded
  tree-width graphs},'' in \emph{Symposium on Discrete Algorithms}, 2001, pp.
  392--401.

\bibitem{Bac02}
F.~Bach and M.~I. Jordan, ``{Thin Junction Trees},'' in \emph{Proc.\ of Neural
  Information Processing Systems}, 2002.

\bibitem{Hec95}
D.~Heckerman and D.~Geiger, ``{L}earning {B}ayesian {N}etworks,'' Microsoft
  Research, Redmond, WA, Tech. Rep. MSR-TR-95-02, December 1994.

\bibitem{Schwarz}
G.~Schwarz, ``{Estimating the dimension of a model},'' \emph{{Annals of
  Statistics}}, vol.~6, pp. 461--464, 1978.

\bibitem{Tib96}
R.~Tibshirani, ``{Regression shrinkage and selection via the Lasso},'' \emph{J.
  Royal. Statist. Soc B.}, vol.~58, no.~1, pp. 267--288, 1996.

\bibitem{Zuk06}
O.~Zuk, S.~Margel, and E.~Domany, ``{On the number of samples needed to learn
  the correct structure of a Bayesian network},'' in \emph{Proceedings of the
  22nd Annual Conference on Uncertainty in Artificial Intelligence (UAI-06)},
  2006.

\bibitem{Choi10}
M.~J. Choi, V.~Y.~F. Tan, A.~Anandkumar, and A.~S. Willsky, ``{Learning Latent
  Tree Graphical Models},'' \emph{submitted to Journal Machine Learning
  Research, on arXiv:1009.2722}, Sep 2010.

\bibitem{Pan03}
L.~Paninski, ``Estimation of entropy and mutual information,'' \emph{{Neural
  Computation}}, vol.~15, pp. 1191 -- 1254, 2003.

\bibitem{Ant01}
A.~Antos and I.~Kontoyiannis, ``Convergence properties of functional estimates
  for discrete distributions,'' \emph{{Random Structures and Algorithms}}, pp.
  163 -- 193, 2001.

\bibitem{Cha05}
J.-R. Chazottes and D.~Gabrielli, ``Large deviations for empirical entropies of
  g-measures,'' \emph{Nonlinearity}, vol.~18, pp. 2545--2563, Nov 2005.

\bibitem{Hut01}
M.~Hutter, ``Distribution of mutual information,'' in \emph{Neural Information
  Processing Systems}.\hskip 1em plus 0.5em minus 0.4em\relax MIT Press, 2001,
  pp. 399--406.

\bibitem{Tan10sp}
V.~Y.~F. Tan, A.~Anandkumar, and A.~S. Willsky, ``{Learning Gaussian Tree
  Models: Analysis of Error Exponents and Extremal Structures},'' \emph{IEEE
  Transactions on Signal Processing}, vol.~58, no.~5, pp. 2701--2714, May 2010.

\bibitem{Kester&Kallenberg:86Stat}
A.~Kester and W.~Kallenberg, ``{Large deviations of estimators},'' \emph{The
  Annals of Statistics}, pp. 648--664, 1986.

\bibitem{Bahadur&Zabell&Gupta:80}
R.~Bahadur, S.~Zabell, and J.~Gupta, ``{Large deviations, tests, and
  estimates},'' \emph{Asymptotic Theory of Statistical Tests and Estimation},
  pp. 33--64, 1980.

\bibitem{Cov06}
T.~M. Cover and J.~A. Thomas, \emph{Elements of Information Theory},
  2nd~ed.\hskip 1em plus 0.5em minus 0.4em\relax Wiley-Interscience, 2006.

\bibitem{Ryu99}
K.~Ryu, ``{Econometric Analysis of Mixed Parameter Models},'' \emph{Journal of
  Economic Theory and Econometrics}, vol.~5, no. 113--124, 1999.

\bibitem{Vantrees}
H.~L.~V. Trees, \emph{Detection, Estimation, and Modulation Theory, Part
  I}.\hskip 1em plus 0.5em minus 0.4em\relax John Wiley \& Sons, 1968.

\bibitem{Csi84}
I.~Csisz\'ar and G.~Tusn\'ady, ``Information geometry and alternating
  minimization procedures,'' \emph{Statistics and Decisions, Supplementary
  Issue No.\ 1}, pp. 205--237, Jul 1984.

\bibitem{Cor03}
T.~Cormen, C.~Leiserson, R.~Rivest, and C.~Stein, \emph{Introduction to
  Algorithms}, 2nd~ed.\hskip 1em plus 0.5em minus 0.4em\relax McGraw-Hill
  Science/Engineering/Math, 2003.

\bibitem{Kruskal}
J.~B. Kruskal, ``{On the Shortest Spanning Subtree of a Graph and the Traveling
  Salesman Problem},'' \emph{Proceedings of the American Mathematical Society},
  vol.~7, no.~1, Feb 1956.

\bibitem{Prim}
R.~C. Prim, ``{Shortest connection networks and some generalizations},''
  \emph{Bell System Technical Journal}, vol.~36, 1957.

\bibitem{Rudin}
W.~Rudin, \emph{Principles of Mathematical Analysis}, 3rd~ed.\hskip 1em plus
  0.5em minus 0.4em\relax McGraw-Hill Science/Engineering/Math, 1976.

\bibitem{Tan10jmlr}
V.~Y.~F. Tan, A.~Anandkumar, and A.~S. Willsky, ``{Learning High-Dimensional
  Markov Forest Distributions: Analysis of Error Rates},'' \emph{submitted to
  Journal Machine Learning Research, on arXiv:1005.0766}, May 2010.

\bibitem{Huber}
P.~J. Huber and V.~Strassen, ``Minimax tests and the neyman-pearson lemma for
  capacities,'' \emph{Annals of Statistics}, vol.~1, pp. 251--263.

\bibitem{Pandit}
C.~Pandit and S.~P. Meyn, ``{Worst-case large-deviations with application to
  queueing and information theory},'' \emph{Stochastic Processes and
  Applications}, vol. 116, no.~5, pp. 724–--756, May 2006.

\bibitem{Zeitouni&Gutman:91IT}
O.~Zeitouni and M.~Gutman, ``{On Universal Hypotheses Testing via Large
  Deviations},'' \emph{IEEE Transactions on Information Theory}, vol.~37,
  no.~2, pp. 285--290, 1991.

\bibitem{Unnikrishnan}
J.~Unnikrishnan, D.~Huang, S.~Meyn, A.~Surana, and V.~V. Veeravalli,
  ``Universal and composite hypothesis testing via mismatched divergence,''
  \emph{IEEE Transactions on Information Theory}, revised May 2010, on arXiv
  http://arxiv.org/abs/0909.2234.

\bibitem{Tan10:ISIT}
V.~Y.~F. Tan, A.~Anandkumar, and A.~S. Willsky, ``{Error Exponents for
  Composite Hypothesis Testing of Markov Forest Distributions},'' in
  \emph{International Symposium on Information Theory}, Austin, TX, June 2010,
  pp. 1613 -- 1617.

\bibitem{Abb08}
E.~Abbe and L.~Zheng, ``{Linear Universal Decoding for Compound Channels: an
  Euclidean Geometric Approach},'' in \emph{International Symposium on
  Information Theory}, 2008, pp. 1098--1102.

\bibitem{Pin64}
M.~S. Pinsker, \emph{{Information and Information Stability of Random
  Variables}}.\hskip 1em plus 0.5em minus 0.4em\relax Oakland, CA: Holden-Day,
  1964.

\bibitem{Lan06}
J.~N. Laneman, ``{On the Distribution of Mutual Information},'' in
  \emph{Information Theory and Applications Workshop}, 2006.

\bibitem{Ama00}
S.-I. Amari and H.~Nagaoka, \emph{Methods of Information Geometry}.\hskip 1em
  plus 0.5em minus 0.4em\relax American Mathematical Society, 2000.

\bibitem{Ama01}
S.-I. Amari, ``Information geometry on hierarchy of probability
  distributions,'' \emph{IEEE Transactions on Information Theory}, vol.~47,
  no.~5, pp. 1701--1711, 2001.

\bibitem{Man43}
H.~B. Mann and A.~Wald, ``{On the statistical treatment of linear stochastic
  difference equations},'' \emph{{Econometrica}}, vol.~11, pp. 173--220, 1943.

\bibitem{Billingsley}
P.~Billingsley, \emph{Weak Convergence of Measures: Applications in
  Probability}.\hskip 1em plus 0.5em minus 0.4em\relax Society for Industrial
  Mathematics, 1987.

\bibitem{Csi97}
I.~Csisz\'ar and J.~Korner, \emph{Information Theory: Coding Theorems for
  Discrete Memoryless Systems}.\hskip 1em plus 0.5em minus 0.4em\relax
  Akademiai Kiado, 1997.

\end{thebibliography}
